\documentclass[twoside,11pt]{article}

\usepackage{blindtext}

\usepackage{jmlr2e}

\usepackage{graphicx}     
\usepackage{tikz}         
\usepackage{float}        
\usepackage{booktabs}     

\usepackage{multicol}     
\usepackage{multirow}     
\usepackage{tabularx}     
\usepackage{array}        
\usepackage{colortbl}     
\usepackage{makecell}     
\usepackage{bm}           

\usepackage{mathtools}    
\usepackage{cancel}       
\usepackage{siunitx}      
\usepackage{empheq}       
\usepackage{latexsym}     
\usepackage{bbm}          
\usepackage{calc}         

\usepackage{comment}      
\numberwithin{equation}{section}
\newtheorem{assumption}[theorem]{Assumption}

\usepackage{lastpage}

\ShortHeadings{Error Bounds for Kernel-Based Statistics}{Ni and Huo}
\firstpageno{1}

\begin{document}

\title{A Uniform Concentration Inequality for Kernel-Based Two-Sample Statistics}

\author{\name Yijin Ni \email yni64@gatech.edu \\
       \name Xiaoming Huo \email huo@gatech.edu \\
       \addr School of Industrial and Systems Engineering\\
       Georgia Institute of Technology\\
       Atlanta, GA 30318, USA}

\editor{}

\maketitle

\begin{abstract}
In many contemporary statistical and machine learning methods, one needs to optimize an objective function that depends on the discrepancy between two probability distributions.
The discrepancy can be referred to as a metric for distributions. 
Widely adopted examples of such a metric include Energy Distance (ED), distance Covariance (dCov), Maximum Mean Discrepancy (MMD), and the Hilbert–Schmidt Independence Criterion (HSIC).
We show that these metrics can be unified under a general framework of kernel-based two-sample statistics.

This paper establishes a novel uniform concentration inequality for the aforementioned kernel-based statistics. 
Our results provide upper bounds for estimation errors in the associated optimization problems, thereby offering both finite-sample and asymptotic performance guarantees. 
As illustrative applications, we demonstrate how these bounds facilitate the derivation of error bounds for procedures such as distance covariance-based dimension reduction, distance covariance-based independent component analysis, MMD-based fairness-constrained inference, MMD-based generative model search, and MMD-based generative adversarial networks.
\end{abstract}


\section{Introduction}

A common task in many contemporary statistical and machine learning frameworks is to optimize an objective function predicated on the distance between two probability distributions.
In many cases, the distance is measured using statistics such as 
Energy Distance (ED; \cite{szekely2013energy}), 
distance Covariance (dCov; \cite{szekely2009brownian}), 
Maximum Mean Discrepancy (MMD, \cite{gretton2012kernel}), 
and Hilbert-Schmidt Independence Criterion (HSIC; \cite{gretton2007kernel}). 
It has been shown that a kernel-based two-sample statistics approach unifies all these seemingly different statistics that have appeared (\cite{sejdinovic2013equivalence}). 


To describe our key technical contribution, we introduce notations for two random variables and their transforms. 
Given a pair of random vectors $X \in \mathcal{X}$ and $Y \in \mathcal{Y}$, let $\mathbf{X}:=(X_1, \dots, X_m)^T$ and $\mathbf{Y}:=(Y_1, \dots, Y_n)^T$ be the data matrices for $X$ and $Y$, respectively, where $X_i \stackrel{\text{i.i.d.}}{\sim} X$ and $Y_j \stackrel{\text{i.i.d.}}{\sim} Y$, $\forall i, j$.
Denote the involved kernel-based two-sample statistic based on observations $\mathbf{X}$ and $\mathbf{Y}$ as $f(\mathbf{X}, \mathbf{Y})$, let $\mathcal{H}$ be a function set mapping from $\mathcal{X} \times \mathcal{Y}$ to $\mathcal{U}\times \mathcal{U}$, where $\mathcal{U}\subseteq \mathbb{R}^d$, that is,
\begin{align*}
    \mathcal{H} := \{h \mid h(X, Y) = (h_\mathcal{X}(X), h_\mathcal{Y}(Y)),\, h_\mathcal{X}: \mathcal{X} \mapsto \mathcal{U}, h_\mathcal{Y}: \mathcal{Y} \mapsto \mathcal{U}\}.
\end{align*}
In this work, we focus on optimization problems over $\mathcal{H}$, and the empirical two-sample statistic $f(h(\mathbf{X}, \mathbf{Y}))$ is a part of the objective function.

Our primary contribution is a novel uniform convergence inequality for general two-sample statistics, encompassing kernel-based metrics as special cases. 
More specifically, we intend to establish the following inequality:
\begin{equation}
\label{eq:intro_general}
\begin{aligned}
    \operatorname{Pr}\left(\sup _{h \in \mathcal{H}} \left| f(h_\mathcal{X}(\mathbf{X}), h_\mathcal{Y}(\mathbf{Y})) - \mathbb{E}f(h_\mathcal{X}(\mathbf{X}), h_\mathcal{Y}(\mathbf{Y}))\right| \geq C_2(m, n, \mathcal{H}, f, \delta)\right) \leq \delta, 
\end{aligned}
\end{equation}
where, $h_\mathcal{X}(\mathbf{X}):=(h_\mathcal{X}(X_1), \dots, h_\mathcal{X}(X_m))^T$ and $h_\mathcal{Y}(\mathbf{Y}):=(h_\mathcal{Y}(Y_1), \dots, h_\mathcal{Y}(Y_n))^T$, respectively.
The constant $C_2(m, n, \mathcal{H}, f, \delta)$ depends solely on the sample sizes $m$, $n$, the function set $\mathcal{H}$, the involved two-sample statistic $f(\cdot, \cdot)$, and the related probability $\delta$.
Notably, the statistic $f$ is not restricted to a specific formulation, and $\mathcal{H}$ includes any function set with finite Gaussian or Rademacher complexities, such as linear projections and neural networks.

By modeling kernel-based two-sample statistics—ED, dCov, MMD, and HSIC—as specific instances of inequality \eqref{eq:intro_general}, we show that uniform concentration inequalities hold for these statistics, under mild assumptions on their reproducing kernel functions.
These inequalities are essential for establishing the consistency of statistical and machine learning methods that utilize these statistics.
As illustrative examples, we will present the following applications in this work (detailed discussions are provided in Section \ref{sec:motivate}):
\begin{enumerate}
\item {\it dCov-based Dimension Reduction.}
Distance covariance is utilized to reduce the dimensionality of data by projecting vectors $X$ and $Y$ into lower-dimensional spaces while preserving their dependence, as demonstrated in \cite{sheng2013direction, sheng2016sufficient, chen2019sufficient, yu2019distance}.

\item {\it dCov-based Independent Component Analysis.}
Minimizing distance covariance serves as a tool for independent component analysis in \cite{matteson2017independent}.

\item {\it Generalized MMD with Kernel Selection.}
To introduce an approach that helps in the selection of a reproducing kernel, \cite{fukumizu2009kernel} proposes a generalized version of MMD.

\item {\it MMD-based Fairness Imposed Inference.}
MMD is incorporated as a penalty to achieve fair representation in supervised learning, as explored in \cite{rychener2022metrizing}.

\item {\it MMD-based Generative Model Search.}
The MMD-based Minimum Distance Estimator (MDE, \cite{bickel1976another}) is employed to search for effective generative models, as studied in \cite{dziugaite2015training, li2017mmd, briol2019statistical, cherief2022finite}.

\item {\it MMD-based Generative Adversarial Network (GAN).}
In MMD-based GANs \cite{li2017mmd, binkowski2018demystifying}, MMD serves as the objective function. The generator minimizes the MMD between the generated and target distributions, while the discriminator maximizes this discrepancy to improve the quality of the generated data.
\end{enumerate}


The rest of the paper is organized as follows. 
In Section \ref{sec:motivate}, we provide detailed explanations for the aforementioned exemplary applications involving kernel-based statistics.
These applications motivate the development of inequality \eqref{eq:intro_general}, which is essential for deriving estimation error bounds.
In Section \ref{sec:preliminary}, we review existing works contributing to the progress of uniform concentration inequalities and discuss their difference with our intentional result \eqref{eq:intro_general}.
In Section \ref{sec:general}, we present Theorem \ref{thm:general}, which provides the explicit formula for inequality \eqref{eq:intro_general}, analyzing its assumptions and extensions.
In Section \ref{sec:kernel-based}, we derive uniform concentration inequalities for ED, dCov, MMD, and HSIC based on our general inequality (Theorem \ref{thm:general}), providing the explicit formulas in specific settings.
In Section \ref{sec:consistency_empirical}, we apply our main result (Theorem \ref{thm:general}) to derive estimation error bounds for the specific application examples described in Section \ref{sec:motivate}.
In Section \ref{sec:discussion1}, we discuss possible alternative concentration inequalities and advantages of our approach in this paper. 
We give some conclusion remarks in Section \ref{sec:conclude}. 
When possible, the technical proofs are relegated to Section \ref{sec:proofs}. 


\section{Motivation}
\label{sec:motivate}

Optimizing an objective function based on the distance between two statistical distributions is widely used in modern statistical and machine learning practices.
In many cases, the distance is measured using a kernel-based statistic. 
As an explanatory example, we present the definition of Maximum Mean Discrepancy (MMD) in Section \ref{sec:define-MMD}, which is a highly utilized kernel-based statistic. 

While using MMD as a representative one, in this paper, we will take a kernel-based two-sample statistics approach.
We will argue that our approach unifies many different statistics that have appeared, including Energy Distance (ED; \cite{szekely2013energy}), 
distance Covariance (dCov; \cite{szekely2009brownian}), 
Maximum Mean Discrepancy (MMD; \cite{gretton2012kernel}), and Hilbert-Schmidt Independence Criterion (HSIC; \cite{gretton2007kernel}).
More specifically, the formulation of MMD can be considered as a general expression for the aforementioned statistics.
Details of our unification can be found in Section \ref{sec:kernel-based}.
For simplicity, in the remainder of this paper, we will also call them MMD-like (or MMD equivalent) statistics.

In the remainder of this section, we describe scenarios in which MMD-like statistics are called into play in the objective function of an optimization problem. 
These associated newly emerged statistical and machine learning approaches include distance covariance-based dimension reduction (Section \ref{sec:dimension-reduction-2}),
distance covariance-based independent component analysis (Section \ref{sec:dis-ica}), 
generalized MMD with kernel selection (Section \ref{sec:generalize-MMD-kernel}), 
fairness imposed inference (Section \ref{sec:fairness-2}), 
MMD-based generative model (Section \ref{sec:mmd-estimators}), and MMD-based
generative adversarial network (GAN) (Section \ref{sec:gan-2}).

\subsection{Definition of Maximum Mean Discrepancy and corresponding estimators} 
\label{sec:define-MMD}

The definition of Maximum Mean Discrepancy (\cite{gretton2012kernel}) is given below. 
\begin{definition}[Maximum Mean Discrepancy]
\label{def:MMD}
    Given probability metrics $P$ and $Q$ embedded in the set $\mathcal{U}$ and random vectors $X \sim P$, $Y \sim Q$. 
    Let $\mathcal{F}$ be a {Reproducing Kernel Hilbert Space} (RKHS) equipped with inner product $\langle \cdot, \cdot \rangle_{\mathcal{F}}$ and norm $\|\cdot\|_{\mathcal{F}}$. 
    Let $k: \mathcal{U} \times \mathcal{U} \mapsto \mathbb{R}$ be the \textit{reproducing kernel} of $\mathcal{F}$.
    The MMD between random vectors $X$ and $Y$, denoted as $\gamma_k(X, Y)$, is defined as the 
    $\mathcal{F}$-distance between the kernel mean embeddings of the corresponding probability metrics. 
    More specifically, we have 
    \begin{equation}
    \label{eq:MMD}
        \gamma_k(X, Y):= \left\|\mathbb{E}_{X \sim P} k(\cdot, X)- \mathbb{E}_{Y \sim Q} k(\cdot, Y) \right\|_{\mathcal{F}}.
    \end{equation}
    \end{definition}

We introduce the formulas for empirical estimators of MMD.
In \cite{gretton2012kernel}, three different estimators of $\gamma_k(X, Y)$ are proposed, including an unbiased estimate, a biased estimate, and a U-statistic estimate.
In the following, we denote the unbiased estimate as $\widehat{\gamma}_{k,u}$, where $u$ refers to `unbiased', the biased estimate as $\widehat{\gamma}_{k,b}$, where $b$ refers to `biased', and the U-statistic estimate as $\widehat{\gamma}_{k,e}$, where $e$ refers to the fact that this estimate is employed in circumstances where the sample sizes collected from $X$ and $Y$ are equal.

\begin{definition}[Empirical MMD Estimators, \cite{gretton2012kernel}] 
\label{def:empiricalMMDestimators}
Given sample matrices $\mathbf{X}:=(X_1,\dots, X_m)^T$ and $\mathbf{Y}:=(Y_1,\dots, Y_n)^T$, where $X_i \stackrel{i.i.d.}{\sim} X$, $Y_j \stackrel{i.i.d.}{\sim} Y$, $\forall i, j$, the empirical estimators of $\gamma_k(X, Y)$ are defined as follows:
\begin{enumerate}
    \item \text{Unbiased Estimator $\widehat{\gamma}_{k,u}$.}
    We have 
    \begin{equation}
    \label{eq:intro_unbiased_MMD}
    \begin{aligned}
        &\widehat{\gamma}_{k, u}^2(\mathbf{X}, \mathbf{Y}) = \frac{\sum_{i\neq j}^m k(X_i, X_j)}{m(m-1)} + \frac{\sum_{i\neq j}^n  k(Y_i, Y_j)}{n(n-1)} - \frac{2\sum_{i,j=1}^{m,n} k(X_i, Y_j)}{mn}.
    \end{aligned}
    \end{equation}
    \item \text{Biased Estimator $\widehat{\gamma}_{k,b}$.}
    We have 
    \begin{equation}
    \label{eq:intro_biased_MMD}
    \begin{aligned}
        &\widehat{\gamma}_{k,b}^2(\mathbf{X}, \mathbf{Y}) = \frac{\sum_{i, j=1}^m k\left(X_i, X_j\right)}{m^2} +\frac{\sum_{i, j=1}^n k\left(Y_i, Y_j\right)}{n^2} -\frac{2\sum_{i, j=1}^{m, n} k\left(X_i, X_j\right)}{m n}.
    \end{aligned}
    \end{equation}
    \item \text{U-statistic Estimator $\widehat{\gamma}_{k,e}$.} 
    Suppose $m = n$, we have 
    \begin{align}
    \label{eq:intro_U_MMD}
        \widehat{\gamma}_{k,e}^2(\mathbf{X}, \mathbf{Y})= \frac{1}{n(n-1)} \sum_{i=1}^n \sum_{j \neq i}^n h(Z_i, Z_j).
    \end{align}
    Here, $Z_i = (X_i, Y_i)$, $i=1,\dots,n$. For any $i, j$, the expression of $h\left(Z_i, Z_j\right)$ is given as follows:
    \begin{align*}
      h\left(Z_i, Z_j\right) =  k\left(X_i, X_j\right)+k\left(Y_i, Y_j\right)-k\left(X_i, Y_j\right)-k\left(X_j, Y_i\right).
    \end{align*}
    \end{enumerate}
\end{definition}

\subsection{Distance covariance-based dimension reduction}
\label{sec:dimension-reduction-2}

Given a response vector $Y\in \mathbb{R}^q$ and a predictor vector $X \in \mathbb{R}^p$.
In \cite{sheng2013direction, sheng2016sufficient, chen2019sufficient, yu2019distance}, distance covariance is applied to reduce the dimension of the pair of random vectors $X$ and $Y$ without loss of mutual information.
More specifically, given a matrix $\boldsymbol{\beta} \in \mathbb{R}^{p \times d}$, $d < p$, such that $\boldsymbol{\beta}^T\boldsymbol{\beta}=I_d$ and $Y \perp\!\!\!\perp X | \boldsymbol{\beta}^TX$, where $\perp\!\!\!\perp$ means independence, then the column space of $\boldsymbol{\beta}$ forms a dimension reduction subspace for $X$ (\cite{cook1996graphics}).
That is, the projected vector $\boldsymbol{\beta}^TX$ preserves the information carried by the predictor $X$, reducing the dimension from $p$ to $d$.
In \cite{yu2019distance}, the estimation of matrix $\boldsymbol{\beta}$ is decomposed into a stepwise procedure.
More specifically, in \cite{yu2019distance}, denote the current empirical estimate of the subspace ${\mathcal{S}}$ as $\widehat{\mathcal{S}}_{\text{old}}$.
Then in the next step, one tried to update the empirical estimate for subspace ${\mathcal{S}}$ through the following optimization problem:
\begin{equation}
\label{eq:ustar}
    u^*:=\max_{\|u\|=1, u \in \widehat{\mathcal{S}}_{\text{old}}^\perp}\mathcal{V}^2(X^Tu, Y), 
\end{equation}
where $\mathcal{V}^2(X^Tu, Y)$ is the distance covariance between the projected random variable $X^Tu$ and random variable $Y$.
If $\mathcal{V}^2(X^Tu^*, Y)= 0$, which means that $X^Tu^* \perp\!\!\!\perp Y$, we consider $\widehat{\mathcal{S}}_{\text{old}}$ as the final estimate of the subspace $\mathcal{S}$, otherwise, we update $\widehat{\mathcal{S}}_{\text{new}}:=\text{span}(u^*, \widehat{\mathcal{S}}_{\text{old}})$.

The objective function on the right-hand side of \eqref{eq:ustar} is a distance measure. 
It measures the distance between the joint distribution of random variable $X^Tu$ and random vector $Y$ and another joint distribution of the same two components, however, assuming the independence between them. 

The connection between dCov, i.e., $\mathcal{V}^2(\cdot, \cdot)$, and MMD, i.e., $\gamma_k(\cdot, \cdot)$, will be discussed in Section \ref{sec::dbstats}.
In particular, in \cite{sejdinovic2013equivalence}, the equivalence between RKHS-based statistics and distance-based statistics is proved.
That is, given a distance-induced reproducing kernel, the ED is equivalent to MMD, and the dCov is equivalent to HSIC.
More details on the unification of four statistics (i.e., energy distance, distance covariance, maximum mean Discrepancy, and Hilbert-Schmidt independence criterion) will be presented in Section \ref{sec:kernel-based}. 

In \cite{yu2019distance}, our uniform concentration inequality for dCov (Corollary \ref{cor:dCov}) is utilized to establish the non-asymptotic statistical consistency for the corresponding empirical optimizer, providing theoretical evidence for the feasibility of the proposed method.
In comparison, this paper's inequality is more general. It is built for a generic kernel-based two-sample statistic, including the MMD, HSIC, ED, and dCov.

\subsection{Distance covariance-based independent component analysis}\label{sec:dis-ica}
Since the distance covariance equals $0$ if and only if the involved random variables are mutually independent.
\cite{matteson2017independent} applied distance covariance $\mathcal{V}^2$ as an independent component analysis tool.
More specifically, given a random vector $Y \in \mathbb{R}^d$, we suppose its covariance matrix is the identity matrix $\mathbf{I}_d$.
In this setting, one aimed to search for an orthogonal matrix $\mathbf{W} \in \mathbb{R}^{d \times d}$, such that the components of the projected random vector, i.e., $S:= \mathbf{W}Y \in \mathbb{R}^d$, are mutually independent, or as close as possible with respect to the value of dCov.

As discussed in \cite{matteson2017independent}, every orthogonal matrix with determinant equal to $1$ corresponds to a product of Givens matrices parameterized by the rotation angles $\theta$, which is a length $p:= d(d-1)/2$ vectorized triangular array of rotation angles, indexed by $\{i,j | 1 \leq i < j \leq d\}$, where $\theta_{k,j} \in [0, \pi)$, $\forall k > 1, j$, and $\theta_{i, j} \in [0, 2\pi)$, $\forall j$.

Denote the projected random vector $S$ utilizing a rotation matrix equipped with angles $\theta$ as $S(\theta)$.
To find the vector $S(\theta)$, which has mutually independent components, the following optimization problem is considered:
\begin{equation}
\label{eq:matteson}
    \theta^* := \min_{\theta}\sum_{k=1}^{d-1}\mathcal{V}^2(S_k(\theta), S_{k^+}(\theta)),
\end{equation}
where $S_k(\theta)$ denotes the $k$-th component of $S(\theta)$, and $S_{k^+}(\theta) \in \mathbb{R}^{d-k}$ denotes the combination from the $k+1$-th to $d$-th components, $\forall k=1,\dots,d-1$.

Note that the optimization problem \eqref{eq:matteson} is complicated.
It is mentioned in \cite[Equation (7)]{matteson2017independent} that the optimization problem \eqref{eq:matteson} can be decomposed as the following series of optimization problems:
\begin{equation}
\label{eq:mattesson_seq}
    \widehat{\theta}_n^{(k: k)}=\left\{\widehat{\theta}_{k, \ell}: k<\ell \leq d\right\}=\underset{\theta^{(k: k)}}{\operatorname{argmin}}\, \mathcal{V}^2\left(S_k(\theta), S_{k^{+}}(\theta)\right),\quad \forall k=1,\dots,d-1,
\end{equation}
in which $\theta^{(1:(k-1))}$ are fixed at $\widehat{\theta}_n^{(1:(k-1))}$ and all elements in $\theta^{((k+1): d)}$ are fixed, but arbitrary.
This sequential manner of estimation leads to a tradeoff between computational complexity and statistical efficiency.

In \cite{matteson2017independent}, the asymptotic consistency of the above empirical estimator $\widehat{\theta}$ is provided, while the convergence rate is not confirmed.
In that case, the uniform concentration inequalities established in this work fill this gap, providing an explicit convergence rate that shows the strength of dCov-based independent component analysis methods compared with those based on other metrics.
We refer to the original paper (\cite{matteson2017independent}) for more technical details. 

\subsection{Generalized MMD with kernel selection}
\label{sec:generalize-MMD-kernel}

Recall the definition of MMD, $\gamma_k(X, Y)$, introduced in Definition \ref{def:MMD}.
Note that the behavior of MMD is heavily influenced by the choice of the corresponding reproducing kernel $k$.
For instance, in an extreme case where $k(u, u') = 0$, $\forall u, u' \in \mathcal{U}$, we have $\gamma_k(P, Q) = 0$, $\forall P, Q$, that is, $\gamma_k$ cannot distinguish distributions.

To introduce an approach that helps to choose a selection of reproducing kernels, \cite{fukumizu2009kernel} proposes a generalized version of MMD.
The technical definition is given as follows:
\begin{definition}[Generalized MMD, \cite{fukumizu2009kernel}]
\label{def:intro_generalizedMMD}
    Denote a class of reproducing kernel functions as $\mathcal{K}$. A generalized MMD is defined as follows:
    \begin{equation}
    \label{eq:intro_fukumizuMMD}
        \gamma(X, Y; \mathcal{K}) := \sup_{k \in \mathcal{K}}\gamma_k(X, Y).
    \end{equation}
    Given data matrices $\mathbf{X}:=(X_1, \dots, X_m)^T$ and $\mathbf{Y}:=(Y_1, \dots, Y_n)^T$, the sample estimate of the generalized MMD $\gamma(X, Y)$ is defined as follows.
    \begin{equation}
    \label{eq:intro_fukumizuMMD_b}
        \widehat{\gamma}_{b}(\mathbf{X}, \mathbf{Y}; \mathcal{K}):= \sup_{k \in \mathcal{K}}\widehat{\gamma}_{k,b}(\mathbf{X}, \mathbf{Y}).
    \end{equation}
\end{definition}
As can be seen, an MMD-like statistic is used in the objective function. 

Note that in \cite{fukumizu2009kernel}, an inequality similar to our uniform concentration inequality for MMD (Corollary \ref{coro:MMD}) is built.
Nevertheless, as mentioned in \eqref{eq:intro_general}, a uniform error bound with respect to a general function class $\mathcal{H}$ is considered in this work.
In contrast, an inequality considering a specified class of reproducing kernel functions is considered in \cite{fukumizu2009kernel}.
Based on the symmetric property of kernel functions, the scenario considered in this paper is not equivalent to the one mentioned in \cite{fukumizu2009kernel}.

We refer to the original paper (\cite{matteson2017independent}) for more technical details.

\subsection{Fairness imposed inference}\label{sec:fairness-2}

We consider an approach investigated in \cite{rychener2022metrizing}, where MMD is used as a penalty to help achieve fair representation in a supervised learning problem.
Given a regression or classification problem,  we consider an empirical risk minimization approach that solves the following optimization problem, 
$$
\min _{h \in \mathcal{H}} \mathbb{E}[L(h(X), Y)],
$$
which aims to predict a property $Y \in \mathcal{Y} \subseteq \mathbb{R}$ (the output) of a human being characterized by a feature vector $X \in \mathcal{X} \subseteq \mathbb{R}^d$ (the input). 
Here, $\mathcal{H}$ represents a family of Borel-measurable hypotheses $h: \mathcal{X} \rightarrow \mathbb{R}$, and $L: \mathbb{R} \times \mathbb{R} \rightarrow \mathbb{R}_{+}$ represents a lower semi-continuous loss function that quantifies the discrepancy between the predicted output $h(X)$ and the actual output $Y$. 
Variable $A \in \mathcal{A}=\{0,1\}$ is a protected attribute that encodes their race, religion, age, sex, etc.

In \cite{rychener2022metrizing}, to seek for a classifier or regressor $h(X)$ such that the information collected from $A$ is not used, the following constraint is imposed: 
$$
\gamma_k^2(h(X|A=0), h(X|A=1)) \leq \epsilon,
$$
where $\epsilon$ is a pre-defined hyperparameter.
In a special case where $\epsilon = 0$, the above constraint is equal to the statement that the probability distribution of random vector $h(X|A=0)$ is the same as that of $h(X|A=1)$.
That is, the predicted output $h(X)$ ignores the impact from the attribute $A$.
Numerically, the corresponding regressor or classifier $h$ is achieved by the following fair learning problem \cite[Equation (3)]{rychener2022metrizing}:
\begin{equation}
\label{eq:fair}
\min _{h \in \mathcal{H}} \mathbb{E}\left[L\left(h(X), Y\right)\right]+\rho\left(\gamma_k^2(h(X|A=0), h(X|A=1))\right),
\end{equation}
where $\rho: \mathbb{R}_{+} \rightarrow \mathbb{R}_{+}$ is a smoothing regularization problem.
Again, we see that an MMD-like statistic is adopted in the objective function.

\cite{rychener2022metrizing} adopts a neural network as the model $h$ and displays a consistently high accuracy and low unfairness of the MMD approach. 
The theoretical consistency of the first term can be proved via the law of large number numbers. 
However, the theoretical consistency of the MMD estimator in \cite{rychener2022metrizing} could be improved. 
The result in this paper will fill in this vacancy, seeing more details in Section \ref{sec::fairness}.

\subsection{MMD-based search for a generative model}\label{sec:mmd-estimators}

Given a random vector $Y$ embedded in $\mathbb{R}^q$.
Define the corresponding data matrix $\mathbf{Y}:=(Y_1,\dots,Y_n)^T$ $\in \mathbb{R}^{n \times q}$. 
A generative model aims to replicate the random vector $Y$ by generating a new random vector $g(X)$, where $g(X)$ is derived from a noisy random vector $X$ through the function $g$.
Typically, the distribution of $X$ is chosen as a standard Gaussian distribution $\mathcal{N}(0,1)$ or a uniform distribution $\operatorname{Unif}[0,1]$.

In \cite{bickel1976another}, the {Minimum Distance Estimator} (MDE) is proposed as an approach to explore the optimum generating function $h \in \mathcal{H}$.
More specifically, suppose $h$ belongs to a parametric function class $\{h_{\theta}: \mathcal{X} \mapsto \mathbb{R}^q \mid \theta \in \Theta\}$.
Denote the random vector following the empirical distribution gathered from data matrix \( \mathbf{Y} \) as \( \boldsymbol{Y}_n \), where $n$ refers to the sample size.
Given a probability metric $d$, the MDE explores the optimal generating function $h_\theta$ through the following minimization problem:
\begin{align}
\label{eq:intro_MDE}
    \widehat{\theta}_n := \arg\min_{\theta \in \Theta} d(\boldsymbol{Y}_n, h_\theta(X)).
\end{align}
As one can see, the objective function measures the distance between two distributions, which is the setting that this paper will consider.

In \cite{dziugaite2015training, li2017mmd, briol2019statistical, cherief2022finite}, the MMD-based MDE is considered; that is, the following estimator is employed:
\begin{equation}
    \label{eq:intro_MDE(MMD,n)}
    \widehat{\theta}_{n} := \arg\min_{\theta \in \Theta} {\gamma}_k\left(\boldsymbol{Y}_n, h_\theta(X)\right).
\end{equation}
The following generalization error bound is established for the empirical estimator $\widehat{\theta}_{n}$:
\begin{equation}
\label{eq:intro_briol}
    \gamma_k\left(h_{\widehat{\theta}_n}(X), Y\right) \leq \inf_{\theta \in \Theta}\gamma_k\left(h_\theta(X), Y\right) + 2\sqrt{\frac{2}{n}\sup_{x\in \mathcal{X}}k(x,x)}\left(1+\sqrt{\log(1/\delta)}\right).
\end{equation}

Nevertheless, in constructing the estimator $\hat{\theta}_n$, we have neglected the discrepancy between the distribution of the generated random vector $h_\theta(X)$ and its empirical distribution $\frac{1}{m} \sum_{i=1}^m \boldsymbol{\delta}_{h_\theta\left(X_i\right)}$ derived from a finite sample.
More specifically, as noted in \cite{mohamed2016learning}, computing the distribution \( P_\theta \) of the generated random vector \( h_\theta(X) \) is challenging when the generating function \( h_\theta \) is complex, as is often the case in the machine learning context.
That is, the estimation of $\widehat{\theta}_n$ cannot be obtained based on the analytical expression of the distribution for \( h_\theta \).
To address this issue, \cite{briol2019statistical, cherief2022finite} approximates the distribution of $h_\theta(X)$ through its empirical probability measure.
Consequently, instead of \eqref{eq:intro_MDE(MMD,n)}, the explicit formulation of the employed estimator is the one given as follows, which accounts for the finite sample size $m$ gathered from the generated random vector $h_\theta(X)$:
\begin{align}
\label{eq:intro_MDE(MMD,m,n)}
    \widehat{\theta}_{m,n} := \arg\min_{\theta \in \Theta} \widehat{\gamma}_{k,u}^2\left(h_\theta(\mathbf{X}), \mathbf{Y}\right).
\end{align}
Here, the unbiased empirical estimate $\widehat{\gamma}_{k,u}^2$ \eqref{eq:intro_unbiased_MMD} in the above expression can be replaced by the biased estimate $\widehat{\gamma}_{k,b}^2$ \eqref{eq:intro_biased_MMD}. 
When $m = n$, it can also be substituted with the U-statistic estimate $\widehat{\gamma}_{k,e}^2$ \eqref{eq:intro_U_MMD}.
For simplicity, we adopt a generalized notation $\widehat{\gamma}_k^2$ to represent the above three empirical estimates in the remainder of this paper.

In Section \ref{sec:::generative}, we will show how our main result can be utilized to derive a generalization error bound for the estimator given in \eqref{eq:intro_MDE(MMD,m,n)}.
To the best of our knowledge, the derived error bound has not been established in related literature.

\subsection{Generative adversarial network}\label{sec:gan-2}
In MMD-based generative adversarial models \cite{li2015generative, li2017mmd, briol2019statistical, cherief2022finite}, the goal is to simultaneously learn an optimal generator $h$ and an MMD-based two-sample test statistic that measures the discrepancy between real and generated data.
More specifically, given a class of reproducing kernels \(\mathcal{K}\) and a class of generating functions \(\mathcal{H}\), we consider the following optimization problem:
\[
\min_{h \in \mathcal{H}} \max_{k \in \mathcal{K}} \widehat{\gamma}_k^2\bigl(h(\mathbf{X}), \mathbf{Y}\bigr),
\]
where \(\mathbf{X} := (X_1, \dots, X_m)^T\) is the data matrix collected from a noisy random vector, and \(\mathbf{Y}\) is the observed data matrix drawn from the distribution we aim to learn.
In the aforementioned works, i.e., \cite{li2015generative, li2017mmd, briol2019statistical, cherief2022finite}, the generating function class $\mathcal{H}$ is considered as a set of neural networks.
We will show in Section \ref{sec::mmdgan} that our main result can be utilized to establish estimation error bounds in the resulting model, which ought to be new results in the literature. 

\section{Existing concentration inequalities and their limitation}
\label{sec:preliminary}
In this section, we review three pivotal concentration inequality results that form the starting points of our analysis. 
We will show that these existing inequalities are not sufficient to develop performance guarantee bounds for the methods that were mentioned in the previous section. 
In Section \ref{sec::mcdiarmid}, we review McDiarmid's inequality, which bounds the deviation between the sampled and expected values of functions evaluated on independent random variables, under the bounded differences condition. 
This condition ensures that small changes in individual variables have a limited effect on the function's value.
In Section \ref{sec::bartlett}, we discuss Bartlett's uniform concentration inequality for sample averages (\cite{bartlett2002rademacher}), which extends McDiarmid's inequality by ensuring that the concentration bound holds uniformly for the sample average of any function transformation of the input random variable within a specified function class.
This uniformity facilitates uniform convergence guarantees that are widely used in machine learning theory. 
In Section \ref{sec::maurer}, we examine Maurer's uniform concentration inequality for nonlinear functions (\cite{maurer2019uniform}), which builds upon Bartlett's result by extending the sample average to nonlinear functions, such as U- and V-statistics and Lipschitz statistics. 
Finally, in Section \ref{sec:need-new-inequality}, we discuss the need of new inequalities to establish performance evaluation for MMD-like statistics. 

\subsection{\texorpdfstring{\cite{mcdiarmid1998concentration}}{McDiarmid (1989)}}
\label{sec::mcdiarmid}
McDiarmid's Inequality is a fundamental concentration inequality that applies to functions of independent random variables with bounded differences.
\begin{lemma}[McDiarmid's Inequality; \texorpdfstring{\cite[Theorem 3.1]{mcdiarmid1998concentration}}{McDiarmid (1989)}]
\label{lem:mcdiarmid}
    Let $f: \mathcal{X}^n \rightarrow \mathbb{R}$ be a function such that for all $i \in\{1, \ldots, n\}$, there exist $c_i<\infty$ for which
    $$
        \sup _{x_1,\dots,x_n, \tilde{x} \in \mathcal{X}}\left|f\left(x_1, \ldots, x_{i-1}, x_i, x_{i+1}, \ldots, x_n\right)-f\left(x_1, \ldots x_{i-1}, \tilde{x}, x_{i+1}, \ldots, x_n\right)\right| \leq c_i .
    $$
    Then, for all random variable $X$ embedded in set $\mathcal{X}$, let $\mathbf{X}:=(X_1, \dots, X_n)^T$, where $X_i \stackrel{\text{i.i.d.}}{\sim} X$, $\forall i$. 
    We have, $\forall \delta \in (0, 1)$, 
    $$
        \operatorname{Pr}\left(\left|f(\mathbf{X})-\mathbb{E}[f(\mathbf{X})]\right| \leq \sqrt{\frac{\sum_{i=1}^n c_i^2 \log(2/\delta)}{2}}\right) \geq 1-\delta,
    $$
    where $f(\mathbf{X}) = (f(X_1), \dots, f(X_n))$.
\end{lemma}

\subsection{\texorpdfstring{\cite{bartlett2002rademacher}}{[Bartlett and Mendelson (2002)]}}
\label{sec::bartlett}
\cite{bartlett2002rademacher} extends concentration results to hold uniformly over a class of functions, particularly focusing on empirical averages versus expectations. 
This is essential for establishing uniform convergence, a cornerstone in statistical learning theory.

To begin with, given an Euclidean set $\mathcal{T} \subseteq \mathbb{R}^n$, we introduce the following complexity measure:
\begin{definition}[Rademacher Complexity]
    \label{def:Rademacher}
    The Rademacher complexity of a set $\mathcal{T} \subseteq \mathbb{R}^n$ is defined as follows:
    \begin{equation}
    \label{eq:Rademacher}
        \mathcal{R}(\mathcal{T}) := \mathbb{E}_\rho\left[\sup_{\mathbf{t} \in \mathcal{T}}\sum_{i=1}^n \rho_i t_i\right],
    \end{equation}
    where $\mathbf{t} = (t_1, \dots, t_n)^T$, $\rho_1, \dots, \rho_n$ are independent uniform $\{ \pm 1\}$-valued random variables.
\end{definition}

Utilizing the above complexity measure, Bartlett's concentration result is stated as follows:
\begin{lemma}[\texorpdfstring{\cite[Theorem 8]{bartlett2002rademacher}}]
\label{thm:bartlett}
    Given a random variable $X \in \mathcal{X} \subseteq \mathcal{U}$, let $\mathbf{X}:=(X_1, \dots, X_n)^T$, where $X_i \stackrel{\text{i.i.d.}}{\sim} X$, $\forall i$.
    $\mathbf{X}'$ be the independent copy of $\mathbf{X}$.
    Let $\mathcal{H}$ be a function set containing functions mapping from $\mathcal{X}$ to $[0, 1]$, $\mathcal{H}(\mathbf{X})\subseteq \mathbb{R}^{n}$ be an Euclidean set defined as follows:
    \begin{equation*}
        \mathcal{H}(\mathbf{X}) := \left\{h(\mathbf{X}) = (h(X_1), \dots, h(X_n))^T \mid h \in \mathcal{H}\right\}.
    \end{equation*}
    Let $\mathcal{R}(\mathcal{H}(\mathbf{X}))$ be the Rademacher complexity of set $\mathcal{H}(\mathbf{X})$, we have
    \begin{align*}
        \mathbb{E}\sup_{h \in \mathcal{H}}\left[ \mathbb{E}_{\mathbf{X}'}\left[\frac{1}{n}\sum_{i=1}^n h\left(X_i'\right)\right]-\frac{1}{n}\sum_{i=1}^n h(X_i)\right] \leq \frac{2}{n} \mathbb{E}[\mathcal{R}(\mathcal{H}(\mathbf{X}))].
    \end{align*}
    Based on McDiarmid's inequality (Lemma \ref{lem:mcdiarmid}), for any $\delta \in (0, 1)$, with probability at least $1-\delta$, we have
    \begin{align}
    \label{eq:bartlett}
        \sup_{h \in \mathcal{H}}\left[ \mathbb{E}_{\mathbf{X}'}\left[\frac{1}{n}\sum_{i=1}^nh\left(X_i'\right)\right]-\frac{1}{n}\sum_{i=1}^n h(X_i)\right] \leq \frac{2}{n} \mathbb{E}[R(\mathcal{H}(\mathbf{X}))] + \sqrt{\frac{\ln(1/\delta)}{2n}}.
    \end{align}
\end{lemma}

\subsection{\texorpdfstring{\cite{maurer2019uniform}}{[Maurer and Pontil (2019)]}}
\label{sec::maurer}
\cite{maurer2019uniform} generalizes concentration results to nonlinear functions of independent random variables. 
This extension is pivotal for analyzing more complex dependencies and interactions within function classes.

More specifically, given a nonlinear function $f: \mathcal{U}^n \mapsto \mathbb{R}$, where $\mathcal{U} \subseteq \mathbb{R}^d$,
Maurer's inequality holds under the following regularity constraints:
\begin{assumption}
\label{assum:maurer}
    Given a function $f: \mathcal{U}^n \to \mathbb{R}$, where $\mathcal{U} \subseteq \mathbb{R}^d$.
    Given seminorms $M_\text{Lip}(f)$, $J_\text{Lip}(f)$, and $M(f)$, we assume that $M_\text{Lip}(f) < \infty$, $J_\text{Lip}(f) < \infty$, and $M(f) < \infty$.
    The technical definitions of the seminorms are given as follows:
    \begin{equation}
    \label{eq:maurer_seminorms}
    \begin{aligned}
        & M_{\operatorname{Lip}}(f) := n \max _k \sup _{\mathbf{u} \in \mathcal{U}^n, y \neq y^{\prime} \in \mathcal{U}} \frac{D_{y y^{\prime}}^k f(\mathbf{u})}{\left\|y-y^{\prime}\right\|}, \\
        & J_{\operatorname{Lip}}(f) := n^2 \max _{k \neq l} \sup _{\mathbf{u} \in \mathcal{U}^n, y \neq y^{\prime}, z, z^{\prime} \in \mathcal{U}} \frac{D_{z z^{\prime}}^l D_{y y^{\prime}}^k f(\mathbf{u})}{\left\|y-y^{\prime}\right\|} ,\\
        & M(f) := n \max _k \sup _{\mathbf{u} \in \mathcal{U}^n, y,y^{\prime} \in \mathcal{U}} D_{y y^{\prime}}^k f(\mathbf{u}),
    \end{aligned}
    \end{equation}
    where $\|\cdot\|$ refers to Euclidean norm, $f(\mathbf{u}) = (f(u_1), \dots, f(u_n))$, $\forall \mathbf{u} \in \mathcal{U}^n$, and $D_{y y^{\prime}}^k f(\mathbf{u})$ is the $k$-th partial difference operator defined as follows:
    \begin{equation}
        \label{eq:partial}
        \begin{aligned}
        D_{y y^{\prime}}^k f(\mathbf{u}) := f
        (\dots, u_{k-1}, y, u_{k+1}, \ldots)-f(\ldots, u_{k-1}, y^{\prime}, u_{k+1}, \ldots), \quad k=1,\dots,n.
        \end{aligned}
    \end{equation}
\end{assumption}
Instead of the Rademacher complexity as given in Definition \ref{def:Rademacher}, the inequality established in \cite{maurer2019uniform} relies on Gaussian complexity, which replaces the Rademacher random variables in Definition \ref{def:Rademacher} with standard normal random variables.
\begin{definition}[Gaussian Complexity]
\label{def:Gauss}
    The Gaussian complexity of a set $\mathcal{T} \subseteq \mathbb{R}^n$ is defined as follows:
    \begin{equation}
    \label{eq:Gaussian}
        \mathcal{G}(\mathcal{T}) := \mathbb{E}_\xi \sup_{\mathbf{t} \in \mathcal{T}}\left[\sum_{i=1}^n \xi_i t_i\right],
    \end{equation}
    where $\mathbf{t} = (t_1, \dots, t_n)^T$, $\xi_i \stackrel{\text{i.i.d.}}{\sim} \mathcal{N}(0, 1)$, $\forall i$.
\end{definition}

It is worth noticing that the Gaussian complexity is closely related to the widely applied Rademacher complexity (Definition \ref{def:Rademacher}).
That is, for a wide range of function classes where its Rademacher complexities are investigated, the Gaussian complexity can be upper bounded via the following proposition.

\begin{proposition}[\protect{Gaussian and Rademacher Complexity}] 
\cite[Exercise 5.5]{wainwright2019high}. 
\label{def:complexity}
    Let $\mathcal{T}$ be a set in $\mathbb{R}^n$, $\mathcal{R}(\mathcal{T})$ and $\mathcal{G}(\mathcal{T})$ be the Rademacher and Gaussian complexities of set $\mathcal{T}$ as defined in \eqref{eq:Rademacher} and \eqref{eq:Gaussian}. We have
    \begin{align*}
        \sqrt{\frac{2}{\pi}}\mathcal{R}(\mathcal{T}) \leq \mathcal{G}(\mathcal{T}) \leq 2\sqrt{\log n} \mathcal{R}(\mathcal{T}).
    \end{align*}
\end{proposition}

Under Assumption \ref{assum:maurer}, utilizing the Gaussian complexity measure mentioned in Definition \ref{def:Gauss}, the uniform concentration inequality proposed in \cite{maurer2019uniform} is stated as follows.

\begin{lemma}[\texorpdfstring{\cite{maurer2019uniform}}{[Maurer and Pontil (2019)]}, Theorem 2, Corollary 3]
\label{lem:maurer}
    Given a nonlinear function $f: \mathcal{U}^n \mapsto \mathbb{R}$, where $\mathcal{U} \subseteq \mathbb{R}^{nd}$.
    Let $\mathbf{X}=\left(X_1, \ldots, X_n\right)$ be a vector of independent random variables taking values in set $\mathcal{X}, \mathbf{X}^{\prime}$ i.i.d. to $\mathbf{X}$. 
    Let $\mathcal{H}$ be a finite function set mapping from set $\mathcal{X}$ to $\mathcal{U}$, $\mathcal{H}(\mathbf{X})\subseteq \mathbb{R}^{nd}$ be a set defined as follows:
    \begin{equation*}
        \mathcal{H}(\mathbf{X}) := \left\{h(\mathbf{X}) = (h(X_1), \dots, h(X_n))^T \mid h \in \mathcal{H}\right\}.
    \end{equation*}
    Let $\mathcal{G}(\mathcal{H}(\mathbf{X}))$ be the Gaussian complexity of set $\mathcal{H}(\mathbf{X})$ (Definition \ref{def:Gauss}), then
    \begin{align}
    \label{eq:maurer_E}
        \mathbb{E}\sup_{h \in \mathcal{H}} \left[\mathbb{E}_{\mathbf{X}^{\prime}}\left[f\left(h\left(\mathbf{X}^{\prime}\right)\right)\right]-f(h(\mathbf{X}))\right] \leq \frac{\sqrt{2 \pi}}{n}\left(2 M_{\text {Lip}}(f)+J_{\text {Lip}}(f)\right) \mathbb{E}[\mathcal{G}(\mathcal{H}(\mathbf{X}))].
    \end{align}
    Moreover, based on McDiarmid's inequality (Lemma \ref{lem:mcdiarmid}), for any $\delta \in(0,1)$, with probability at least $1-\delta$, we have
    \begin{equation}
    \label{eq:maurer_prop}
    \begin{aligned}
        & \sup_{h \in \mathcal{H}}\bigg[ \mathbb{E}_{\mathbf{X}'}\left[f\left(h\left(\mathbf{X}^{\prime}\right)\right)\right]-f(h(\mathbf{X}))\bigg] \\
         \leq & \frac{\sqrt{2 \pi}}{n}\left(2 M_{\text {Lip}}(f)+J_{\text{Lip}}(f)\right) \mathbb{E}[\mathcal{G}(\mathcal{H}(\mathbf{X}))] + M(f) \sqrt{\frac{\ln (1 / \delta)}{n}}.
    \end{aligned}
    \end{equation}
\end{lemma}
Notably, the cardinality of the involved function set $\mathcal{H}$ is required to be finite to avoid problems of measurability.
Nevertheless, the cardinality of $\mathcal{H}$ can be arbitrarily large since it was not present in the upper bound of Lemma \ref{lem:maurer}.

\subsection{Need of new inequality for MMD-based statistics}
\label{sec:need-new-inequality}

The following describes why the aforementioned existing inequalities are not sufficient to provide inequality to study the performance of MMD-like statistics.
\begin{enumerate}
\item {Inapplicability of Classical Deviation Bounds.} 
Classical non-asymptotic deviation bounds based on \cite{hoeffding1963probability} and those that were reviewed in Section \ref{sec::mcdiarmid} through \ref{sec::maurer}  cannot be directly applied. 
Take MMD as a case study of kernel-based statistics. 
Classical bounds consider the probability bound for the difference $|\widehat{\gamma}_k^2 - \gamma_k^2|$.
However, the consistency analysis of the minimum MMD estimator (\cite{briol2019statistical}) requires bounding the expression $|\min_{h \in \mathcal{H}}\widehat{\gamma}_k^2(h(\mathbf{X}), \mathbf{Y}) - \min_{h \in \mathcal{H}}\gamma_k^2(h(X), Y)|$, which accounts for the difference between the empirical minimizer and the expected minimizer.
Furthermore, in the metrizing fairness example (Section \ref{sec:fairness-2}), MMD serves as a penalty term in the objective function. 
Consequently, the empirical optimizer $g^*$ is neither a minimizer nor a maximizer of the corresponding empirical estimator, i.e., $\widehat{\gamma}_k^2$. 
Therefore, probability bounds specified solely for an empirical minimizer or maximizer are insufficient.
In other words, a uniform concentration inequality for kernel-based statistics is necessary to provide estimation error bounds for the kernel-based statistics related optimization problems.

\item {Dependence in Empirical Estimators.} 
The empirical estimators of these kernel-based statistics (MMD, HSIC, ED, and dCov) involve summations of dependent terms. 
This dependence precludes the direct application of uniform concentration inequalities established for sample averages, such as those by \cite{bartlett2002rademacher} and \cite{koltchinskii2002empirical} (Lemma \ref{thm:bartlett}).
More specifically, compare the formulation of sample average as mentioned in Theorem \ref{thm:bartlett} with the empirical estimators of MMD (Definition \ref{def:empiricalMMDestimators}), it can be observed that the difference between their summation forms needs to be addressed.

\item {Two-Sample Statistic Complexity.} 
MMD, HSIC, ED, and dCov are inherently two-sample statistics. 
To establish a uniform concentration inequality that accounts for the employed functions, we consider the set $\mathcal{H}(\mathbf{X}, \mathbf{Y})$ in $\mathcal{U}^{m+n}$, which is defined as follows:
$$
    \left\{(h_\mathcal{X}(\mathbf{X}), h_\mathcal{Y}(\mathbf{Y})) = \left(h_\mathcal{X}(X_1), \dots, h_\mathcal{X}(X_m), h_\mathcal{Y}(Y_1), \dots, h_\mathcal{Y}(Y_n)\right) \mid h \in \mathcal{H}\right\},
$$
where $h(x, y) := (h_\mathcal{X}(x), h_\mathcal{Y}(y))$, $\forall x \in \mathcal{X}$, $y \in \mathcal{Y}$, $\forall h \in \mathcal{H}$.
Note that the above set \(\mathcal{H}(\mathbf{X}, \mathbf{Y})\) accounts for both the difference in sample sizes and the choice of functions for the involved pair of random vectors, i.e., $(X, Y)$.
Consequently, it can be concluded that it is not straightforward to adapt \eqref{eq:bartlett} to the two-sample statistic case, where the involved set $\mathcal{H}(\mathbf{X})$ is replaced by \(\mathcal{H}(\mathbf{X}, \mathbf{Y})\).
Even if \cite{maurer2019uniform} extended \eqref{eq:bartlett} to nonlinear cases (Lemma \ref{lem:maurer}), a similar problem occurred.
\end{enumerate}

\section{Main result: A uniform concentration inequality for two-sample statistics}
\label{sec:general}
Building on Maurer's uniform concentration inequality for nonlinear functions (Lemma \ref{lem:maurer}), we present our primary contribution: an extension of this inequality to the framework of two-sample statistics. 
While Maurer's result provides robust concentration bounds for the nonlinear functions of a single set of independent random variables, many practical statistical methods, such as hypothesis testing, fairness representation learning, and generative models, involve comparing two independent samples. 
Our extended inequality accommodates such two-sample scenarios by deriving concentration bounds that account for the interactions between two distinct datasets. 
In this section, we detail the technical formulation for our main result (in Theorem \ref{thm:general}) and its underlying assumptions.

We present our main result in Section \ref{sec:main-result}. 
In Section \ref{sec:cover-number}, the condition of finite covering numbers is discussed. 
The Gaussian complexities that occurred in the upper bound of Theorem \ref{thm:general} are discussed in Section \ref{sec::gaussianexample}. 

\subsection{Main result}\label{sec:main-result}
To begin, we formally define a two-sample statistic. Let \( X \) and \( Y \) be random vectors embedded in set $\mathcal{U} \subseteq \mathbb{R}^d$.
Unlike the nonlinear function \( f: \mathcal{U}^n \to \mathbb{R} \) discussed in Section~\ref{sec::maurer}, we consider sample matrices \( \mathbf{X} = (X_1, \dots, X_m)^\top \in \mathcal{U}^m \) and \( \mathbf{Y} = (Y_1, \dots, Y_n)^\top \in \mathcal{U}^n \), where $X_i \stackrel{\text{i.i.d.}}{\sim} X$ and $Y_j \stackrel{\text{i.i.d.}}{\sim} Y$, $\forall i, j$.
We define the function $f$ as a general two-sample statistic that maps from $\mathcal{U}^{m+n}$ to $\mathbb{R}$, involving a pair of random vectors that may have different sample sizes. 

We then define the function transform. 
Note the above two-sample statistics are defined for transformed random variables. 
Let \( \mathcal{H} \) be a specified set of functions mapping from \( \mathcal{X} \times \mathcal{Y} \) to \( \mathcal{U} \times \mathcal{U} \). Specifically, for each \( h \in \mathcal{H} \), the function \( h \) is defined as
\[
h(X, Y) = \left( h_\mathcal{X}(X),\ h_\mathcal{Y}(Y) \right),
\]
where \( h_\mathcal{X}: \mathcal{X} \to \mathcal{U} \) and \( h_\mathcal{Y}: \mathcal{Y} \to \mathcal{U} \).
Recall that the cardinality of the involved function set $\mathcal{H}$ is required to be finite in Maurer's uniform concentration inequality (Lemma \ref{lem:maurer}) to avoid problems of measurability.

In the following, we relaxed the finite cardinality assumption to the finite covering number assumption.
The technical statement is detailed as follows:
\begin{assumption}
    \label{assum:covering}
    Let $\mathcal{H}$ be a function set mapping from $\mathcal{X}\times \mathcal{Y}$ to $\mathcal{U}\times \mathcal{U}$, where $\mathcal{U} \subseteq \mathbb{R}^d$. 
    $\forall \epsilon > 0$, suppose the covering number $N(\epsilon; \mathcal{H}, \|\cdot\|_\infty) < \infty$, where $\|\cdot\|_\infty$ is a function norm defined as follows:
    \begin{align*}
        \|h\|_{\infty} := \inf\left\{C \geq 0 \,\bigg| \sup_{(x, y) \in \mathcal{X}\times\mathcal{Y}}\left|h(x, y)\right|\leq C \right\}.
    \end{align*}
    The covering number $N(\epsilon; \mathcal{H}, \|\cdot\|)$ is defined as the minimum cardinality of a set $\mathcal{H}_\epsilon$, such that $\forall h \in \mathcal{H}$, there exists a function $h_\epsilon \in \mathcal{H}_\epsilon$, such that $\|h_\epsilon - h\| \leq \epsilon$.
\end{assumption}

Under Assumption \ref{assum:maurer} and \ref{assum:covering}, the technical statement for our main result is detailed as follows:

\begin{theorem}
\label{thm:general}
    Given a pair of random vectors $X$ and $Y$ embedded in sets $\mathcal{X}$ and $\mathcal{Y}$, respectively.
    Let $\mathbf{X}:=(X_1, \dots, X_m)^T$ and $\mathbf{Y}:=(Y_1, \dots, Y_n)^T$, where $X_i \stackrel{\text{i.i.d.}}{\sim} X$ and $Y_j \stackrel{\text{i.i.d.}}{\sim} Y$, $\forall i, j$,
    $\mathbf{X}'$ and $\mathbf{Y}'$ be independent copies of $\mathbf{X}$ and $\mathbf{Y}.$
    Let $\mathcal{H}$ be a function set equipped with finite covering number (Assumption \ref{assum:covering}), $\mathcal{H}(\mathbf{X}, \mathbf{Y})$ be an Euclidean set in $\mathbb{R}^{(m+n)d}$ stated as follows:
    \begin{align}
    \label{eq:set_twosample}
        \mathcal{H}(\mathbf{X}, \mathbf{Y}) := \left\{(h_\mathcal{X}(\mathbf{X}), h_\mathcal{Y}(\mathbf{Y})) = \left(h_\mathcal{X}(X_1), \dots, h_\mathcal{X}(X_m), h_\mathcal{Y}(Y_1), \dots, h_\mathcal{Y}(Y_n)\right) \mid h \in \mathcal{H}\right\}.
    \end{align}
    Let $f: \mathcal{U}^{m}\times\mathcal{U}^n \mapsto \mathbb{R}$ be a general two-sample statistic under Assumption \ref{assum:maurer}, $\mathcal{G}(\mathcal{H}(\mathbf{X}, \mathbf{Y}))$ be the Gaussian complexity (Definition \ref{def:Gauss}) of set $\mathcal{H}(\mathbf{X}, \mathbf{Y})$. 
    Then $\forall \delta \in (0,1)$, with probability at least $1-\delta$, we have
    \begin{equation*}
    \begin{aligned}
        \sup _{h \in \mathcal{H}} | f(h_\mathcal{X}(\mathbf{X}), h_\mathcal{Y}(\mathbf{Y})) & - \mathbb{E}_{\mathbf{X}', \mathbf{Y}'}f(h_\mathcal{X}(\mathbf{X}'), h_\mathcal{Y}(\mathbf{Y}'))| \\
        \leq & \frac{\sqrt{2 \pi}}{m+n}(2 M_{\text {Lip }}(f)+J_{\text {Lip }}(f))\mathbb{E}[\mathcal{G}(\mathcal{H}(\mathbf{X}, \mathbf{Y}))] + M(f)\sqrt{\frac{\ln(2/\delta)}{m+n}}.
\end{aligned}
\end{equation*}
\end{theorem}
A proof of the above theorem can be found in Section \ref{sec:proof-1}.

\subsection{Discussion: Finite covering numbers}\label{sec:cover-number}

For most function sets in the machine learning context, the corresponding covering numbers are finite, indicating that the assumptions made in our work are generally satisfied.
In the following, we review the covering numbers for two commonly used function classes as illustrative examples, including the linear function class (Lemma \ref{lem:zhang2002}) and feed-forward neural networks embedded in a compact Euclidean set (Lemma \ref{lem:cover_NN}).

\begin{lemma}[Linear Functions; \texorpdfstring{\cite[Theorem 4]{zhang2002covering}}{[Zhang (2002)]}]
\label{lem:zhang2002}
    Let $\mathcal{H}$ be a set of linear functions defined as follows:
    \begin{align*}
        \mathcal{H}:=\{h: \mathcal{X}^{n} \mapsto \mathbb{R}^n \mid h(\mathbf{x})= (w^T x_1,\dots, w^Tx_n)^T,
        x_i \in \mathcal{X}\subset \mathbb{R}^d, \forall i, w \in \mathbb{R}^d, \|w\|_2 \leq C\}.
    \end{align*}
    Suppose the input domain $\mathcal{X}$ is bounded, that is, there exists a constant $B > 0$, such that $\sup_{x \in \mathcal{X}}\|x\| \leq B$. Then $\forall \epsilon > 0$, we have
    $$
        \log _2 N(\epsilon; \mathcal{H}, \|\cdot\|_\infty) \leq \left(\frac{6BC}{\epsilon}\right)^2 \log _2[2\lceil 4 BC / \epsilon+2\rceil n+1],
    $$
    where $\lceil \cdot \rceil$ refers to the ceiling function that maps $x$ to the smallest integer greater than or equal to $x$.
\end{lemma}

\begin{lemma}[\protect{Shallow Neural Networks, \cite[Theorem 2]{shen2023complexity}}]
\label{lem:cover_NN}
Let $\mathcal{H}$ be a set of two-layer neural networks mapping from $\mathcal{X}$ to $\mathbb{R}$, i.e., $\mathcal{H}:= \{h(x;\theta) |\theta \in [-B, B]^\mathcal{S} \}$, where
    \begin{align*}
        \theta = \left(W^{(1)},b^{(1)},W^{(2)}, b^{(2)}\right),\, h(x;\theta) = W^{(2)}\sigma\left(W^{(1)}x + b^{(1)}\right)+b^{(2)},
    \end{align*}
where $\sigma$ is an activation function and $\mathcal{S}$ is the total number of parameters in the network. 
Let the vector $(d_0, d_1)$ represent the dimensions of the layers of the neural network $h(x;\theta)$ given in the above formula, that is, we have $W^{(1)}\in \mathbb{R}^{d_1\times d_0}$, $b^{(1)} \in \mathbb{R}^{d_1 \times 1}$, $W^{(2)} \in \mathbb{R}^{1 \times d_1}$, and $b^{(2)} \in \mathbb{R}$.
Suppose the radius of $\mathcal{X}$ is bounded by a constant $B_\mathcal{X} > 0$, and the activation function $\sigma$ is $l_\sigma$-Lipschitz. Then $\forall \epsilon > 0$, we have
\begin{align*}
    N(\mathcal{H}, \epsilon, \|\cdot\|_\infty) \leq \left(16 B^2\left(B_\mathcal{X}+1\right) \sqrt{d_0} d_1 / \epsilon\right)^{d_0 d_1+2 d_1+1} \times l_\sigma^{d_0 d_1+d_1} / d_{1}!.
\end{align*}
\end{lemma}

\subsection{Discussion: Computable Gaussian complexities}
\label{sec::gaussianexample}

An important property of the Gaussian complexity is that it can be estimated from a single sample $\left(X_1, \ldots, X_n\right)$ and a single realization of the Gaussian variables.
The following result follows from McDiarmid's inequality and the concentration inequality for the standard normal distribution. 
The proof is referred to Section \ref{sec:proof-2}.

\begin{proposition}[\protect{Concentration of Gaussian Complexity}]
\label{prop:Gaussian}
    Given a function class  $\mathcal{H}:=\{h: \mathcal{X} \mapsto \mathcal{U}\}$.
    Denote the set $\left\{h(x) \mid h \in \mathcal{H}, x \in \mathcal{X}\right\}$ in $\mathcal{U}$ as $\mathcal{H}(\mathcal{X})$.
    Let $\mathcal{H}(\mathbf{X})$ be a set in $\mathcal{U}^n$ defined as $\mathcal{H}(\mathbf{X}):=\left\{(h(X_1), \dots, h(X_n)) \mid h \in \mathcal{H}\right\}$.
    Let $D(\mathcal{H}(\mathcal{X})):=\sup_{u, u' \in \mathcal{H}(\mathcal{X})}\|u-u'\|$ be the Euclidean width of the set $\mathcal{H}(\mathcal{X})$, $\mathcal{G}(\mathcal{H}(\mathbf{X}))$ be the Gaussian complexity of the set $\mathcal{H}(\mathbf{X})$ as defined in \eqref{eq:Gaussian}. 
    Then $\forall \delta \in (0,1)$, with probability at least $1-\delta$, we have
    \begin{align*}
        \left|\sup_{h \in \mathcal{H}} \sum_{i=1}^{n}\langle \xi_i, h(X_i) \rangle - \mathcal{G}(\mathcal{H}(\mathbf{X})) \right| \leq D(\mathcal{H}(\mathcal{X}))\sqrt{n\log \left(\frac{2}{\delta}\right)},
    \end{align*}
    where $\xi_i \stackrel{\text{i.i.d.}}{\sim} \mathcal{N}(\mathbf{0}_d, I_d)$, $\forall i$.
    Moreover, based on McDiarmid's inequality (Lemma \ref{lem:mcdiarmid}), we have
    \begin{align*}
        \left|\mathcal{G}(\mathcal{H}(\mathbf{X})) - \mathbb{E}\mathcal{G}(\mathcal{H}(\mathbf{X}))\right| \leq D(\mathcal{H}(\mathcal{X}))\sqrt{\frac{nd}{2} \log\left(\frac{2}{\delta}\right)}.
    \end{align*}
\end{proposition}
From the above proposition, it seems that the estimation of Gaussian complexity is particularly convenient. 
However, as mentioned before, the computation involves an optimization over the class $\mathcal{H}$, which can be complex for complicated function classes. 
Nonetheless, as investigated in \cite{bartlett2002rademacher}, suppose a function class is combined by several simple function classes, then its Gaussian complexity can be upper bounded by the combination of the Gaussian complexities of the simple function classes.
For instance, a decision tree can be expressed as a fixed boolean function of the functions appearing in each decision node, voting methods use thresholded convex combinations of functions from a simpler class, and neural networks are compositions of fixed squashing functions with linear combinations of functions from some class.

In the following, we first provide the contraction principle, which allows us to provide a bound for the Gaussian (or Rademacher) complexity of a composition of functions in terms of the complexity of the original function class and the Lipschitz constant of the outer function.
Note that it can be considered as a version of Talagrand's contraction lemma.
Since our definition of Gaussian complexity does not use absolute values, we provide an explicit proof in Section \ref{sec:proof-4.6}.

\begin{lemma}[Contraction Inequality for Gaussian Complexity]
\label{lem:contraction}
    Let $\mathcal{F}$ be a class of functions mapping from a set $\mathcal{X}$ to $\mathbb{R}^{d_1}$, where $\mathcal{X}$ is an Euclidean set in $\mathbb{R}^{d_0}$.
    Let $X$ be a random vector embedded in set $\mathcal{X}$, $\mathbf{X}:=(X_1, \dots, X_n)^T$, $X_i \stackrel{\text{i.i.d.}}{\sim} X$, $\forall i$. 
    Let $\phi$ be an element-wise non-decreasing activation function with Lipschitz constant $l_\phi$. 
    Then:
    $$
        \mathcal{G}(\phi \circ \mathcal{F}(\mathbf{X})) \leq l_\phi \cdot \mathcal{G}(\mathcal{F}(\mathbf{X})),
    $$
    where $\phi \circ \mathcal{F}=\{\phi \circ f: f \in \mathcal{F}\}$, and $\mathcal{G}(\cdot)$ denotes the Gaussian complexity.
\end{lemma}

For a set of linear functions under certain regularity constraints, its Gaussian complexity is provided as follows, the proof is referred to Section \ref{sec:proof-4.7}.
\begin{proposition}[Linear Functions]
\label{prop:linear}
   For $x \in \mathbb{R}^{d}$, define
    $$
        \mathcal{F}=\left\{x \mapsto \boldsymbol{w}^T x: \boldsymbol{w} \in \mathbb{R}^{d},\|\boldsymbol{w}\|_{1} \leq \omega\right\},
    $$
    where $\|\boldsymbol{w}\|_1 = \sum_{i=1}^{d} |w_i|$.
    For any data matrix $\mathbf{X} := (X_1, \ldots, X_n)^T \in \mathbb{R}^{nd_0}$, we have
    $$
        \mathcal{G}\left(\mathcal{F}(\mathbf{X})\right) \leq \omega \sqrt{2\log(2 d)} \max_k \sqrt{\sum_{i=1}^n X_{i, k}^2}.
    $$
\end{proposition}

Based on the contraction inequality (Lemma \ref{lem:contraction}) and the Gaussian complexity for a linear function class, the Gaussian complexity for the regularized feed-forward neural networks is stated as follows, the proof is referred to Section \ref{sec:proof-prop4.8}. 
\begin{proposition}[\protect{Feed-Forward Neural Networks}] 
\cite[Theorem 18]{bartlett2002rademacher} 
\label{prop:fnn}
    Consider a feed-forward neural network with depth $\iota$, which is given by the function $f^{\iota}_{nn}: \mathbb{R}^{d} \mapsto \mathbb{R}$ defined as follows
    \begin{eqnarray*}
        f_{n n}^{(\iota)}(x):=l^{(\iota)} \circ \cdots \circ l^{(1)}(x) \equiv l^{(\iota)}\left(\cdots l^{(2)}\left(l^{(1)}(x)\right) \cdots\right),
    \end{eqnarray*}
    where $d_0 = d$, $d_\iota = 1$, and $l^{(\iota)} := W^{(\iota)}x$ for a specified matrix $\mathbb{R}^{1 \times d_{\iota - 1}}$.
    Here, for $k=1, \dots, \iota - 1$, $l^{(k)}: \mathbb{R}^{d_{k-1}} \mapsto \mathbb{R}^{d_{k}}$ is the $k$-th hidden layer consists of a coordinate-wise composition of an activation function $\sigma: \mathbb{R} \mapsto \mathbb{R}$ and an affine map, namely, $l^{(k)}(x):=\phi(W^{(k)}x)$ for an given interaction matrix $W^{(k)} \in \mathbb{R}^{d_k \times d_{k-1}}$.
    Let the interaction matrices be the parameters to be tuned, the corresponding class of neural networks is given as follows:
    \begin{align*}
        \mathcal{F}:=\left\{f_{n n}^{(\iota)}(x) \,\bigg| \left\|W^{(k)}\right\|_{1, \infty} \leq \omega, \forall k\right\},
    \end{align*}
    where for a given matrix $W$, the $\|\cdot\|_{1,\infty}$ norm is defined as $\|W\|_{1, \infty} = \max_{i}\sum_j|W_{i,j}|$.
    Suppose the activation function $\sigma$ is $\lambda$-Lipschitz, let $\mathbf{X}:=(X_1, \dots, X_n)^T \in \mathbb{R}^{n\times d_0}$, we have 
    $$
        \mathcal{G}(\mathcal{F}(\mathbf{X})) \leq (2\omega)^{\iota }\lambda^{\iota - 1} \sqrt{2 \log (2 d)} \max_k\sqrt{\sum_{i=1}^n X_{i, k}^2}.
    $$
\end{proposition}

Note that the Gaussian complexities in the above examples are equipped with order $O(n^{1/2})$. 
In that case, the upper bound in Therorem \ref{thm:general} can be reformulated in the following formula:
\begin{align*}
    \frac{C_1}{\sqrt{m+n}} + C_2\sqrt{\frac{\log(1/\delta)}{m+n}},
\end{align*}
where $C_1$ and $C_2$ are constants irrelavent with the sample size $n$.

Readers can refer to \cite{bartlett2002rademacher, meir2003generalization, ambroladze2007complexity, biau2008performance, kakade2008complexity, kakade2012regularization} for more techniques to bound Rademacher and Gaussian averages in various contexts.

Recall the Gaussian complexity $\mathcal{G}(\mathcal{H}(\mathbf{X}, \mathbf{Y}))$ incorporated in our general uniform concentration inequality for two-sample statistics (Theorem \ref{thm:general}).
Here, for any function $h$ in the function class $\mathcal{H} = \{(h_\mathcal{X}, h_\mathcal{Y}) \mid h_\mathcal{X}: \mathcal{X} \to \mathcal{U}, h_\mathcal{Y}: \mathcal{Y} \to \mathcal{U}\}$, it is composed of functions $h_\mathcal{X}$ and $h_\mathcal{Y}$.
To leverage existing upper bounds for specific function sets in machine learning, we employ the following decomposition, derived directly from the definition of Gaussian complexity (Definition \ref{def:Gauss}):
\begin{lemma}
    Given the function class $\mathcal{H}=\{(h_\mathcal{X}, h_{\mathcal{Y}}) \mid 
    h_\mathcal{X}: \mathcal{X} \mapsto \mathcal{U}, 
    h_\mathcal{Y}: \mathcal{Y} \mapsto \mathcal{U}\}$. 
    Define $\mathcal{H}_{\mathcal{X}}:= \{h_\mathcal{X} \mid (h_\mathcal{X}, h_\mathcal{Y}) \in \mathcal{H}\}$ and $\mathcal{H}_{\mathcal{Y}}:= \{h_\mathcal{Y} \mid (h_\mathcal{X}, h_\mathcal{Y}) \in \mathcal{H}\}$. 
    Based on the definition of Gaussian complexity \eqref{eq:Gaussian}, we have
    \begin{align*}
        \mathbb{E}\mathcal{G}(\mathcal{H}(\mathbf{X}, \mathbf{Y})) \leq \mathbb{E}\mathcal{G}(\mathcal{H}_{\mathcal{X}}(\mathbf{X})) + \mathbb{E}\mathcal{G}(\mathcal{H}_{\mathcal{Y}}(\mathbf{Y})).
    \end{align*}
\end{lemma}
That is, to upper bound $\mathcal{G}(\mathcal{H}(\mathbf{X}, \mathbf{Y}))$, one can simply utilize upper bounds for $\mathcal{G}(\mathcal{H}_\mathcal{X}(\mathbf{X}))$ and $\mathcal{G}(\mathcal{H}_\mathcal{Y}(\mathbf{X}))$ and then call the inequality presented in the above lemma.

\section{Application of our main result to kernel-based statistics}
\label{sec:kernel-based}

In this section, we show how our main result in Theorem \ref{thm:general} can be applied to a range of kernel-based statistics, including maximum mean discrepancy (Section \ref{sec:mmd-2}), 
Hilbert-Schmidt independence criterion (Section \ref{sec::HSIC}), and 
distance covariance (Section \ref{sec::dbstats}). 
In Section \ref{sec::verification},  we show that the assumptions (Assumption \ref{assum:BandL}) required for the reproducing kernel function $k$ is satisfied for several commonly used reproducing kernel functions.

\subsection{Maximum mean discrepancy}
\label{sec:mmd-2}

Recall the definition of maximum mean discrepancy and its empirical versions that were defined in Section \ref{sec:define-MMD}. 
To implement our general uniform concentration inequality (Theorem \ref{thm:general}) to MMD, we investigate the assumptions under which the seminorms $M_\text{Lip}$, $J_\text{Lip}$, and $M$ (Assumption \ref{assum:maurer}) of empirical MMD estimators are finite.
More specifically, given a reproducing kernel function $k: \mathcal{U} \times \mathcal{U} \mapsto \mathbb{R}$, the technical statement for the assumption is stated as follows:

\begin{assumption}
\label{assum:BandL}
Given a reproducing kernel Hilbert space $\mathcal{F}$ consisting of functions $f: \mathcal{U} \mapsto \mathbb{R}$, where $\mathcal{U}\subset \mathbb{R}^d$.
Let $k: \mathcal{U} \times \mathcal{U} \mapsto \mathbb{R}$ be the reproducing kernel function of space $\mathcal{F}$, that is, $\forall f \in \mathcal{F}$, we have $f(x) = \langle f, k(x, \cdot)\rangle_{\mathcal{F}}$, where $\langle \cdot, \cdot \rangle_{\mathcal{F}}$ is the inner product defined in space $\mathcal{F}$.
Suppose the reproducing kernel $k$ is built under the following regularity conditions:
    \begin{enumerate}
        \item (\text{Bounded}). $\exists \nu \in \mathbb{R}$, such that we have $\sup_{u \in \mathcal{U}}k(u,u) \leq \nu$.
        \item (\text{Lipschitz}). $\exists l \in \mathbb{R}$, such that we have $M_{\text{Lip}}(k) \leq l$.
    \end{enumerate}
\end{assumption}

\begin{corollary}[Maximum Mean Discrepancy]
\label{coro:MMD}
    Given matrices $\mathbf{X}:=(X_1,\dots, X_m)^T$ and $\mathbf{Y}:=(Y_1,\dots,Y_n)^T$, where $X_i \stackrel{\text{i.i.d.}}{\sim} P$, and $Y_j \stackrel{\text{i.i.d.}}{\sim} Q$, $\forall i, j$. Denote $m/(m+n)$ as $\rho_x$, and $n/(m+n)$ as $\rho_y$, where $\rho_x+\rho_y = 1$. Adopt the constants proposed in Assumption \ref{assum:BandL}, regarding the unbiased estimate $\widehat{\gamma}_{k, u}$, \eqref{eq:intro_unbiased_MMD}, we have
    \begin{align*}
        &M_{\text{Lip}}(\widehat{\gamma}_{k, u}^2) \leq 2l\max \left\{\rho_x^{-1}, \rho_y^{-1}\right\},\\
        &J_{\text{Lip}}(\widehat{\gamma}_{k, u}^2) \leq 4l\max\left\{\rho_x^{-2}, \rho_y^{-2}\right\}, \text{ and}\\
        &M\left(\widehat{\gamma}_{k, u}^2\right) \leq 8\nu \max \left\{\rho_x^{-1}, \rho_y^{-1}\right\}.
    \end{align*}
    Moreover, regarding the biased estimate $\widehat{\gamma}_{k,b}$ \eqref{eq:intro_biased_MMD}, we have 
    \begin{align*}
        \sup_{(\mathbf{x}, \mathbf{y})\in \mathcal{U}^{m+n}}\left|\widehat{\gamma}_{k,u}^2(\mathbf{x}, \mathbf{y}) - \widehat{\gamma}_{k,b}^2(\mathbf{x}, \mathbf{y})\right| \leq 2\nu \frac{\rho_x^{-1}\rho_y^{-1}}{m+n}.
    \end{align*}
    Suppose $m = n$, regarding the U-statistic estimate $\widehat{\gamma}_{k, e}$, \eqref{eq:intro_U_MMD}, we have
    \begin{align*}
        \sup_{(\mathbf{x}, \mathbf{y})\in \mathcal{U}^{2n}}\left|\widehat{\gamma}_{k,u}^2(\mathbf{x}, \mathbf{y}) - \widehat{\gamma}_{k,e}^2(\mathbf{x}, \mathbf{y})\right| \leq \frac{4\nu}{n}.
    \end{align*}
    In conclusion, suppose $\mathcal{H}:=\{(h_\mathcal{X}, h_\mathcal{Y}) \mid h_\mathcal{X}: \mathcal{X} \mapsto \mathcal{U}, h_\mathcal{Y}: \mathcal{Y} \mapsto \mathcal{U}\}$ is a function class equipped with finite covering number, i.e., $\forall \epsilon > 0$, $N(\epsilon; \mathcal{H}, \|\cdot\|_\infty) < \infty$ (Assumption \ref{assum:covering}), let $\mathcal{G}(\mathcal{H}(\mathbf{X}, \mathbf{Y}))$ be the empirical Gaussian complexity defined in Equation \eqref{eq:Gaussian} regarding the set $\mathcal{H}(\mathbf{X}, \mathbf{Y})$ \eqref{eq:set_twosample}. 
    Then
    $\forall \delta \in (0,1)$, with probability at least $1-\delta$, we have either one of the following inequalities:
    \begin{align*}
    \sup_{(h_\mathcal{X}, h_\mathcal{Y}) \in \mathcal{H}} \bigg|\widehat{\gamma}_{k,u}^2(h_\mathcal{X}(\mathbf{X}), h_\mathcal{Y}(\mathbf{Y})) - \gamma_k^2(h_\mathcal{X}(X), h_\mathcal{Y}(Y))\bigg| \leq 8\nu \max\left\{\rho_x^{-1}, \rho_y^{-1}\right\}\sqrt{\frac{\log (2 / \delta)}{m+n}}\\
    + \frac{4\sqrt{2\pi}l}{m+n}\max\bigg\{\rho_y^{-1}
    \left(1 + \rho_y^{-1}\right), \rho_x^{-1}\left(1 + \rho_x^{-1}\right)\bigg\} \mathbb{E}\left[\mathcal{G}(\mathcal{H}(\mathbf{X}, \mathbf{Y}))\right];
    \end{align*}
    \begin{align*}
        \sup_{(h_\mathcal{X}, h_\mathcal{Y}) \in \mathcal{H}} \bigg|\widehat{\gamma}_{k,b}^2(h_\mathcal{X}(\mathbf{X}), h_\mathcal{Y}(\mathbf{Y})) - \gamma_k^2(h_\mathcal{X}(X), h_\mathcal{Y}(Y))\bigg| \leq 8\nu \max\left\{\rho_x^{-1}, \rho_y^{-1}\right\}\sqrt{\frac{\log (2 / \delta)}{m+n}}\\
        + 2\nu\frac{\rho_x^{-1}\rho_y^{-1}}{m+n} + \frac{4\sqrt{2\pi}l}{m+n}\max\bigg\{\rho_y^{-1}\left(1 + \rho_y^{-1}\right), \rho_x^{-1}\left(1 + \rho_x^{-1}\right)\bigg\} \mathbb{E}\left[\mathcal{G}(\mathcal{H}(\mathbf{X}, \mathbf{Y}))\right].
    \end{align*}
    In cases where $m = n$, $\forall \delta \in (0, 1)$, with probability at least $1-\delta$, we have
    \begin{align*}
        \sup_{(h_\mathcal{X}, h_\mathcal{Y}) \in \mathcal{H}} \bigg|\widehat{\gamma}_{k,e}^2(h_\mathcal{X}(\mathbf{X}), h_\mathcal{Y}(\mathbf{Y})) - \gamma_k^2(h_\mathcal{X}(X), h_\mathcal{Y}(Y))\bigg| \leq 8\sqrt{2}\nu \sqrt{\frac{\log (2 / \delta)}{n}}+ \frac{4\nu}{n}\\ 
        + \frac{12\sqrt{2\pi}l}{n}\mathbb{E}\left[\mathcal{G}(\mathcal{H}(\mathbf{X}, \mathbf{Y}))\right].
    \end{align*}
\end{corollary}
A proof of the above Corollary can be found in Section \ref{sec:proof-3}. 

To the best of our knowledge, the above inequality is the first uniform concentration bound for empirical MMD estimators over a specified set of functions.

Note that $\gamma_k^2(h_\mathcal{X}(X), h_\mathcal{Y}(Y))$ is the population quantity and the corresponding empirical estimate is $\widehat{\gamma}_{k,\cdot}^2(h_\mathcal{X}(\mathbf{X}), h_\mathcal{Y}(\mathbf{Y}))$. 
Therefore, the aforementioned inequalities give an upper bound of the estimation error. 
This can be utilized to generate performance guarantees when an MMD-estimator is adopted. 

\subsection{Hilbert-Schmidt independence criterion}
\label{sec::HSIC}
Given a pair of random vectors $X$ and $Y$ equipped with the joint distribution $P_{XY}$ and marginal distributions $P_X$, $P_Y$.
In \cite{gretton2005measuring, gretton2007kernel}, MMD is employed to measure the distance between probability metrics $P_{XY}$ and $P_X P_Y$, serving as a measure of statistical dependence, which is named as the \textit{Hilbert-Schmidt independence criterion} (HSIC).
The technical statement is given as follows.
\begin{definition}[HSIC] 
\cite[Definition 11]{sejdinovic2013equivalence}
\label{def:hsic}
    Let $X \sim P_X$ and $Y\sim P_Y$ be random variables on $\mathcal{X}$ and $\mathcal{Y}$, respectively, having joint distribution $P_{XY}$. 
    Furthermore, let $k_{\mathcal{X}}: \mathcal{X} \times \mathcal{X} \mapsto \mathbb{R}$ and $k_{\mathcal{Y}}: \mathcal{Y} \times \mathcal{Y} \mapsto \mathbb{R}$ be reproducing kernels on RKHSs $\mathcal{F}_{\mathcal{X}}$ and $\mathcal{F}_{\mathcal{Y}}$, respectively. 
    Let $\mathcal{F}$ be an RKHS isometrically isomorphic to the tensor product $\mathcal{F}_{\mathcal{X}} \otimes \mathcal{F}_{\mathcal{Y}}$.
    The following function can be considered as a reproducing kernel for space $\mathcal{F}$ \cite[Lemma 4.6]{steinwart2008support}:
    \begin{align}
    \label{eq:HSIC_kernel}
        k\left((x, y),\left(x^{\prime}, y^{\prime}\right)\right)=k_{\mathcal{X}}\left(x, x^{\prime}\right) k_{\mathcal{Y}}\left(y, y^{\prime}\right),
    \end{align}
    The {Hilbert-Schmidt Independence Criterion} (HSIC) of $X$ and $Y$ is the squared MMD, $\gamma_k^2$, between the joint distribution $P_{XY}$ and the product of its marginals $P_XP_Y$, that is, let $Z \sim P_XP_Y$, we have
    $$
    \begin{aligned}
        \gamma_k^2\left((X, Y), Z\right) =\left\|\mathbb{E}\left[k(\cdot, (X, Y))\right]-\mathbb{E} k_{\mathcal{X}}(\cdot, X) \otimes \mathbb{E} k_{\mathcal{Y}}(\cdot, Y)\right\|^2_\mathcal{F}.
    \end{aligned}
    $$
    Given sample matrices $\mathbf{Z}:=((X_1, Y_1)^T, \cdots, (X_n, Y_n)^T)^T$, $\mathbf{X} = (X_1,\dots, X_n)^T$, and $\mathbf{Y}:= (Y_1,\dots, Y_n)^T$, where $(X_i, Y_i) \stackrel{\text{i.i.d.}}{\sim} P_{XY}$, $\forall i$.
    There are two versions of the empirical estimates for the HSIC, i.e., $\gamma_k^2\left((X, Y), Z\right)$, which are derived based on the biased estimate $\widehat{\gamma}_{k, b}$ and the U-statistic estimate $\widehat{\gamma}_{k, e}$, respectively.
    More specifically, let $\mathbf{K}$ and $\mathbf{H}$ be the $n \times n$ kernel matrices for $\mathbf{X}$ and $\mathbf{Y}$, respectively, with entries $K_{i j}=k_{\mathcal{X}}\left(X_i, X_j\right)$ and $L_{i j}=k_{\mathcal{Y}}\left(Y_i, Y_j\right)$, $\forall i, j$. Let $\mathbf{H}=\mathbf{I}_n-\frac{1}{n} \mathbf{1}_n \mathbf{1}_n^{\top}$ be the centering matrix, we have
    \begin{align*}
        \widehat{\gamma}_{k, b}^2\left((\mathbf{X}, \mathbf{Y}), \mathbf{Z}\right) &:= \frac{1}{n^2} \operatorname{tr}(\mathbf{K H L H});\\
        \widehat{\gamma}_{k, e}^2\left((\mathbf{X}, \mathbf{Y}), \mathbf{Z}\right) &:= \frac{1}{n(n-1)} \sum_{i \neq j}\left[ K_{i j} L_{i j}+K_{i i} L_{j j}-K_{i  j} L_{j j}-K_{i i} L_{i j}\right].
    \end{align*}
\end{definition}

In the following, we show that as long as the margin reproducing kernel functions $k_{\mathcal{X}}$ and $k_{\mathcal{Y}}$ satisfy the conditions listed in Assumption \ref{assum:BandL}, the product kernel $k$ involved in the HSIC is naturally equipped with the required constraints.
That is, our inequality for MMD (Corollary \ref{coro:MMD}) can be extended to HSIC.
The proof is referred to Section \ref{sec:proof-5.4}.

\begin{corollary}[HSIC]
\label{coro:HSIC}
    Adopt notations in Assumption \ref{assum:BandL}. 
    Suppose the kernels $k_{\mathcal{X}}$ and $k_{\mathcal{Y}}$ are equipped with constants $(\nu_{\mathcal{X}}, l_{\mathcal{X}})$ and $(\nu_{\mathcal{Y}}, l_{\mathcal{Y}})$, respectively. Regarding the kernel $k$ given in Equation (\ref{eq:HSIC_kernel}), we have 
    \begin{enumerate}
        \item $\sup_{u \in \mathcal{U}} k(u,u) \leq \nu_{\mathcal{X}}\nu_{\mathcal{Y}}$;
        \item $M_{\text{Lip}}(k) \leq \sqrt{2}\max\{\nu_\mathcal{X}l_\mathcal{Y},\nu_\mathcal{Y}l_\mathcal{X}\}$.
    \end{enumerate}
    That is, by replacing the kernel function $k(\cdot, \cdot)$ in Corollary \ref{coro:MMD} with the product kernel function $k_{\mathcal{X}}\left(x, x^{\prime}\right) k_{\mathcal{Y}}\left(y, y^{\prime}\right)$, we derive uniform concentration inequalities for the corresponding empirical estimates of HSIC. In this case, the constants $\nu$ and $l$ in Corollary \ref{coro:MMD} are replaced by $\nu_{\mathcal{X}} \nu_{\mathcal{Y}}$ and $\sqrt{2} \max \left\{\nu_{\mathcal{X}} l_{\mathcal{Y}}, \nu_{\mathcal{Y}} l_{\mathcal{X}}\right\}$, respectively. More specifically, $\forall \delta \in (0, 1)$, with probability at least $1-\delta$, we have
    \begin{align*}
        \sup_{h \in \mathcal{H}}\Bigg|\widehat{\gamma}_{k}^2\left((h_\mathcal{X}(\mathbf{X}), h_\mathcal{Y}(\mathbf{Y})), h(\mathbf{Z})\right) - {\gamma}_{k}^2\left(h(X, Y), h(Z)\right)\Bigg| \leq& 8\sqrt{2}\nu_{\mathcal{X}} \nu_{\mathcal{Y}}\sqrt{\frac{\log(2/\delta)}{n}} +\\ 
         \frac{4\nu_{\mathcal{X}} \nu_{\mathcal{Y}}}{n} + 48\sqrt{\pi}\max &\left\{\nu_{\mathcal{X}} l_{\mathcal{Y}}, \nu_{\mathcal{Y}} l_{\mathcal{X}}\right\}\mathbb{E}[\mathcal{G}(\mathcal{H}(\mathbf{X}, \mathbf{Y}))],
    \end{align*}
    where $\widehat{\gamma}_{k}^2((h_\mathcal{X}(\mathbf{X}), h_\mathcal{Y}(\mathbf{Y})), h(\mathbf{Z}))$ can be considered as either the biased estimate $\widehat{\gamma}_{k, b}^2$ or the U-statistic estimate $\widehat{\gamma}_{k, u}^2$ for HSIC as mentioned in Definition \ref{def:hsic}.
\end{corollary}

To the best of our knowledge, the above corollary serves as the first uniform concentration inequality for the empirical estimators of HSIC regarding a specified function set.

\subsection{Distance-based statistics}
\label{sec::dbstats}
In \cite{sejdinovic2013equivalence}, the equivalence between RKHS-based statistics and the distance-based statistic is proved.
That is, given a distance-induced reproducing kernel, the energy distance is equivalent to MMD, and the distance covariance is equivalent to HSIC.
Consequently, our concentration result for MMD (Corollay \ref{coro:MMD}) and HSIC (Corollay \ref{coro:HSIC}) can be extended to the energy distance and distance covariance, respectively.
In the following, we present more technical details along this line of the literature.

Proposed in \cite{rizzo2016energy}, Energy Distance (ED) is a probability metric applied in numerous topics, including the hypothesis testing (\cite{rizzo2016energy}), Crámer GAN (\cite{bellemare2017cramer}), etc. 
The mathematical definition is given as follows.
\begin{definition}[Energy Distance, \cite{rizzo2016energy}]
\label{def:ED}
Given random vectors $X$ and $Y$ equipped with finite first moments. Let $X'$ and $Y'$ be independent copies of $X$ and $Y$, respectively. The Energy Distance (ED) between $X$ and $Y$ is defined as follows:
\begin{equation}
    D_E(X, Y) = 2\mathbb{E}_{XY}\|X-Y\|-\mathbb{E}_{XX'}\|X-X'\| - \mathbb{E}_{YY'}\|Y-Y'\|.
\end{equation}
\end{definition}
Similar to the extension from MMD to HSIC, in \cite{szekely2009brownian}, the energy distance is applied to measure the distance between a joint distribution and the product of marginal distributions, serving as a measure of statistical dependence, named as distance Covariance (dCov).
dCov has found applications in dimension reduction (\cite{sheng2013direction, sheng2016sufficient, chen2019sufficient, yu2019distance}), independent component analysis (\cite{matteson2017independent}), etc.
The technical expression for dCov is given as follows:

\begin{definition}[Distance Covariance, \protect{\cite{szekely2009brownian}}]
    Given Borel probability measures $P$ and $Q$ equipped with finite first moments. 
    Given random variables $X \sim P$ and $Y \sim Q$.
    The distance Covariance (dCov), $\mathcal{V}(X, Y)$, between $X$ and $Y$ is defined as follows:
    \begin{equation}
        \begin{aligned}
            \mathcal{V}^2(X, Y)=  \mathbb{E}_{XY} \mathbb{E}_{X^{\prime} Y^{\prime}}\left\|X-X^{\prime}\right\|_2\left\|Y-Y^{\prime}\right\|_2 &+\mathbb{E}_X \mathbb{E}_{X^{\prime}}\left\|X-X^{\prime}\right\|_2 \mathbb{E}_Y \mathbb{E}_{Y^{\prime}}\left\|Y-Y^{\prime}\right\|_2 \\
            -2&\mathbb{E}_{XY}\left[\mathbb{E}_{X^{\prime}}\left\|X-X^{\prime}\right\|_2 \mathbb{E}_{Y^{\prime}}\left\|Y-Y^{\prime}\right\|_2\right],
        \end{aligned}
    \end{equation}
    where $X, X' \stackrel{i.i.d.}{\sim} P$ and $Y, Y' \stackrel{i.i.d.}{\sim} Q$.
\end{definition}

In \cite{sejdinovic2013equivalence}, a positive definite kernel, termed distance-induced kernel, is defined, such that the MMD corresponds exactly to the ED.
This equivalence readily extends to dCov and HSIC using kernels on the product space.
The formulation for the distance-induced kernel and the proof for its constants, as mentioned in Assumption \ref{assum:BandL}, is stated as follows.
The proof is referred to Section \ref{sec:proof-5.7}.

\begin{lemma}[Distance-Induced Kernel]
\label{ex:dk}
Given a set $\mathcal{U} \subset \mathbb{R}^d$ and a fixed point $u_0 \in \mathcal{U}$, the \textit{distance-induced kernel} $k: \mathcal{U} \times \mathcal{U} \mapsto \mathbb{R}$ is given as follows:
\begin{equation}
    \label{eq:dkernel}
        k(u, u'):= \frac{1}{2}\left[\|u-u_0\| + \|u'-u_0\| - \|u-u'\| \right].
\end{equation}
Suppose there exists a positive constant $\nu_{\text{dis}}$, such that $\sup_{u \in\mathcal{U}} \|u-u_0\| \leq \nu_{\text{dis}}$, we have
\begin{enumerate}
    \item $\sup_{u \in \mathcal{U}} k(u,u) \leq \nu_{\text{dis}}$, that is, $\nu \leq \nu_{\text{dis}}$;
    \item $M_\text{Lip}(k) \leq 2$.
\end{enumerate}
\end{lemma}

Considering the RKHS-based statistics (MMD, HSIC) equipped with the above distance-induced kernel, their equivalence with the aforementioned distance-based statistics (ED, dCov) is detailed as follows:
\begin{lemma}[\cite{sejdinovic2013equivalence}, Theorem 22]
\label{lem:DE_MMD}
Adopt notations in Lemma \ref{ex:dk}, the equivalence between MMD and ED, as well as between dCov and HSIC, is summarized as follows:
\begin{enumerate}
    \item (\textbf{MMD and Energy Distance}) Let $P$ and $Q$ be two Borel probability measures embedded in set $\mathcal{U}$ with finite first moments, random vectors $X\sim P$, $Y\sim Q$, and $\mathcal{F}$ be an RKHS equipped with a distance-induced kernel $k$ given in Lemma \ref{ex:dk}.
    In this case, we have
    $$
        D_E(X, Y) = 2\gamma_k^2(X, Y).
    $$
    Moreover, the unbiased, biased, and U-statistic empirical estimate of $D_E(P, Q)$ can be obtained by multiplying the corresponding empirical estimates of MMD by $2$. That is, given data matrices $\mathbf{X} = (X_1,\dots, X_m)^T$ and $\mathbf{Y}=(Y_1,\dots, Y_n)^T$, where $X_i \stackrel{\text{i.i.d.}}{\sim} X$ and $Y_i \stackrel{\text{i.i.d.}}{\sim} Y$, we have
    \begin{align*}
        \widehat{D}_{E, u}(X, Y):=& \frac{\sum_{i \neq j}^m \|X_i-X_j\|}{m(m-1)} + \frac{\sum_{i\neq j}^n \|Y_i - Y_j\|}{n(n-1)} -\frac{2\sum_{i=1}^m\sum_{j=1}^n \|X_i - Y_j\|}{mn};\\
        \widehat{D}_{E, b}(X, Y):=& \frac{\sum_{i, j}^m \|X_i-X_j\|}{m^2} + \frac{\sum_{i, j}^n \|Y_i - Y_j\|}{n^2} -\frac{2\sum_{i=1}^m\sum_{j=1}^n \|X_i - Y_j\|}{mn}.
    \end{align*}
    Suppose $m = n$, then the U-statistic estimate of the energy distance $D_E(X, Y)$ is given as follows:
    \begin{align*}
        \widehat{D}_{E, e}(X, Y):= \frac{1}{n(n-1)}\sum_{i=1}^n\sum_{j \neq i}\left[\|X_i-X_j\|+\|Y_i-Y_j\|-\|X_i-Y_j\|-\|X_j-Y_i\|\right].
    \end{align*}
    \item (\textbf{HSIC and dCov}) Let $X \sim P_X$ and $Y \sim P_Y$ be random variables embedded in sets $\mathcal{X}$ and $\mathcal{Y}$, respectively. 
    The joint distribution is denoted as $P_{XY}$.
    Furthermore, let $k_{\mathcal{X}}$ and $k_{\mathcal{Y}}$ be distance-induced kernels (given in Lemma \ref{ex:dk}) on $\mathcal{X}$ and $\mathcal{Y}$ with respective RKHS, i.e., $\mathcal{F}_{\mathcal{X}}$ and $\mathcal{F}_{\mathcal{Y}}$, and $k((x, y), (x', y')) := k_{\mathcal{X}}(x, x')k_{\mathcal{Y}}(y, y')$. 
    Then,
    $$
        \mathcal{V}^2(X, Y)=4\gamma_k^2\left((X, Y), Z\right),
    $$
    where $Z$ follows the product of marginal distributions, $P_XP_Y$.
    Moreover, the biased and U-statistic empirical estimate of $\mathcal{V}^2(X, Y)$ can be obtained by multiplying the corresponding empirical estimates of HSIC by $4$. 
    That is, given sample matrices $\mathbf{Z}:=((X_1, Y_1)^T, \cdots, (X_n, Y_n)^T)^T$. 
    Let $\mathbf{X} = (X_1,\dots, X_n)^T$, and $\mathbf{Y}:= (Y_1,\dots, Y_n)^T$, where $(X_i, Y_i) \stackrel{\text{i.i.d.}}{\sim} P_{XY}$, $\forall i$.
    Let $\mathbf{K}$ and $\mathbf{H}$ be the $n \times n$ kernel matrices for $\mathbf{X}$ and $\mathbf{Y}$, respectively, with entries $K_{i j}=\left\|X_i - X_j\right\|$ and $L_{i j}=\left\|Y_i- Y_j\right\|$, $\forall i, j$. 
    Let $\mathbf{H}=\mathbf{I}_n-\frac{1}{n} \mathbf{1}_n \mathbf{1}_n^{\top}$ be the centering matrix, we have
    \begin{align*}
        \widehat{\mathcal{V}}_{b}^2\left(X, Y\right) &:= \frac{1}{n^2} \operatorname{tr}(\mathbf{K H L H});\\
        \widehat{\mathcal{V}}_{u}^2\left(X, Y\right) &:= \frac{1}{n(n-1)} \sum_{i \neq j}\left[ K_{i j} L_{i j}+K_{i i} L_{j j}-K_{i  j} L_{j j}-K_{i i} L_{i j}\right].
    \end{align*}
\end{enumerate}
\end{lemma}

Combining the above results (Lemma \ref{ex:dk} and Lemma \ref{lem:DE_MMD}), we can first substitute the constants $\nu$ and $l$ mentioned in the concentration inequalities for MMD (Corollary \ref{coro:MMD}) and the one for HSIC (Corollary \ref{coro:HSIC}) with the constants in Lemma \ref{ex:dk} to get the inequalities for MMD and HSIC with the distance-induced kernel.
By multiplying the upper bound with the factor $2$ and $4$ as mentioned in Lemma \ref{lem:DE_MMD}, the uniform concentration inequalities for ED and dCov are derived as follows, the proofs are referred to Section \ref{sec:proof-3} and Section \ref{sec:proof-4}, respectively.
\begin{corollary}[Energy Distance]
\label{cor:ED}
    Adopt notations in Corollary \ref{coro:MMD}.
    Suppose there exists a positive constant $\nu_{\text{dis}}$ and a fixed point $u \in \mathcal{U}$, such that $\sup_{u \in\mathcal{U}} \|u-u_0\| \leq \nu_{\text{dis}}$. 
    
    Replace the kernel function $k(\cdot, \cdot)$ in Corollary \ref{coro:MMD} with the distance-induced kernel function based on point $u_0$ (Lemma \ref{ex:dk}).
    Let $c_1$ and $c_2$ be constants defined as follows:
    \begin{align*}
        c_1 :=& 16\nu_{\text{dis}} \max\left\{\rho_x^{-1}, \rho_y^{-1}\right\},\\
        c_2 :=& 16\sqrt{2\pi}\max\bigg\{\rho_y^{-1}\left(1 + \rho_y^{-1}\right), \rho_x^{-1}\left(1 + \rho_x^{-1}\right)\bigg\}.
    \end{align*}
    Then, $\forall \delta \in (0, 1)$, with probability at least $1-\delta$, the following inequality holds for the unbiased estimator $\widehat{D}_{E, u}$:
    \begin{align*}
        \sup_{(h_\mathcal{X}, h_\mathcal{Y}) \in \mathcal{H}} \bigg|\widehat{D}_{E, u}(h_\mathcal{X}(\mathbf{X}), h_\mathcal{Y}(\mathbf{Y})) -& D_{E, u}(h_\mathcal{X}(X), h_\mathcal{Y}(Y))\bigg|\\
        &\leq c_1\sqrt{\frac{\log \left(\frac{2}{\delta}\right)}{m+n}} + \frac{c_2}{m+n} \mathbb{E}\left[\mathcal{G}(\mathcal{H}(\mathbf{X}, \mathbf{Y}))\right].
    \end{align*}
    Similarly, $\forall \delta \in (0, 1)$, with probability at least $1-\delta$, the following inequality holds for the biased estimator $\widehat{D}_{E, b}$:
    \begin{align*}
        \sup_{(h_\mathcal{X}, h_\mathcal{Y}) \in \mathcal{H}} \bigg|\widehat{D}_{E, b}(h_\mathcal{X}(\mathbf{X}),& h_\mathcal{Y}(\mathbf{Y})) - D_{E, b}(h_\mathcal{X}(X), h_\mathcal{Y}(Y))\bigg|\\
        &\leq 4\nu_{\text{dis}}\frac{\rho_x^{-1}\rho_y^{-1}}{m+n} + c_1\sqrt{\frac{\log \left(\frac{2}{\delta}\right)}{m+n}} + \frac{c_2}{m+n} \mathbb{E}\left[\mathcal{G}(\mathcal{H}(\mathbf{X}, \mathbf{Y}))\right].
    \end{align*}
    In cases where $m = n$, $\forall \delta \in (0, 1)$, with probability at least $1-\delta$, we have
    \begin{align*}
        \sup_{(h_\mathcal{X}, h_\mathcal{Y}) \in \mathcal{H}} \bigg|\widehat{D}_{E, e}(h_\mathcal{X}(\mathbf{X}), h_\mathcal{Y}(\mathbf{Y})) &- D_{E, e}(h_\mathcal{X}(X), h_\mathcal{Y}(Y))\bigg|\\ 
        \leq & 16\sqrt{2}\nu_{\text{dis}} \sqrt{\frac{\log \left(\frac{2}{\delta}\right)}{n}} + \frac{8\nu_{\text{dis}}}{n} + \frac{48\sqrt{2\pi}}{n}\mathbb{E}\left[\mathcal{G}(\mathcal{H}(\mathbf{X}, \mathbf{Y}))\right].
    \end{align*}
    \end{corollary}

\begin{corollary}[Distance Covariance]
\label{cor:dCov}
    Suppose there exists a positive constant $\nu_{\text{dis}}$, and fixed points $x_0 \in \mathcal{X}$, $y_0 \in \mathcal{Y}$, such that $\sup_{x \in  \mathcal{X}} \|x-x_0\| \leq \nu_{\text{dis}}$ and $\sup_{y \in \mathcal{Y}} \|y-y_0\| \leq \nu_{\text{dis}}$.
    Let $k_\mathcal{X}$ and $k_\mathcal{Y}$ be the distance-induced kernels (Lemma \ref{ex:dk}) based on $(\mathcal{X}, x_0)$ and $(\mathcal{Y}, y_0)$, respectively.
    Let $k$ be a reproducing kernel function such that $k((x, y), (x', y')) = k_\mathcal{X}(x, x')k_\mathcal{Y}(y, y')$. Then, replacing the kernel function referred in Corollary \ref{coro:MMD}, similar concentration inequalities hold for the empirical estimates of dCov. More specifically, $\forall \delta \in (0, 1)$, with probability at least $1-\delta$, we have either one of the following inequalities:
    \begin{align*}
        \sup_{h \in \mathcal{H}}\Bigg|\widehat{\mathcal{V}}_e^2\left(\mathbf{X}, \mathbf{Y}\right) - \mathcal{V}^2\left(X, Y\right)\Bigg| \leq 32\sqrt{2}\nu_{\text{dis}}^2\sqrt{\frac{\log(2/\delta)}{n}} + \frac{192\sqrt{\pi}}{n}\nu_{\text{dis}}\mathbb{E}[\mathcal{G}(\mathcal{H}(\mathbf{X}, \mathbf{Y}))],
    \end{align*}
    \begin{align*}
        \sup_{h \in \mathcal{H}}\Bigg|\widehat{\mathcal{V}}_b^2\left(\mathbf{X}, \mathbf{Y}\right) - \mathcal{V}^2(X, Y)\Bigg| \leq& \frac{4\nu_{\text{dis}}^2}{n} + \\
        &32\sqrt{2}\nu_{\text{dis}}^2\sqrt{\frac{\log(2/\delta)}{n}} + \frac{192\sqrt{\pi}}{n}\nu_{\text{dis}}\mathbb{E}[\mathcal{G}(\mathcal{H}(\mathbf{X}, \mathbf{Y}))].
    \end{align*}
\end{corollary}

Notice that the above corollaries serve as the first known uniform concentration inequalities for the empirical estimators of energy distance and distance covariance regarding a specified set of function transforms.

\subsection{Verification of conditions for kernels}
\label{sec::verification}
In this section, we show that the assumptions (Assumption \ref{assum:BandL}) required for the reproducing kernel function $k$ are satisfied for several commonly used reproducing kernel functions. 
The examples listed in this section are summarized in Table \ref{tab:kernels}.
The first column refers to the name of the related kernels; 
the second one refers to the technical expression for it; 
the third one is the assumption required to upper bound the constants of the kernels as mentioned in Assumption \ref{assum:BandL}; 
and the last two columns refer to the constants $\nu$ and $l$ as mentioned in Assumption \ref{assum:BandL}, respectively.

\begin{table}[htbp]
\caption{Verification of Assumption \ref{assum:BandL} for commonly used kernels in Corollary \ref{coro:MMD}.}
\scriptsize
    \centering
    \begin{tabular}{ccccc}
    \toprule
         \textbf{Kernel} & $\boldsymbol{k(u, u')}$ & \textbf{Assumption} & $\sup_{u \in \mathcal{U}}\boldsymbol{k(u,u)}$ & $\boldsymbol{M}_\text{Lip}\boldsymbol{(k)}$\\
         \rowcolor[gray]{.9}
         Bilinear & $u^Tu'$ & $\sup_{u \in \mathcal{U}}\|u\| \leq b_{\text{bi}}$ & $b_{\text{bi}}^2$ & $2b_{\text{bi}}$ \\
         \rowcolor[gray]{.85}
         Polynomial & $(\alpha u^T u'+1)^d$ & $\sup_{u \in \mathcal{U}}\|u\| \leq b_{\text{pl}}$ & $(\alpha b_{\text{pl}}^2+1)^d$ & $2\alpha db_{\text{pl}}(\alpha b_{\text{pl}}^2+1)^{d-1}$\\
         \rowcolor[gray]{.9}
         Sigmoid & $\operatorname{tanh}(\alpha u^Tu' + \beta)$ & $\sup_{u \in \mathcal{U}}\|u\| \leq b_{\text{sig}}$ & $1$ & $2\alpha b_{\text{sig}}$ \\
         \rowcolor[gray]{.85}
         Gaussian & $\exp(-\sigma^{-2}\|u-u'\|^2)$ & $\sigma > 0$ & $1$ & $2\sigma^{-1}\sqrt{2/e}$\\
         \rowcolor[gray]{.9}
         Laplacian & $\exp(-\sigma^{-1}\|u-u'\|_1)$ & $\sigma > 0$& $1$ & $2\sigma^{-1}\sqrt{d}$\\
    \bottomrule
    \end{tabular}
    \label{tab:kernels}
\end{table}

The kernels that we examine include 
dot product kernels (Section \ref{sec:dot-kernel}) and 
translation invariant kernels (Section \ref{sec:translation-invariant-kernel}). 

\subsubsection{Dot product kernels}
\label{sec:dot-kernel}
Dot product kernels are a fundamental class of kernel functions used in machine learning algorithms, particularly in kernel methods like \textit{support vector machines} (SVMs) (\cite{scholkopf2002learning}), {Kernel Principal Component Analysis} (KPCA) (\cite{scholkopf1998nonlinear}), and {Kernel Ridge Regression, (\cite{saunders1998ridge})}. 
These kernels are based on the inner product (dot product) of input feature vectors and enable algorithms to capture linear and non-linear patterns by implicitly mapping data into higher-dimensional feature spaces without explicitly computing the coordinates in that space.

In the following, we verify our assumptions with three dot product reproducing kernel examples.
To begin with, we consider the bilinear reproducing kernel, which corresponds to the inner product operation in the Euclidean space.

\begin{lemma}[Bilinear Kernel]
\label{ex:bilkernel}
    Given a \textit{bilinear kernel} $k(u, u'): \mathcal{U}\times\mathcal{U} \mapsto \mathbb{R}$, where $\mathcal{U} \subseteq \mathbb{R}^d$ defined as follows:
    \begin{align*}
        k(u, u'):= u^Tu', \quad \forall u, u' \in \mathcal{U}.
    \end{align*}
    Suppose $\sup_{u \in \mathcal{U}} \|u\| \leq b_{\text{bi}}$, we have
    \begin{enumerate}
        \item $\sup_{u \in \mathcal{U}} k(u, u) \leq b_{\text{bi}}^2$, that is, $\nu \leq b_{\text{bi}}^2$;
        \item $M_\text{Lip}(k) \leq 2 b_{\text{bi}}$.
    \end{enumerate}
\end{lemma}
A proof of the above lemma is in Section \ref{sec:proof-4}.

As a higher-order extension of the bilinear kernel, we investigate the polynomial kernel in the following lemma.
\begin{lemma}[Polynomial Kernel]
\label{ex:polykernel}
    Given a positive number $\alpha > 0$ and an integer $d \in \mathbb{N}$, the polynomial kernel on $\mathcal{U} \subseteq \mathbb{R}^d$ is defined as
    \begin{align*}
        k(u, u'):= \left(\alpha u^Tu'+1\right)^d, \quad \forall u, u' \in \mathcal{U}.
    \end{align*}
    Suppose $\sup_{u \in \mathcal{U}} \|u\| \leq b_{\text{pl}}$, we have
    \begin{enumerate}
        \item $\sup_{u \in \mathcal{U}} k(u, u) \leq \left(\alpha b_{\text{pl}}^2 + 1\right)^d$, that is, $\nu \leq \left(\alpha b_{\text{pl}}^2 + 1\right)^d$;
        \item $M_\text{Lip}(k) \leq 2\alpha db_{\text{pl}}\left(\alpha b_{\text{pl}}^2 + 1\right)^{d-1}$.
    \end{enumerate}
\end{lemma}
A proof of the above lemma is in Section \ref{sec:proof-5}.

As a kernel function that resembles the activation function used in the early layers of a neural network, the sigmoid kernel is widely applied in support vector machines and neural network contexts. 
We verify the Assumption \ref{assum:BandL} for the sigmoid kernel in the following example.
\begin{lemma}[Sigmoid / Hyperbolic Tangent Kernel]
\label{lem:sigmoid}
    Given constants $\alpha, \beta \in \mathbb{R}$, the sigmoid kernel on $\mathcal{U} \subseteq \mathbb{R}^d$ is defined as
    \begin{align*}
        k(u, u') := \operatorname{tanh}(\alpha u^Tu'+\beta), \quad \forall u, u' \in \mathcal{U},
    \end{align*}
    which is similar to the sigmoid function in logistic regression.
    Here, observed from the $\operatorname{tanh}(x):= {1}/{(1+e^{-x})}$ function, suppose 
$\sup_{u \in \mathcal{U}}\|u\| \leq b_{\text{sig}}$, we have
\begin{enumerate}
    \item $\sup_{u \in \mathcal{U}} k(u,u) \leq 1$, that is, $\nu \leq 1$;
    \item $M_\text{Lip}(k) \leq 2\alpha b_{\text{sig}}$.
\end{enumerate}
\end{lemma}
A proof of the above lemma is in Section \ref{sec:proof-6}.

In the above bilinear, polynomial, and sigmoid kernel examples, the compact support assumption is required for the realization of the bounded condition $\sup_{u}k(u,u) < \infty$.

\subsubsection{Translation invariant kernels}
\label{sec:translation-invariant-kernel}
Translation invariant kernel is a widely applied function group considered in existing MMD-related works, including \cite{tolstikhin2016minimax, muandet2017kernel, JMLR:v11:ghiasi-shirazi10a}.
It is also considered an extension of the radial kernels mentioned in \cite{tolstikhin2016minimax}.
Examples include the Gaussian, Matérn and inverse multiquadric kernels.

In the following, we first consider the translation invariant kernel, showing the sufficient conditions under which Assumption \ref{assum:BandL} is satisfied.
Afterward, we consider the Gaussian and Laplacian kernels as illustrative examples.
\begin{lemma}[Translation Invariant Kernel]
\label{lem:translation}
   Given a set $\mathcal{U} \subset \mathbb{R}^d$, a kernel function $k: \mathcal{U} \times \mathcal{U} \mapsto \mathbb{R}$ is called \textit{translation invariant} iff $k(u_1+\Delta u, u_2+\Delta u) = k(u_1, u_2)$, $\forall u_1, u_2 \in \mathcal{U}$ and $u_1+\Delta u \in \mathcal{U}$. In other words, there exists a function $\widetilde{k}: \mathbb{R}^d \mapsto \mathbb{R}$, such that $k(u_1, u_2) = \widetilde{k}(u_1-u_2)$, $\forall u_1,u_2\in \mathcal{U}$.
   Suppose there exist positive constants $\nu_{\text{t}}, l_{\text{t}}$, such that the function $\widetilde{k}$ is $l_{\text{t}}$-Lipschitz, and $\widetilde{k}(\mathbf{0}_d) \leq \nu_{\text{t}}$, we have
   \begin{enumerate}
    \item $\sup_{u \in \mathcal{U}} k(u,u) \leq \nu_{\text{t}}$, that is, $\nu \leq \nu_t$;
    \item $M_\text{Lip}(k) \leq 2l_t$.
\end{enumerate}
\end{lemma}
A proof of the above lemma is in Section \ref{sec:proof-7}.

Next, we discuss two popular choices of translation invariant kernels: the Gaussian and Laplacian kernels.

\begin{lemma}[Gaussian Kernel]
\label{ex:Gk}
    Given a set $\mathcal{U} \subset \mathbb{R}^d$ and a positive constant $\sigma$, the Gaussian kernel $k_\sigma: \mathcal{U} \times \mathcal{U} \mapsto \mathbb{R}$ is defined as follows:
    \begin{align*}
        k(u, u'):=\exp\left(-\sigma^{-2}\|u-u'\|^2\right).
    \end{align*}
    Without extra assumptions, we have
    \begin{enumerate}
    \item $\sup_{u \in \mathcal{U}} k(u,u) \leq 1$, that is, $\nu \leq 1$;
    \item $M_\text{Lip}(k) \leq 2\sigma^{-1}\sqrt{2/e}$.
\end{enumerate}
\end{lemma}
A proof of the above lemma is in Section \ref{sec:proof-8}.

\begin{lemma}[Laplacian Kernel]
\label{ex:Lk}
Given a set $\mathcal{U} \subset \mathbb{R}^d$ and a positive constant $\sigma$, the Laplacian kernel $k_\sigma: \mathcal{U} \times \mathcal{U} \mapsto \mathbb{R}$ is defined as follows:
    \begin{align*}
        k(u, u'):=\exp\left(-\sigma^{-1}\|u-u'\|_1\right),
    \end{align*}
    where $\|\cdot\|_1$ is the $l_1$-norm in $\mathbb{R}^d$.
    Without extra assumptions, we have
    \begin{enumerate}
    \item $\sup_{u \in \mathcal{U}} k(u,u) \leq 1$, that is, $\nu \leq 1$;
    \item $M_\text{Lip}(k) \leq 2\sigma^{-1}\sqrt{d}$.
    \end{enumerate}
\end{lemma}
A proof of the above lemma is in Section \ref{sec:proof-9}.

\section{Applications of our inequalities in empirical kernel-based methods}
\label{sec:consistency_empirical}
In this section, we show details of how our main result can be applied in various settings to establish new and improved error bounds.
We discuss the dCov-based dimension reduction (\cite{sheng2013direction, sheng2016sufficient, chen2019sufficient, yu2019distance}) in Section \ref{sec::dimensionreduction}, 
dCov-based independent component analysis (\cite{matteson2017independent}) in Section \ref{sec::ica}, 
generalized MMD regarding kernel choice (\cite{fukumizu2009kernel}) in Section \ref{sec:::kernel_generalized}, 
MMD-based fair representations (\cite{rychener2022metrizing}) in Section \ref{sec::fairness}, 
the minimum MMD-based generative model search (\cite{briol2019statistical, cherief2022finite}) in Section \ref{sec:::generative}, 
Finally, we discuss MMD-based generative adversarial networks (\cite{li2017mmd}) in Section \ref{sec::mmdgan}.

\subsection{Dimension reduction}
\label{sec::dimensionreduction}

Recall the dimension reduction approach in Section \ref{sec:dimension-reduction-2}. 
Consider the empirical estimate of the subspace $\mathcal{S}$ proposed in \cite{yu2019distance}. 
In the following, we show that based on the uniform concentration inequality for dCov, the non-asymptotic estimation error bounds can be derived.
The proof is referred to Section \ref{sec:proof-cor6.1}.

\begin{corollary}[Dimension Reduction; \cite{yu2019distance}]
\label{ex:DISCA}
    Given the sample set $\mathbf{X}:= (X_1,\dots,X_n)^T \in \mathbb{R}^{n \times p}$, where $X_i \stackrel{i.i.d.}{\sim} P$, $\forall i$, and $\mathbf{Y}:= (Y_1, \dots, Y_n)^T  \in \mathbb{R}^{n \times q}$, with $Y_i \stackrel{i.i.d.}{\sim} Q$, $\forall i$. Let $S^{p-1}:=\{u\in \mathbb{R}^p: \|u\|=1\}$.
    Suppose $\sup _{x\in\mathcal{X}}\|x\|\leq \nu_{\text{dis}}$, and $\sup _{y\in\mathcal{Y}}\|y\|\leq \nu_{\text{dis}}$, then the Gaussian complexity of the projected set $(\operatorname{P}_{S^{p-1}}\mathbf{X}, \mathbf{Y}):=\{(\mathbf{X}u, \mathbf{Y}) \mid u \in S^{p-1}\}$, could be bounded as follows.
    \begin{align*}
        \mathbb{E}[\mathcal{G}(\operatorname{P}_{S^{p-1}}\mathbf{X}, \mathbf{Y})]=\mathbb{E}\left[\sup_{u \in \mathcal{S}}\sum_{i=1}^n\left\langle \xi_i, (u^TX_i, Y_i) \right\rangle\right] \leq \nu_{\text{dis}}n^{1/2}.
    \end{align*}
    That is, based on the uniform concentration inequality for dCov (Corollary \ref{cor:dCov}), there exists constants $C_1$, $C_2$, such that, $\forall \delta \in (0,1)$, with probability $1-\delta$, we have:
    \begin{align*}
    \left|\sup_{\|u\|=1, u \in \widehat{\mathcal{S}}_{\text{old}}^\perp}\widehat{\mathcal{V}}^2(\mathbf{X}u, \mathbf{Y})-\sup_{\|u\|=1, u \in \widehat{\mathcal{S}}_{\text{old}}^\perp}{\mathcal{V}}^2(Xu, Y) \right| \leq \nu_{\text{dis}}^2n^{-\frac{1}{2}}\left(C_1+C_2\sqrt{\ln\left(2/{\delta}\right)}\right),
    \end{align*}
    where $\widehat{\mathcal{V}}^2$ refers to both the biased estimate $\widehat{\mathcal{V}}^2_b$ and the U-statistic estimate $\widehat{\mathcal{V}}^2_e$ of dCov (Lemma \ref{lem:DE_MMD}).
\end{corollary}
As observed in \cite[Corollary 3.4, Theorem 3.5, Theorem 3.8, Corollary 3.10]{yu2019distance}, in cases where the output dimension reduction subspace is equipped with a correct dimension estimate, the above uniform concentration inequality can be employed to prove the statistical consistency of the estimated subspace.
Moreover, under mild regularization conditions, the above inequality can be employed to derive a probability bound on the correctness of the estimated direction \(u\) in each step.

\subsection{Independent component analysis}
\label{sec::ica}

Recall the dCov-based independent component analysis that is described in Section \ref{sec:dis-ica}. 
Suppose the aforementioned optimization problem \eqref{eq:matteson} can be minimized to $0$, 
denote the solution as $\theta^*$, then $\{S_{k}\}_{k=1}^d$ would be the independent components collected from samples.
The following inequality provides the error bounds for the sequential estimate as mentioned in \eqref{eq:mattesson_seq}.
The proof is referred to \ref{sec:proof-cor6.2}.

\begin{corollary}[ICA, \cite{matteson2017independent}]
\label{ex:ICA}
Let $Y$ be a random vector embedded in set $\mathcal{Y} \subset \mathbb{R}^d$, where $\sup_{y \in \mathcal{Y}}\|y\| \leq  B$.
Given the data matrix $\mathbf{Y}:=(Y_1, \dots, Y_n)^T$, where $Y_i \stackrel{i.i.d.}{\sim} Y$, $i=1,\dots,n$.

Define $\theta$ as the length $p=d(d-1)/2$ vectorized triangular array of rotation angles, indexed by $\{i,j: 1\leq i<j \leq d\}$. 
Set the valid set for the rotation angles in the $k$-th row of $\Theta$ as $\Theta_k:=\{\theta^{(k:k)}=(\theta_{k,k+1}, \dots, \theta_{k,d}): \theta_{k,j} \in [0, (1+I_{(k=1)})\pi), j=i+1,\dots, d\}$, $\forall k$.
Denote the rotation matrix according to angles $\theta$ as $\mathbf{W}_{\theta}$.
Define the column-wise separation of random vector $Y$ rotated with $\theta$, i.e., $\mathbf{W}_{\theta}Y$, as $S := (S_1,\dots,S_d)^T$, equipped with the data matrix $\boldsymbol{S}:= \mathbf{Y}\mathbf{W}_{\theta}^T = (\boldsymbol{S}_1(\theta), \dots, \boldsymbol{S}_d(\theta))$. Denote the components of $\boldsymbol{S}(\theta)$ after index $k$ as $\boldsymbol{S}_{k^+}(\theta):=(\boldsymbol{S}_{k+1}(\theta) \cdots \boldsymbol{S}_{d}(\theta))$, $ k=1,\dots,d-1$.
Denote the sequential estimate of $\theta$ as $\widehat{\theta}_n$.
The Gaussian complexity regarding the set $\Theta_k(\mathbf{Y}):=\{\left(\boldsymbol{S}_k(\theta), \boldsymbol{S}_{k^+}(\theta)\right) \mid \theta^{(k:k)} \in \Theta_k\}$ can be bounded as follows:
\begin{align*}
\mathbb{E}\left[\mathcal{G}(\Theta_k(\mathbf{Y}))\right]=\mathbb{E}\left[\sup_{\theta^{(k:k)}\in \Theta_k}\sum_{i=1}^n \left\langle \xi_i, \left(\boldsymbol{S}_k(\theta), \boldsymbol{S}_{k^+}(\theta)\right)_i\right\rangle\right]
\leq \sqrt{n(d-k+1)}, \quad\forall k.
\end{align*}
That is, based on the uniform concentration inequality for dCov (Corollary \ref{cor:dCov}), there exist constants $C_1$ and $C_2$, such that, for any $k = 1,\dots,d$, with probability at least $1-\delta$, we have
\begin{align*}
    \Bigg|\min_{\theta^{(k:k)}\in \Theta_k} \widehat{\mathcal{V}}^2\left(\boldsymbol{S}_k(\theta), \mathbf{S}_{k^{+}}(\theta)\right) - \min_{\theta^{(k:k)}\in \Theta_k} \mathcal{V}^2 & \left(S_k(\theta), S_{k^{+}}(\theta)\right)\Bigg| \\
   \leq & B\Bigg(C_1\sqrt{\frac{d-k+1}{n}} + C_2B\sqrt{\frac{\ln(2/\delta)}{n}}\Bigg),
\end{align*}
where $\widehat{\mathcal{V}}^2$ refers to both the biased estimate $\widehat{\mathcal{V}}^2_b$ and the U-statistic estimate $\widehat{\mathcal{V}}^2_e$ of dCov (Lemma \ref{lem:DE_MMD}).
\end{corollary}

In \cite{matteson2017independent}, the asymptotic consistency of the above empirical estimator $\widehat{\theta}$ was established.
However, the non-asymptotic error bounds and the specific convergence rates were not discussed.
As shown with the above corollary, our inequality fills this gap.
Our result can be used to provide explicit non-asymptotic error bounds for the corresponding empirical estimators, showing the theoretical performance of MMD-based models with a specific convergence rate.

\subsection{Generalized MMD regarding kernel choice}
\label{sec:::kernel_generalized}

Recall the Generalized MMD regarding kernel choice that is described in Section \ref{sec:generalize-MMD-kernel}. 
In \cite{fukumizu2009kernel}, the statistical consistency of the generalized MMD, i.e., $\gamma(X, Y)$, is established in the following inequality, that is, $\forall \delta \in (0,1)$, with probability at least $1-\delta$, we have
\begin{equation}
\label{eq:intro_kernelMMD}
\begin{aligned}
    &\left|\widehat{\gamma}_{b}(\mathbf{X}, \mathbf{Y}; \mathcal{K}) - \gamma_k(X, Y; \mathcal{K})\right| \leq\\
    &\,\, 4\sqrt{\frac{1}{n(n-1)} R_m\left(\mathbf{X};\mathcal{K}\right)+ R_n\left(\mathbf{Y};\mathcal{K}\right)}+2\left(\sqrt{2}+3\sqrt{\log \frac{4}{\delta}}\right)\sqrt{\sup_{k\in\mathcal{K}, u \in \mathcal{U}}k(u,u)}\sqrt{\frac{m+n}{mn}},
\end{aligned}
\end{equation}
where $R_m(\mathbf{X};\mathcal{K})$ refers to the empirical \textit{Rademacher chaos} \cite[p. 105]{ledoux2013probability} of the kernel class $\mathcal{K}$ with respect to the data matrix $\mathbf{X}$, that is, 
\begin{equation}
\label{eq:rademacherchaos}
    R_m(\mathbf{X};\mathcal{K}) = \mathbb{E}_\rho\left[\sup_{k \in \mathcal{K}}\bigg|\sum_{i<j}\rho_i\rho_jk(X_i, X_j)\bigg|\Bigg| \mathbf{X}\right].
\end{equation}
Here, $\rho_1,\dots,\rho_n$ are independent uniform $\{\pm 1\}$-valued Rademacher random variables.

Comparing the above inequality with the one established from Theorem \ref{thm:general}, it is worth noticing that the above inequality is established in a setting different from the context of this paper.
More specifically, we have: 1. The Rademacher chaos is a tool established in the \textit{U-process} theory \cite{de2012decoupling}, which can be considered as a higher-order variant of the Gaussian complexity mentioned in our main result. 2. Due to the symmetric property of the reproducing kernel function, i.e., $k(x, y) = k(y, x)$, $\forall x, y \in \mathcal{U}, $ $\forall k \in \mathcal{K}$, the kernel function class is not equivalent with the function class $\mathcal{G}:=\{(h_\mathcal{X},h_\mathcal{Y})\}$ considered in our work. 
Nonetheless, in the case where the kernel class is given in the following formula, taking the supremum over $k \in \mathcal{K}$ is equivalent to taking the supremum over a function set $\mathcal{F}$.
That is, given a reproducing kernel $k^*$, we have
\begin{align}
\label{eq:ek_class}
    \mathcal{K} = \left\{k^*(f(X), f(Y)) \mid f: \mathcal{X} \mapsto \mathcal{U}, f \in \mathcal{F}\right\}.
\end{align}
In the above setting, suppose the involved function class $\mathcal{F}$ is not a singleton, it can be observed that the classical non-asymptotic deviation bound cannot be applied to the convergence analysis of the empirical estimate $\widehat{\gamma}_{b}$ for the generalized MMD (Definition \ref{def:intro_generalizedMMD}).

In the following, we compare the upper bounds derived from our main result (Theorem \ref{thm:general}) with the inequality \eqref{eq:intro_kernelMMD} proposed in \cite{maurer2019uniform}, in the above special case \eqref{eq:ek_class}.
For simplicity, we assume that the kernel class $\mathcal{K}$ is composed of Gaussian kernels with different bandwidths, that is, we have
\begin{align}
\label{eq:Gaussiankernelclass}
    \mathcal{K} := \{k_\sigma \mid k_\sigma(u,u')=\exp\left(\|u-u'\|^2/\sigma^2\right), \sigma \in [\sigma_l, \sigma_h]\},
\end{align}
where $\sigma_l$ and $\sigma_h$ are the lower and upper bounds of the bandwidth $\sigma$.
In that case, the uniform concentration inequality for MMD (Corollary \ref{coro:MMD}) leads to the following probability bound. 
The proof is referred to Section \ref{sec:proof-cor6.3}.
\begin{corollary}
\label{cor:generalizedMMD}
    Let $\gamma$ and $\widehat{\gamma}_b$ be the notations given in \eqref{def:intro_generalizedMMD}, \eqref{eq:intro_fukumizuMMD_b}, respectively. Let $\mathcal{K}$ be a class of Gaussian kernels given in \eqref{eq:Gaussiankernelclass}. Suppose $\mathbb{E}\|X\|^2 < \infty$, $\mathbb{E}\|Y\|^2 < \infty$, and $m/n< \infty$, $n/m < \infty$ as sample size $m, n \rightarrow \infty$. Then there exists constant $C_1$ and $C_2$, such that $\forall \delta \in (0,1)$, with probability at least $1-\delta$, we have
\begin{align}
\label{ineq:intro_fukumizufrommain}
    \left|\gamma^2(X, Y; \mathcal{K}) -\widehat{\gamma}^2_b(\mathbf{X}, \mathbf{Y}; \mathcal{K})\right| \leq  \frac{C_1}{\sqrt{m+n}} + C_2\sqrt{\frac{\log(2/\delta)}{m+n}}.
\end{align}
\end{corollary}

For comparison, based on the result provided in \cite{fukumizu2009kernel}, the upper bound provided in \eqref{eq:intro_kernelMMD} in the above special case can be reformulated as follows:
\begin{align*}
    \left(C_1+C_2\sqrt{\log(2/\delta)}\right)\sqrt{\frac{m+n}{mn}}.
\end{align*}
Under the assumption that $m/n$ and $n/m < \infty$, it can be inferred that $m/(m+n)$ and $n/(m+n)$ are both constrained within a certain range. As a consequence, the term $(m+n)/mn$ is equipped with the same order as the term $1/(m+n)$. That is, even if the focus of the theoretical results built in \cite{fukumizu2009kernel} is different, in a special setting \eqref{eq:ek_class} where the choice of kernel functions is equivalent with the choice of function transformations, the order of the upper bounds derived from our main result (Theorem \ref{thm:general}) is the same as the one \eqref{eq:intro_kernelMMD} given in \cite{fukumizu2009kernel}.

\subsection{Measuring fairness}
\label{sec::fairness}

Recall the work that is described in Section \ref{sec:fairness-2} where MMD is used as a penalty to help achieve fair representation in a supervised learning problem.
Based on the uniform concentration inequality for MMD (Corollary \ref{coro:MMD}), in the following context, we provide an estimation error bound for the MMD penalty term built in the above problem.
For simplicity, we assume that the function $\rho$ is an identity map in the following context.
The following corollary is a straightforward conclusion derived from the uniform concentration inequality for MMD (Corollary \ref{coro:MMD}).
\begin{corollary}[Metrizing Fairness, \cite{rychener2022metrizing}]
\label{ex:fair}
    Given a protected attribute $A \in \mathcal{A}=\{0,1\}$, a feature vector $X$ embedded in set $\mathcal{X} \subseteq \mathbb{R}^d$, a sample matrix $\mathbf{X}:= (X_1,\dots, X_n)^T$, where $X_i \stackrel{i.i.d.}{\sim} P$, $\forall i$, and a family of Borel-measurable functions $\mathcal{H}$.
    Let $h^*$ be the empirical estimator obtained in the optimization problem \eqref{eq:fair}, under the assumptions stated in Corollary \ref{coro:MMD}, there exist constants $C_1$, $C_2$, such that $\forall \delta \in (0,1)$, with probability at least $1-\delta$, we have
    \begin{align*}
        \gamma_k^2(h^*(X|A=0), h^*(X|A=1)) \leq & \widehat{\gamma}_k^2(h^*(X|A=0), h^*(X|A=1)) \\
        & + \frac{C_1}{n} \mathbb{E}\left[\mathcal{G}(\mathcal{H}(\mathbf{X}))\right] + C_2\sqrt{\frac{\ln(2/\delta)}{n}},
    \end{align*}
    where $\mathbf{X}|A=0$ and $\mathbf{X}|A=1$ refers to the data matrix $\mathbf{X}$ conditioned on the protective attribute $A$, and $\widehat{\gamma}_k$ refers to any of the three empirical estimates of MMD, i.e., $\widehat{\gamma}_{u, k}$ \eqref{eq:intro_unbiased_MMD}, $\widehat{\gamma}_{b, k}$ \eqref{eq:intro_biased_MMD}, or $\widehat{\gamma}_{k,e}$ \eqref{eq:intro_U_MMD}.
\end{corollary}
Note that in circumstances where neural networks are utilized, the Gaussian complexity in the above inequality can be replaced with the one for neural networks as stated in Section \ref{sec::gaussianexample}.

While \cite{rychener2022metrizing} does not establish estimation error bounds for related empirical MMD estimators---leaving no theoretical justification that their MMD penalty indeed achieves a fair representation \(h^*\)---our result fills this gap. 
Specifically, given a sufficiently small empirical MMD estimate $\widehat{\gamma}_k$, our corollary implies that as \(m\) and \(n\) tend to infinity, the underlying population MMD will closely approximate the empirical MMD, thereby ensuring a fair representation \(h\).

\subsection{Minimum MMD-based generative model}
\label{sec:::generative}

Recall the MMD-based search for a generative model described in Section \ref{sec:mmd-estimators}. 
In scenarios where set $\Theta$ is not a singleton, the classical non-asymptotic deviation bound cannot be applied to the convergence analysis of the above empirical estimator $\widehat{\theta}_{m,n}$ \eqref{eq:intro_MDE(MMD,m,n)}.
On the contrary, our inequality for MMD (Corollary \ref{coro:MMD}) provides a generalization error bound for $\widehat{\theta}_{m,n}$ \eqref{eq:intro_MDE(MMD,m,n)}, presented in a formula similar to \eqref{eq:intro_briol}.
More specifically, the following estimation error bound is derived. 
The proof is referred to Section \ref{sec:proof-cor6.5}.
\begin{corollary}
\label{coro:mmd(m,n)}
    Let $\widehat{\theta}_{m,n}$ be the estimator given in \eqref{eq:intro_MDE(MMD,m,n)}. 
    Suppose that as sample size $m, n \rightarrow \infty$, we have $m/n$ and $n/m < \infty$.
    Under certain regularity conditions for the corresponding reproducing kernel function $k$, there exist constants $C_1 > 0$ and $C_2 > 0$, such that $\forall \delta \in (0,1)$, with probability at least $1-\delta$, we have
    \begin{align}
    \label{ineq:intro_MMD(n,n)_bound}
        \gamma_k^2\left(h_{\widehat{\theta}_{m,n}}(X), Y\right) \leq \inf_{\theta \in \Theta}\gamma_k^2(h_{\theta}(X), Y) + \frac{C_1}{m}\mathbb{E}[\mathcal{G}(\mathcal{H}_\Theta\left(\mathbf{X}\right))] + C_2\sqrt{\frac{\log(2/\delta)}{m+n}}.
    \end{align}
\end{corollary}
Note that the estimation error bound \eqref{eq:intro_briol} derived in \cite{briol2019statistical} is irrelevant with the sample size $m$ collected from the generated random vector $h_\theta(X)$.
On the contrary, the above error bound \eqref{ineq:intro_MMD(n,n)_bound} derived from our main result (Theorem \ref{thm:general}) reveals the estimation error stemmed from the involved function class $\mathcal{H}_\Theta$ and the sample size $m$.
As an illustrating example, suppose the function class $\mathcal{H}$ as mentioned in the estimation process of $\widehat{\theta}_{m,n}$ is a feed-forward neural network.
In that case, the Gaussian complexity $\mathbb{E}[\mathcal{G}(\mathcal{H}_\Theta\left(\mathbf{X}\right))]$ is equipped with order $O(m^{1/2})$ (\ref{prop:Gaussian}).
That is, as the sample size $m \rightarrow \infty$ and $n \rightarrow \infty$, the estimation error derived in inequality \eqref{ineq:intro_MMD(n,n)_bound} converges to zero in order $O(1/\sqrt{m+n})$.

\subsection{MMD-based generative adversarial network}
\label{sec::mmdgan}

Recall the MMA-based Generative Adversarial Network (GAN) described in Section \ref{sec:gan-2}. 
In the following, we show that an estimation error bound similar to the one presented in \eqref{eq:intro_briol} can be derived for the MMD GAN example, based on our uniform concentration inequality for MMD (Corollary \ref{coro:MMD}).
The proof is referred to Section \ref{sec:proof-cor6.6}.
It is worth noticing that in existing MMD-based generative models-related works \cite{li2017mmd, deka2023mmd}, the theoretical investigation of the corresponding optimizer is not mentioned.

\begin{corollary}[Error Bound in MMD GAN]
\label{coro:MMDGAN}
Let $X$, $Y$ be random vectors embedded in sets $\mathcal{X}$, $\mathcal{Y}$, respectively. Let $\mathcal{F}:=\{f:\mathcal{Y}\mapsto \mathcal{U}\}$ be the function class constructed for the composite reproducing kernel, and $\mathcal{H}:=\{h:\mathcal{X}\mapsto \mathcal{Y}\}$ be the function class given for the generative model.
The following min-max problem is considered in the MMD GAN (\cite{li2017mmd}):
\begin{align*}
    \min_{h \in \mathcal{H}}\max_{f \in \mathcal{F}}\widehat{\gamma}_{k\circ f}^2\left(h(\mathbf{X}), \mathbf{Y}\right),
\end{align*}
where $\mathbf{X}$ and $\mathbf{Y}$ refers to the data matrix collected from random vector $X$ and $Y$, respectively, and $k\circ f$ refers to the reproducing kernel function such that $k \circ f(x, x'):= k(f(x), f(x'))$, $\forall f$.
Let $f^*$, $h^*$ be the corresponding empirical estimators.
Under the assumptions utilized in Corollary \ref{coro:MMD}, there exists constants $C_1, C_2 >0$, such that, $\forall \delta \in (0,1)$, with probability at least $1-\delta$, we have
    \begin{align*}
        \gamma_{k \circ f^*}^2\left(h^*(X), Y\right) \leq & \inf_{h \in \mathcal{H}}\sup_{f \in \mathcal{F}}\gamma_{k \circ f}^2 \left(h(X), Y\right)\\
        & + \frac{C_1}{m+n}\mathbb{E}[\mathcal{G}(\mathcal{F}(\mathbf{Y}))+ \mathcal{G}(\mathcal{F}\circ \mathcal{H}(\mathbf{X}))] + C_2\sqrt{\frac{\ln \left(2/\delta\right)}{m+n}},
    \end{align*}
where $\mathcal{F}\circ \mathcal{H}:=\{f\circ h: f\circ h(x) = f(h(x)) \mid f \in \mathcal{F}, h \in \mathcal{H}\}$ is the function class consisting of composite functions, and $\widehat{\gamma}_k$ refers to any of the three empirical estimates of MMD, i.e., $\widehat{\gamma}_{u, k}$ \eqref{eq:intro_unbiased_MMD}, $\widehat{\gamma}_{b, k}$ \eqref{eq:intro_biased_MMD}, or $\widehat{\gamma}_{k,e}$ \eqref{eq:intro_U_MMD}.
\end{corollary}
Notice that the generalization bound built for the minimum MMD estimator could be considered as a special case of the above Corollay \ref{coro:MMDGAN}, where the function class $\mathcal{F}$ is chosen such that only an identity map is contained. 

The above generalization error bounds haven't been discussed in existing MMD GAN works.
In other words, our generalization error bounds can be considered the first theoretical error bounds regarding the MMD GAN model.

\section{Alternative approaches and advantages of our choice}
\label{sec:discussion1}

Our approach in this paper follows the framework in Maurer’s uniform concentration inequality (Lemma~\ref{lem:maurer}).
More specifically, we establish a general concentration inequality (Theorem \ref{thm:general}) with conditions such as the continuity seminorms in Assumption \ref{assum:maurer} and Gaussian Complexity in Definition \ref{def:Gauss} for the involved function class. 
We then show that our theorem applies when the function class corresponds to the kernel-based two-sample statistics. 
Consequently, a concentration inequality for MMD (Corollary \ref{coro:MMD}), as well as for other statistics such as HSIC, ED, and dCov, can be derived accordingly. 
In the following, we describe two alternative ways to establish concentration inequalities and the advantages of our approach. 
In Section \ref{sec:Rademacher-chaos}, an approach that follows the one in \cite{bartlett2002rademacher} eventually leads to an upper bound that involve \textit{Rademacher chaos} for the associated function class \cite[p. 105]{ledoux2013probability}.
We argue that our approach is more general.
In Section \ref{sec:supreme-ep}, a supremum of empirical processes approach is reviewed. 
Again, we will argue that our approach is more unified and, in many cases, leads to lower upper bounds.

\subsection{Rademacher chaos-based upper bound}
\label{sec:Rademacher-chaos}

The first inequality is built based on a modification for the proofs in \cite{bartlett2002rademacher}.
We can establish the following proposition. 
A proof can be found in Section \ref{sec:proof-prop42}.
\begin{proposition}[Alternative Uniform Concentration Inequality for MMD]
\label{prop:alternative-1}
Adopt notations in Corollary \ref{coro:MMD}. 
Recall a pair of random vectors $X$, $Y$, the corresponding data matrices $\mathbf{X}:=(X_1, \dots, X_m)^T$, $\mathbf{Y}:=(Y_1, \dots, Y_n)^T$, and an associated function set $\mathcal{H}:=\{h \mid h(x, y) = (h_\mathcal{X}(x), h_\mathcal{Y}(y))\}$.
Suppose $\sup_{u} k(u,u) \leq \nu$ (Assumption \ref{assum:BandL}).
$\forall \delta \in (0,1)$, with probability at least $1-\delta$, we have
\begin{align*}
\sup_{h \in \mathcal{H}}|\widehat{\gamma}_{k,b}(h_\mathcal{X}(\mathbf{X}), h_\mathcal{Y}(\mathbf{Y})) - {\gamma}_k&(h_\mathcal{X}(X), h_\mathcal{Y}(Y))| \leq \\
\left(\frac{1}{\sqrt{m}} + \frac{1}{\sqrt{n}}\right)\sqrt{{2 \nu} \log \frac{1}{\delta}} & + \frac{2}{\sqrt{m}}\left[\mathbb{E} \sup_{h_\mathcal{X}}\left|\sum_{i=1}^m\sum_{j=1}^m \rho_i \rho_j k(h_\mathcal{X}(X_i), h_\mathcal{X}(X_j))\right|\right]^{1/2}\\
& + \frac{2}{\sqrt{n}}\left[\mathbb{E} \sup_{h_\mathcal{X}}\left|\sum_{i=1}^n\sum_{j=1}^n \rho_i \rho_j k(h_\mathcal{Y}(Y_i), h_\mathcal{Y}(Y_j))\right|\right]^{1/2}.
\end{align*}
\end{proposition}
Note that the above quantity $\mathbb{E} \sup_{h_\mathcal{X}}\left|\sum_{i=1}^m\sum_{j=1}^m \rho_i \rho_j k(h_\mathcal{X}(X_i), h_\mathcal{X}(X_j))\right|$ is similar to the Rademacher complexity (Definition \ref{def:Rademacher}).
This quantity is named as the \textit{Rademacher chaos} for the set $\{k(h_\mathcal{X}(X_i), h_\mathcal{X}(X_j))\}$ with degree $2$ \cite[p. 105]{ledoux2013probability}.

Comparing the above inequality with the uniform concentration inequality for MMD (Corollary \ref{coro:MMD}), it can be observed that the above proposition does not impose a Lipschitz assumption regarding the involved reproducing kernel function $k$.
On the other hand, it requires a calculation of the corresponding Rademacher chaos, which is not required in our result. 
Nevertheless, we did not adopt the above inequality as our main result due to the following reasons:
\begin{enumerate}
\item In contrast to the previous proposition—which is limited to kernel-based statistics—our main result (Theorem \ref{thm:general}) goes beyond that scope. Specifically, Theorem \ref{thm:general} establishes a uniform concentration inequality for general Lipschitz two-sample statistics, with kernel-based statistics included as special cases.
    
\item Although \textit{Rademacher chaos} has been widely studied in kernel learning contexts (\cite{ying2010rademacher, zhang2023nearly}), the \textit{Rademacher chaos} for the machine learning models, such as multi-layer neural networks (Section \ref{sec:fairness-2}, \ref{sec:mmd-estimators}, \ref{sec:gan-2}) haven't been well studied yet, leading to an uncertain upper bound.

\end{enumerate}

\subsection{Supremum of empirical processes}
\label{sec:supreme-ep}
The second alternative way to prove a uniform error bound analogous to Corollary \ref{coro:MMD} is to decompose the difference $\widehat{\gamma}_{k,u}(h_\mathcal{X}(\mathbf{X}), h_\mathcal{Y}(\mathbf{Y})) - \mathbb{E}\left[\widehat{\gamma}_{k,u}(h_\mathcal{X}(\mathbf{X}), h_\mathcal{Y}(\mathbf{Y})\right]$ into a sum of terms capturing the maximal deviation of the empirical mean from the true mean, mirroring the left-hand side of \eqref{eq:bartlett}.
Here, we consider the unbiased estimator $\widehat{\gamma}_{k,u}$ (Definition \ref{def:empiricalMMDestimators}) as the empirical estimator of $\gamma_k$.
First, based on the fact that $\mathbb{E}\widehat{\gamma}_{k,u} = \gamma_k$, we have
\begin{align*}
    \widehat{\gamma}_{k,u}(h_\mathcal{X}(\mathbf{X}), h_\mathcal{Y}(\mathbf{Y})) - \mathbb{E}&\left[\widehat{\gamma}_{k,u}(h_\mathcal{X}(\mathbf{X}), h_\mathcal{Y}(\mathbf{Y})\right] =\\
    & \underbrace{\frac{1}{m(m-1)}\sum_{i\neq j}^m \left[k(h_\mathcal{X}(X_i), h_\mathcal{X}(X_j))
    - \mathbb{E}k(h_\mathcal{X}(X), h_\mathcal{X}(X'))\right]}_{(\star_{x,x'})}\\ 
    &+ \underbrace{\frac{1}{n(n-1)}\sum_{i\neq j}^n \left[k(h_\mathcal{Y}(Y_i), h_\mathcal{Y}(Y_j)) - \mathbb{E}k(h_\mathcal{Y}(Y), h_\mathcal{Y}(Y'))\right]}_{(\star_{y,y'})}\\
    &- \underbrace{\frac{2}{mn}\sum_{i=1}^m\sum_{j=1}^n \left[k(h_\mathcal{X}(X_i), h_\mathcal{Y}(Y_j)) - \mathbb{E}k(h_\mathcal{X}(X), h_\mathcal{Y}(Y))\right]}_{(\star_{x,y})}.
\end{align*}
Utilize Hoeffding's decomposition, the terms $(\star_{x,x'})$ and $(\star_{y,y'})$ can be considered as a summation of degenerate U-statistics.
In the following, we provide the decomposition of $(\star_{x,x'})$ as a representative example:
\begin{align*}
    (\star_{x,x'}) = \frac{1}{m} \sum_{i=1}^m g_1\left(h_\mathcal{X}(X_i)\right)+\frac{1}{m(m-1)} \sum_{1 \leq i<j \leq n} g_2\left(h_\mathcal{X}(X_i), h_\mathcal{X}(X_j)\right),
\end{align*}
where $g_1$ and $g_2$ are functions defined as follows:
\begin{align*}
    &g_1(h_\mathcal{X}(x)):=\mathbb{E}_X k(h_\mathcal{X}(x), h_\mathcal{X}(X)) - \mathbb{E}\left[k(h_\mathcal{X}(X), h_\mathcal{X}(X'))\right],\\
    &g_2(h_\mathcal{X}(x), h_\mathcal{X}(x')):=k(h_\mathcal{X}(x), h_\mathcal{X}(x')) - g_1(x) - g_1(x') - \mathbb{E}\left[k(h_\mathcal{X}(X), h_\mathcal{X}(X'))\right].
\end{align*}
Note that the term $\sup_{h_\mathcal{X}}g_1(h_\mathcal{X}(x))$ can be upper bounded utilizing the empirical process theory or Maurer's uniform concentration inequality (Lemma \ref{lem:maurer}).
Regarding the second order U-statistic term $\frac{1}{m(m-1)} \sum_{1 \leq i<j \leq n} g_2\left(h_\mathcal{X}(X_i), h_\mathcal{X}(X_j)\right)$, utilizing the decoupling theory for U-statistic \cite[Theorem 3.1.2]{de2012decoupling}, under the assumption that $\sup_{u} k(u,u) \leq \nu$, the following conclusion can be derived:
\begin{align*}
    \mathbb{E}\sup_{h_\mathcal{X}}\left| \sum_{1 \leq i<j \leq n} g_2\left(h_\mathcal{X}(X_i), h_\mathcal{X}(X_j)\right)\right| \leq 12 \mathbb{E}\sup_{h_\mathcal{X}}\left| \sum_{1 \leq i<j \leq n} g_2\left(h_\mathcal{X}(X_i), h_\mathcal{X}(X_j')\right)\right|,
\end{align*}
where $X_1', \dots, X_m'$ are independent copies of $X_1, \dots, X_m$, respectively.

To summarize, we have the following derivation:
\begin{eqnarray*}
    &&\mathbb{E}\sup_h \left|\widehat{\gamma}_{k,u}(h_\mathcal{X}(\mathbf{X}), h_\mathcal{Y}(\mathbf{Y})) - \mathbb{E}\left[\widehat{\gamma}_{k,u}(h_\mathcal{X}(\mathbf{X}), h_\mathcal{Y}(\mathbf{Y})\right] \right|\\
    &=& \mathbb{E}\sup_{h} \left|(\star_{x,x'}) + (\star_{y,y'}) + (\star_{x,y})\right|\\
    &\leq& \mathbb{E}\sup_{h_\mathcal{X}} \left|(\star_{x,x'})\right| + \mathbb{E}\sup_{h_\mathcal{Y}}\left|(\star_{y,y'})\right| + \mathbb{E}\sup_h\left|(\star_{x,y})\right|.
\end{eqnarray*}
Here, the term $(\star_{x,x'})$ can be upper bounded via the following summation:
\begin{eqnarray*}
    && \mathbb{E}\sup_{h} \left|(\star_{x,x'})\right|\\
    &=& \mathbb{E}\sup_{h_\mathcal{X}} \left|\frac{1}{m} \sum_{i=1}^m g_1\left(h_\mathcal{X}(X_i)\right)+\frac{1}{m(m-1)} \sum_{1 \leq i<j \leq n} g_2\left(h_\mathcal{X}(X_i), h_\mathcal{X}(X_j)\right)\right|\\
    &\leq& \frac{1}{m(m-1)}\mathbb{E}\sup_{h_\mathcal{X}}\left| \sum_{1 \leq i<j \leq n} g_2\left(h_\mathcal{X}(X_i), h_\mathcal{X}(X_j)\right)\right| + \frac{1}{m}\mathbb{E}\sup_{h_\mathcal{X}}\left|\sum_{i=1}^m g_1(h_\mathcal{X}(X_i))\right|\\
    &\leq& \frac{12}{m(m-1)}\mathbb{E}\sup_{h_\mathcal{X}}\left| \sum_{1 \leq i<j \leq n} g_2\left(h_\mathcal{X}(X_i), h_\mathcal{X}(X_j')\right)\right| + \frac{1}{m}\mathbb{E}\sup_{h_\mathcal{X}}\left|\sum_{i=1}^m g_1(h_\mathcal{X}(X_i))\right|,
\end{eqnarray*}
where both terms $\mathbb{E} \sup _{h_{\mathcal{X}}}\left|\frac{\sum_{i=1}^m g_1\left(h_{\mathcal{X}}\left(X_i\right)\right)}{m}\right|$ and $\mathbb{E} \sup _{h_{\mathcal{X}}}\left|\frac{\sum_{1 \leq i<j \leq m} g_2\left(h_{\mathcal{X}}\left(X_i\right), h_{\mathcal{X}}\left(X_j^{\prime}\right)\right)}{m(m-1)}\right|$ represent the supremum of sample averages. 
A similar decomposition applies to $\left(\star_{y, y^{\prime}}\right)$.

Consequently, the following summation can be upper bounded by the summation of terms considered as the suprema of sample averages over specified function sets:
$$
\mathbb{E} \sup _{h_{\mathcal{X}}}\left|\left(\star_{x, x^{\prime}}\right)\right|+\mathbb{E} \sup _{h_{\mathcal{Y}}}\left|\left(\star_{y, y^{\prime}}\right)\right|+\mathbb{E} \sup _h\left|\left(\star_{x, y}\right)\right|.
$$
Following the above derivation steps, the uniform error bounds for kernel-based statistics (Corollaries \ref{coro:MMD}, \ref{coro:HSIC}, \ref{cor:ED}, and \ref{cor:dCov}) can be obtained by adopting uniform concentration inequalities for one-sample statistics (Lemma \ref{thm:bartlett}, \ref{lem:maurer}) or maximal inequalities for empirical processes (\cite{pollard1989asymptotics}).

Nevertheless, we choose to build a general inequality as presented in Theorem \ref{thm:general}, aiming to focus on a broader range of possible involved statistics, with the kernel-based statistics considered as special cases.
Moreover, in the machine learning context, following the paths outlined above will not lead to upper bounds better than the ones presented from Corollary \ref{coro:MMD} to Corollary \ref{cor:dCov}.
Specifically, if we utilize the above derivation steps and choose to employ either Bartlett's inequality (Lemma \ref{thm:bartlett}) or maximal inequality for the empirical process to build the uniform concentration inequalities for kernel-based statistics, the following points then arise:
\begin{enumerate}
    \item \textit{Bartlett's Inequality (Theorem \ref{thm:bartlett}).}
    The resulting bounds closely mirror those from Corollary \ref{coro:MMD} to Corollary \ref{cor:dCov}, yielding an upper bound of order $\mathcal{O}\left(m^{-1 / 2}+n^{-1 / 2}\right)$. 
    These bounds are derived using Rademacher complexity, which is equivalent to Gaussian complexity up to a logarithm factor (Proposition \ref{def:complexity}).
    \item \textit{Maximal Inequalities for Empirical Processes.}
    Existing inequalities such as Hoeffding’s \cite[Theorem 12.1]{10.1093/acprof:oso/9780199535255.001.0001}, Bernstein’s \cite[Theorem 12.2]{10.1093/acprof:oso/9780199535255.001.0001}, Bousquet’s \cite[Theorem 12.5]{10.1093/acprof:oso/9780199535255.001.0001}, and the Klein–Rio bound \cite[Theorem 12.10]{10.1093/acprof:oso/9780199535255.001.0001} build their upper bounds by leveraging the variance of the empirical process.
    Even if the formulations of these inequalities are simpler, they tend to be less adaptive than Rademacher-based or Gaussian-complexity-based bounds, which can better capture the actual data or distribution (Proposition \ref{prop:Gaussian}).
\end{enumerate}
Most importantly, our approach is more systematic, while in the above approach, one might need to work out details on each individual case.

\section{Conclusion}
\label{sec:conclude}
There is a widespread need in modern statistical and machine learning methods to optimize objectives based on the distance between two probability distributions. 
This paper establishes a uniform concentration inequality for two-sample statistics (Theorem \ref{thm:general}), which are used to estimate the distance between two probability distributions.
As discussed in Sections \ref{sec:kernel-based} and \ref{sec:consistency_empirical}, our derived concentration bound provides a brand-new general theoretical tool for establishing statistical consistency in the methods mentioned above, filling the theoretical gaps that occurred in the existing literature.


\acks{
The authors are partially sponsored by a subcontract of NSF grant 2229876, the A. Russell Chandler III Professorship at Georgia Institute of Technology, and the NIH-sponsored Georgia Clinical \& Translational Science Alliance project.}


\newpage

\appendix
\section{Technical Proofs}
\label{sec:proofs}
This section contains all the relegated technical proofs. 
Section \ref{sec:preliminary2} provides several preliminary results.
From Section \ref{sec:proof-1} to Section \ref{sec:proof-cor6.6}, we then present, in sequence, the proofs of Theorem \ref{thm:general}, Proposition \ref{prop:Gaussian}, Lemma \ref{lem:contraction}, Proposition \ref{prop:linear}, Proposition \ref{prop:fnn}, Corollary \ref{coro:MMD}, Corollary \ref{coro:HSIC}, Lemma \ref{ex:dk}, Lemma \ref{ex:bilkernel}, Lemma \ref{ex:polykernel}, Lemma \ref{lem:sigmoid}, Lemma \ref{lem:translation}, Lemma \ref{ex:Gk}, Lemma \ref{ex:Lk}, Corollary \ref{ex:DISCA}, Corollary \ref{ex:ICA}, Corollary \ref{cor:generalizedMMD}, Corollary \ref{coro:mmd(m,n)}, Corollary \ref{coro:MMDGAN}, and Proposition \ref{prop:alternative-1}.

\subsection{Preliminary results}
\label{sec:preliminary2}
As a preliminary, we provide the following proposition.
\begin{proposition}
\label{prop:intro}
    $\forall f_1, f_2 \in \{f: \Theta \mapsto \mathbb{R}\}$, we have
    \begin{align*}
        \left|\sup_{\theta \in \Theta}f_1(\theta)-\sup_{\theta \in \Theta}f_2(\theta)\right| \leq& \sup_{\theta \in \Theta}\left|f_1(\theta)-f_2(\theta)\right|, \quad \text{and}\\
        \left|\inf_{\theta \in \Theta}f_1(\theta)-\inf_{\theta \in \Theta}f_2(\theta)\right| \leq& \sup_{\theta \in \Theta}\left|f_1(\theta)-f_2(\theta)\right|.
    \end{align*}
\end{proposition}
\begin{proof}
    The proof begins with the upper bound of $\sup_{\theta \in \Theta}f_1(\theta) - \sup_{\theta \in \Theta}f_2(\theta)$.
    First, we have $\sup_{\theta \in \Theta}f_1(\theta) = \sup_{\theta \in \Theta}\left[f_2(\theta) + \left(f_1(\theta)-f_2(\theta)\right)\right]$. As a consequence, the following inequalities are derived.
    \begin{align*}
        &\sup_{\theta \in \Theta}f_1(\theta) - \sup_{\theta \in \Theta}f_2(\theta)\\
        \leq & \sup_{\theta \in \Theta}\left[f_2(\theta) + \left|f_1(\theta)-f_2(\theta)\right|\right]- \sup_{\theta \in \Theta}f_2(\theta)\\
        \leq & \sup_{\theta \in \Theta}f_2(\theta) + \sup_{\theta \in \Theta}\left|f_1(\theta)-f_2(\theta)\right| - \sup_{\theta \in \Theta}f_2(\theta)\\
        =& \sup_{\theta \in \Theta}\left|f_1(\theta)-f_2(\theta)\right|.
    \end{align*}
    Exchange $f_1$ and $f_2$, a similar inequality is derived. As a consequence, we have 
    $$
        \left|\sup_{\theta \in \Theta}f_1(\theta) - \sup_{\theta \in \Theta}f_2(\theta)\right| \leq \sup_{\theta \in \Theta}|f_1(\theta) - f_2(\theta)|.
    $$
    
    For the upper bound of $\inf_{\theta \in \Theta}f_1(\theta) - \inf_{\theta \in \Theta}f_2(\theta)$, let $\{\theta_i\}_{i=1}^{\infty}$ be a sequence in $\Theta$, such that $\lim_{i \rightarrow \infty} f_2(\theta_i) = \inf_{\theta \in \Theta}f_2(\theta)$. That is, $\forall \epsilon > 0$, $\exists N(\epsilon) \in \mathbb{N}_+$, such that $\forall i > N(\epsilon)$, we have $f_2(\theta_i) < \inf_{\theta \in \Theta}f_2(\theta) + \epsilon$. More specifically, let $i_\epsilon$ be any index greater than $N(\epsilon)$, then we have
    \begin{align*}
        &\inf_{\theta \in \Theta}f_1(\theta) - \inf_{\theta \in \Theta}f_2(\theta)\\
        \leq & \inf_{\theta \in \Theta}\left[f_2(\theta) + \left|f_1(\theta)-f_2(\theta)\right|\right]- \inf_{\theta \in \Theta}f_2(\theta)\\
        \leq & f_2(\theta_{i_\epsilon}) + \left|f_1(\theta_{i_\epsilon})-f_2(\theta_{i_\epsilon})\right| - \inf_{\theta \in \Theta}f_2(\theta)\\
        \leq & \inf_{\theta \in \Theta}f_2(\theta) + \epsilon + \sup_{\theta \in \Theta}\left|f_1(\theta)-f_2(\theta)\right| - \inf_{\theta \in \Theta}f_2(\theta)\\
        = & \epsilon + \sup_{\theta \in \Theta}\left|f_1(\theta)-f_2(\theta)\right|.
    \end{align*}
    Since the above inequality holds for any $\epsilon > 0$, let $\epsilon \rightarrow 0$, we have
    $$
        \inf_{\theta \in \Theta}f_1(\theta) - \inf_{\theta \in \Theta}f_2(\theta) \leq \sup_{\theta \in \Theta}\left|f_1(\theta)-f_2(\theta)\right|.
    $$
    Exchange $f_1$ and $f_2$, a similar inequality is derived. As a consequence, we have 
    $$
        \left|\inf_{\theta \in \Theta}f_1(\theta) - \inf_{\theta \in \Theta}f_2(\theta)\right| \leq \sup_{\theta \in \Theta}|f_1(\theta) - f_2(\theta)|.
    $$
\end{proof}

\subsection{Proof of Theorem \ref{thm:general}}
\label{sec:proof-1}
\begin{proof}
To begin with, we consider the following two lemmas proved in \cite{maurer2019uniform}.
\begin{lemma}[\cite{maurer2019uniform}, Lemma 9]
\label{lem:lem9_maurer}
    Let $\mathbf{x} := (x_1, \dots, x_n)$, and $\mathbf{x}^{\prime}:= (x_1', \dots, x_n') \in \mathcal{U}^n$.
    Then we have
    $$
        f(\mathbf{x})-f\left(\mathbf{x}^{\prime}\right)=\sum_{k=1}^n F_k\left(\mathbf{x}, \mathbf{x}^{\prime}\right),
    $$
    where for $k = 1, \dots, n$, we have
    \begin{equation}
    \label{eq:F-K}
        F_k\left(\mathbf{x}, \mathbf{x}^{\prime}\right) = \frac{1}{2^k} \sum_{A \subseteq \{1, 2, \dots, k-1\}}\left(D_{x_k, x_k^{\prime}}^k f\left(\mathbf{x}^A\right)+D_{x_k, x_k^{\prime}}^k f\left(\mathbf{x}^{A^c}\right)\right).
    \end{equation}
    Here, $D^k_{x_k, x_k'}$ refers to the difference operator defined in Assumption \ref{assum:maurer}. We use $\mathbf{x}^A$ to denote the vector in $\mathcal{U}^n$ defined by
    $$
        x_i^A=\left\{\begin{array}{lll}
        x_i^{\prime}, & \text { if } & i \in A; \\
        x_i, & \text { if } & i \notin A.
        \end{array}\right.
    $$
    We use $A^c$ to denote $\{1, \dots, n\}\backslash A$.
    \end{lemma}

\begin{lemma}[\cite{maurer2019uniform}, Lemma 10] 
\label{lem:lem10_maurer} 
    For $\left(\mathbf{x}, \mathbf{x}^{\prime}\right),\left(\mathbf{y}, \mathbf{y}^{\prime}\right) \in \mathcal{U}^{2 n}$ and $k \in \{1, \dots, n\}$, we have
    $$
        F_k\left(\mathbf{x}, \mathbf{x}^{\prime}\right)-F_k\left(\mathbf{y}, \mathbf{y}^{\prime}\right) \leq \sqrt{\frac{\pi}{2}} \mathbb{E}\left|\left\langle\boldsymbol{\xi}_k, v_k\left(\mathbf{x}, \mathbf{x}^{\prime}\right)-v_k\left(\mathbf{y}, \mathbf{y}^{\prime}\right)\right\rangle\right|,
    $$
    where $\boldsymbol{\xi}_k=\left(\xi_{k 1}, \ldots, \xi_{k n}, \xi_{k 1}^{\prime}, \ldots, \xi_{k n}^{\prime}\right) \stackrel{\text{i.i.d.}}{\sim} \mathcal{N}(\boldsymbol{0}_{2n}, I_{2n})$.
    For $\left(\mathbf{x}, \mathbf{x}^{\prime}\right) \in \mathcal{U}^{2 n}$ and $k = 1, \dots, n$, we define the vector $v_k\left(\mathbf{x}, \mathbf{x}^{\prime}\right) \in \mathcal{U}^{2 n}$ by
    $$
    \begin{aligned}
    &v_i^k\left(\mathbf{x}, \mathbf{x}^{\prime}\right)=\left\{\begin{array}{ll}
    2 M x_k & \text { if } \quad i=k \\
    J n^{-1 / 2} x_i & \text { if } \quad i \neq k, i \leq n \\
    2 M x_k^{\prime} & \text { if } \quad i=n+k \\
    J n^{-1 / 2} x_{i-n}^{\prime} & \text { if } \quad i \neq n+k, i>n
    \end{array} .\right.
    \end{aligned}
    $$
\end{lemma}
Based on the above lemmas, the following lemma can be proved through mathematical induction.
\begin{lemma}
\label{lem:maurer_twosample}
    Let $\mathbf{X}:=(X_1, \dots, X_m)^T$ and $\mathbf{Y}:=(Y_1, \dots, Y_n)^T$ be the data matrices of random vectors $X$ and $Y$, respectively, where $X_i \stackrel{\text{i.i.d.}}{\sim} X$ and $Y_j \stackrel{\text{i.i.d.}}{\sim} Y$, $\forall i, j$. Let $\mathbf{X}'$ and $\mathbf{Y}'$ be independent copies of $\mathbf{X}$ and $\mathbf{Y}$, respectively. We have
    \begin{align}
    \label{eq:pfThm2_maurer}
        \mathbb{E} \sup_{h \in \mathcal{H}} f(h(\mathbf{X}, \mathbf{Y})) -f\left(h\left(\mathbf{X}^{\prime}, \mathbf{Y}'\right)\right) \leq \sqrt{\frac{\pi}{2}} \mathbb{E} \sup_{h \in \mathcal{H}} \sum_{k=1}^{m+n}\left\langle\boldsymbol{\xi}_k, v_k\left(h(\mathbf{X}, \mathbf{Y}), h\left(\mathbf{X}^{\prime}, \mathbf{Y}^{\prime}\right)\right)\right\rangle,
    \end{align}
    where $\boldsymbol{\xi}_k \sim \mathcal{N}(\boldsymbol{0}_{2d(m+n)}, I_{2d(m+n)})$.
    Here, $v_k\left(h(\mathbf{X}, \mathbf{Y}), h(\mathbf{X}', \mathbf{Y}')\right)$ is defined as the composition of terms $v_k(h(\mathbf{X}, \mathbf{Y}))$ and $v_k(h(\mathbf{X}', \mathbf{Y}'))$, $\forall k$. 
    For $k=1, \dots, m$, the term $v_k$ is defined as follows:
    \begin{align*}
        v_k&\left(h(\mathbf{X}, \mathbf{Y})\right) = \frac{J_{\text{Lip}}(f)}{(m+n)\sqrt{m+n-1}}\bigg(h_\mathcal{X}(X_1), \dots, h_\mathcal{X}(X_{k-1}),\\
        &\frac{2M_{\text{Lip}}(f)\sqrt{m+n-1}}{J_{\text{Lip}}(f)}h_\mathcal{X}(X_{k}), h_\mathcal{X}(X_{k+1}),
        \dots,h_\mathcal{X}(X_{m}),h_\mathcal{Y}(Y_1), \dots, h_\mathcal{Y}(Y_n)\bigg).
    \end{align*}
    For $k = m+1,\dots, m+n$, the term $v_k$ is defined as follows:
    \begin{align*}
        v_k(h(\mathbf{X}, \mathbf{Y})) = \frac{J_{\text{Lip}}(f)}{(m+n)\sqrt{m+n-1}}\bigg(h_\mathcal{X}(X_1), \dots, h_\mathcal{X}(X_m), h_\mathcal{Y}(Y_1), \dots, h_\mathcal{X}(Y_{k-m-1}),\\
        \frac{2M_{\text{Lip}}(f)\sqrt{m+n-1}}{J_{\text{Lip}}(f)}h_\mathcal{Y}(Y_{k-m}), h_\mathcal{Y}(Y_{k-m+1}), \dots, h_\mathcal{Y}(Y_{n})\bigg).
    \end{align*}
\end{lemma}
\begin{proof}
To prove Lemma \ref{lem:maurer_twosample}, we consider induction on $l \in \{0, \dots, m+n\}$ regarding the following inequality:
\begin{equation}
\label{eq:induction}
    \begin{aligned}
    &\mathbb{E} \sup_{h \in \mathcal{H}} f(h(\mathbf{X}, \mathbf{Y}))-f\left(h\left(\mathbf{X}^{\prime}, \mathbf{Y}'\right)\right) \\
    \leq & \mathbb{E} \sup_{h \in \mathcal{H}} \left[\sqrt{\frac{\pi}{2}} \sum_{k=1}^{l}\left\langle\boldsymbol{\xi}_k, v_k\left(h(\mathbf{X}, \mathbf{Y}), h\left(\mathbf{X}^{\prime}, \mathbf{Y}^{\prime}\right)\right)\right\rangle + \sum_{k=l+1}^{m+n}  F_k(h(\mathbf{X}, \mathbf{Y}), h(\mathbf{X'}, \mathbf{Y'}))\right].
    \end{aligned}
\end{equation}
Recall that function $F_k$ was defined in \eqref{eq:F-K}. 
For $l = 0$, the formulation of the above inequality is equivalent to the one derived in Lemma \ref{lem:lem9_maurer}.
Suppose the above inequality holds for $l-1$.
For simplicity, given $h \in \mathcal{H}$, we define a real valued random variable $R_h$ as follows:
\begin{align*}
    R_h = \sup_{h \in \mathcal{H}} \sqrt{\frac{\pi}{2}} \sum_{k=1}^{l-1}\left\langle\boldsymbol{\xi}_k, v_k\left(h(\mathbf{X}, \mathbf{Y}), h\left(\mathbf{X}^{\prime}, \mathbf{Y}^{\prime}\right)\right)\right\rangle + \sum_{k=l+1}^{m+n}  F_k(h(\mathbf{X}, \mathbf{Y}), h(\mathbf{X'}, \mathbf{Y'})).
\end{align*}
That is, we have
\begin{align*}
    \mathbb{E} \sup_{h \in \mathcal{H}} f(h(\mathbf{X}, \mathbf{Y}))-f\left(h\left(\mathbf{X}^{\prime}, \mathbf{Y}'\right)\right) \leq \mathbb{E}\sup_{h \in \mathcal{H}} \left[F_l(h(\mathbf{X}, \mathbf{Y}), h(\mathbf{X'}, \mathbf{Y'})) + R_h\right].
\end{align*}
Without loss of generality, we suppose the $l$-th position of vector $h(\mathbf{X}, \mathbf{Y})$ is $h_\mathcal{X}(X_l)$.
If we exchange the positions of $X_l$ and $X_l'$, then the $F_l$ term in the above formulation would be given in the following formula:
\begin{align*}
    \frac{1}{2^k} \sum_{A \subseteq \{1, 2, \dots, k-1\}}\left(D_{X_l', X_l}^l f\left(h(\mathbf{X}, \mathbf{Y})^A\right)+D_{X_l', X_l}^l f\left(h(\mathbf{X}, \mathbf{Y})^{A^c}\right)\right),
\end{align*}
which is equal to $- F_l(h(\mathbf{X}, \mathbf{Y}), h(\mathbf{X}', \mathbf{Y}'))$.

Note that $\boldsymbol{\xi}_k = (\xi_{k,1}, \dots, \xi_{k,m+n}, \xi_{k,1}', \dots, \xi_{k,m+n}')$. 
Consequently, from the definition of $v_k$, $\forall k$ as given in Lemma \ref{lem:maurer_twosample}, by exchanging both $X_l, X_l'$ and $\boldsymbol{\xi}_{k, l}'$, $\forall k$, the summation $\sum_{k=1}^{l-1}\left\langle\boldsymbol{\xi}_k, v_k\left(h(\mathbf{X}, \mathbf{Y}), g\left(\mathbf{X}^{\prime}, \mathbf{Y}^{\prime}\right)\right)\right\rangle$ remains unchanged.
Moreover, for $k = l+1, \dots, m+n$, denote $D^k_{X_k, X_k'}f(h(\mathbf{X}, \mathbf{Y})^A)$ as $\varphi(A)$, then we have:
\begin{align*}
    &F_k((\mathbf{X}, \mathbf{Y}), (\mathbf{X}', \mathbf{Y}'))\\ 
    =& \frac{1}{2^k}\sum_{A \subset \{1, \dots, k-1\}}\left[\varphi(A) + \varphi(A^c)\right]\\
    =& \frac{1}{2^k}\sum_{A \subset \{1, \dots, k-1\}}\left[\varphi(A) + \varphi(A \cup \{k, \dots, n\})\right]\\
    =& \frac{1}{2^k}\sum_{A \subset \{1, \dots, k-1\}\backslash {l}}\left[\varphi(A) + \varphi(A \cup \{l\}) + \varphi(A \cup \{k, \dots, n\}) + \varphi(A \cup \{k, \dots, n\} \cup \{l\})\right].
\end{align*}
After exchanging $X_l$ and $X_l'$, for $A \subset \{1, \dots, k-1\}\backslash {l}$, the positions of $\varphi(A)$ and $\varphi(A \cup \{l\})$ will be exchanged. Similarly, the positions of $\varphi(A \cup \{k, \dots, n\})$ and $\varphi(A \cup \{k, \dots, n\} \cup \{l\})$ will be exchanged.
That is, $F_k((\mathbf{X}, \mathbf{Y}), (\mathbf{X}', \mathbf{Y}'))$ remains unchanged for $k = l+1, \dots, m+n$.
This leads to the conclusion that the random variable $R_g$ will remain unchanged after we exchange the position of both $X_l, X_l'$ and $\boldsymbol{\xi}_{k, l}'$, $\forall k$.
Since the expectation would be unchanged after the exchange operation, we have
\begin{align*}
    \mathbb{E}\sup_{h \in \mathcal{H}} \left[F_l(h(\mathbf{X}, \mathbf{Y}), h(\mathbf{X'}, \mathbf{Y'})) + R_h\right]
    = \mathbb{E}\sup_{h \in \mathcal{H}} \left[-F_l(h(\mathbf{X}, \mathbf{Y}), h(\mathbf{X'}, \mathbf{Y'})) + R_{h}\right].
\end{align*}
Consequently, term $\mathbb{E}\sup_{h \in \mathcal{H}} \left[F_l(h(\mathbf{X}, \mathbf{Y}), h(\mathbf{X'}, \mathbf{Y'})) + R_h\right]$ can be bounded as follows:
$$
    \frac{1}{2}\mathbb{E}\left[\sup_{h_1} \left[F_l(h_1(\mathbf{X}, \mathbf{Y}), h_1(\mathbf{X'}, \mathbf{Y'})) + R_{h_1}\right] + \sup_{h_2} \left[-F_l(h_2(\mathbf{X}, \mathbf{Y}), h_2(\mathbf{X'}, \mathbf{Y'})) + R_{h_2}\right]\right].
$$
Regarding the above expression, it is equivalent to the following expression:
$$
    \frac{1}{2}\mathbb{E}\sup_{h_1, h_2 \in \mathcal{H}} \left[F_l(h_1(\mathbf{X}, \mathbf{Y}), h_1(\mathbf{X'}, \mathbf{Y'})) -F_l(h_2(\mathbf{X}, \mathbf{Y}), h_2(\mathbf{X'}, \mathbf{Y'})) + R_{h_1} + R_{h_2}\right].
$$
This is no greater than the following one equipped with an absolute value:
$$
    \frac{1}{2}\mathbb{E}\sup_{h_1, h_2 \in \mathcal{H}} \left[\left|F_l(h_1(\mathbf{X}, \mathbf{Y}), h_1(\mathbf{X'}, \mathbf{Y'})) -F_l(h_2(\mathbf{X}, \mathbf{Y}), h_2(\mathbf{X'}, \mathbf{Y'}))\right| + R_{h_1} + R_{h_2}\right].
$$
Utilize Lemma \ref{lem:lem10_maurer}, the above quantity is smaller than or equal to the following expression:
$$
    \frac{1}{2}\mathbb{E} \sup_{h_1, h_2} \left[\sqrt{\frac{\pi}{2}} \mathbb{E}_{\boldsymbol{\xi}_l}\left|(\star)\right| + R_{h_1} + R_{h_2}\right],
$$
where $(\star):= \left\langle\boldsymbol{\xi}_l, v_l\left(h_1(\mathbf{X}, \mathbf{Y}), h_1(\mathbf{X}^{\prime}, \mathbf{Y}')\right)-v_l\left(h_2(\mathbf{X}, \mathbf{Y}), h_2(\mathbf{X}^{\prime}, \mathbf{Y}')\right)\right\rangle$.
Next, notice that we have $\sup_{h_1, h_2}\mathbb{E}_{\boldsymbol{\xi}_l}|(\star)| \leq \mathbb{E}_{\boldsymbol{\xi}_l}\sup_{h_1, h_2}|(\star)|$, that is, the expectation symbol $|(\star)|$ can be moved outside the bracket, which is given as follows:
\begin{eqnarray*}
    \frac{1}{2}\mathbb{E}\sup_{h_1, h_2} \left[\sqrt{\frac{\pi}{2}} \left|(\star)\right| + R_{h_1} + R_{h_2}\right].
\end{eqnarray*}
Here, if $(\star) < 0$, we can exchange $h_1$ and $h_2$, making it greater than zero, while $R_{h_1} + R_{h_2}$ remains unchanged.
Consequently, we have:
\begin{align*}
    &\mathbb{E} \sup_{h \in \mathcal{H}} f(h(\mathbf{X}, \mathbf{Y}))-f\left(h\left(\mathbf{X}^{\prime}, \mathbf{Y}'\right)\right)\\
    \leq & \frac{1}{2}\mathbb{E}\sup_{h_1, h_2 \in \mathcal{H}} \left[\sqrt{\frac{\pi}{2}}\left|(\star)\right| + R_{h_1} + R_{h_2}\right]\\
    = & \frac{1}{2}\mathbb{E}\sup_{h_1, h_2 \in \mathcal{H}} \left[\sqrt{\frac{\pi}{2}}(\star) + R_{h_1} + R_{h_2}\right]\\
    =& \frac{1}{2}\mathbb{E}\sup_{h_1}\left[ \sqrt{\frac{\pi}{2}}\left\langle\boldsymbol{\xi}_l, v_l\left(h_1(\mathbf{X}, \mathbf{Y}), h_1(\mathbf{X}^{\prime}, \mathbf{Y}')\right)\right\rangle+ R_{h_1}\right] + \\
    &\quad \quad \quad \quad \quad \quad \quad \frac{1}{2}\mathbb{E}\sup_{h_2}\left[-\sqrt{\frac{\pi}{2}}\left\langle \boldsymbol{\xi}_l, v_l\left(h_2(\mathbf{X}, \mathbf{Y}), h_2(\mathbf{X}^{\prime}, \mathbf{Y}')\right)\right\rangle + R_{h_2}\right].
\end{align*}
Based on the symmetry property of Gaussian distribution, we have
\begin{align*}
    & \mathbb{E}\sup_{h_2}\left[-\sqrt{\frac{\pi}{2}}\left\langle \boldsymbol{\xi}_l, v_l\left(h_2(\mathbf{X}, \mathbf{Y}), h_2(\mathbf{X}^{\prime}, \mathbf{Y}')\right)\right\rangle + R_{h_2}\right]\\ 
    = & \mathbb{E}\sup_{h_2}\left[\sqrt{\frac{\pi}{2}}\left\langle \boldsymbol{\xi}_l, v_l\left(h_2(\mathbf{X}, \mathbf{Y}), h_2(\mathbf{X}^{\prime}, \mathbf{Y}')\right)\right\rangle + R_{h_2}\right].
\end{align*}
Consequently, combining the above results, we have
\begin{align*}
    &\mathbb{E} \sup_{h \in \mathcal{H}} f(h(\mathbf{X}, \mathbf{Y}))-f\left(h\left(\mathbf{X}^{\prime}, \mathbf{Y}'\right)\right)\\
    \leq &\mathbb{E}\sup_{h \in \mathcal{H}}\left[ \sqrt{\frac{\pi}{2}}\left\langle\boldsymbol{\xi}_l, v_l\left(h(\mathbf{X}, \mathbf{Y}), h(\mathbf{X}^{\prime}, \mathbf{Y}')\right)\right\rangle+ R_{g}\right].
\end{align*}
That is, \eqref{eq:induction} holds for $l$. This completes the mathematical induction.
\end{proof}

Return to the proof of Theorem \ref{thm:general}.
Note that the following summation is equivalent to the term $\mathbb{E}\sup_{h \in \mathcal{H}}\sum_{k=1}^{m+n}\left\langle\boldsymbol{\xi}_k, v_k\left(h(\mathbf{X}, \mathbf{Y}), h\left(\mathbf{X}^{\prime}, \mathbf{Y}^{\prime}\right)\right)\right\rangle$:
\begin{align*}
    \mathbb{E}\sup_{h \in \mathcal{H}}\sum_{k=1}^{m+n}\bigg[\langle(\xi_{k, 1},  \dots, \xi_{k, m+n}), v_k(h(\mathbf{X},& \mathbf{Y}))\rangle +\\
    &\langle(\xi_{k, m+n+1}, \dots, \xi_{k, 2(m+n)}), v_k\left(h(\mathbf{X}', \mathbf{Y}')\right)\rangle\bigg].
\end{align*}
Since $\mathbf{X}'$ and $\mathbf{Y}'$ are independent copies of data matrices $\mathbf{X}$ and $\mathbf{Y}$, respectively, the above summation is smaller than or equal to the following expression:
\begin{align*}
    2\mathbb{E}\sup_{h \in \mathcal{H}}\sum_{k=1}^{m+n}\left\langle(\xi_{k, 1}, \dots, \xi_{k, m+n}), v_k\left(h(\mathbf{X}, \mathbf{Y})\right)\right\rangle.
\end{align*}
Recall notation $v_k$ given in Lemma \ref{lem:maurer_twosample}, let $\xi_{k, i} \stackrel{\text{i.i.d.}}{\sim} \mathcal{N}(\boldsymbol{0}_d, I_d)$, $\forall k, i$, we have
\begin{eqnarray}
    && 2\mathbb{E}\sup_{h \in \mathcal{H}}\sum_{k=1}^{m+n}\left\langle(\xi_{k, 1}, \dots, \xi_{k, m+n}), v_k\left(h(\mathbf{X}, \mathbf{Y})\right)\right\rangle = \nonumber\\
    && 2\mathbb{E}\sup_{h \in \mathcal{H}}\Bigg[\frac{2M_{\text{Lip}}(f)}{m+n} \underbrace{\left(\sum_{k=1}^{m} \xi_{k, k}^Th_\mathcal{X}(X_k) + \sum_{k=m+1}^{m+n} \xi_{k, k}^Th_\mathcal{Y}(Y_{k-m}) \right)}_{(\star_1)} \label{eq:star1}\\
    &+& \frac{J_{\text{Lip}}(f)}{(m+n)}\underbrace{\frac{1}{\sqrt{m+n-1}}\left(\sum_{k=1}^m \sum_{j \neq k}\xi_{k, j}^Th_\mathcal{X}(X_k) + \sum_{k=m+1}^{m+n} \sum_{j \neq k} \xi_{k, j}^Th_\mathcal{Y}(Y_{k-m})\right)}_{(\star_2)} \Bigg].\label{eq:star2}
\end{eqnarray}
That is, based on the definition of terms $(\star_1)$ and $(\star_2)$ given in \eqref{eq:star1} and \eqref{eq:star2}, we have
\begin{eqnarray*}
    && 2\mathbb{E}\sup_{h \in \mathcal{H}}\sum_{k=1}^{m+n}\left\langle(\xi_{k, 1}, \dots, \xi_{k, m+n}), v_k\left(h(\mathbf{X}, \mathbf{Y})\right)\right\rangle\\
    &\leq& 2\mathbb{E}\sup_{h \in \mathcal{H}}\Bigg[\frac{2M_{\text{Lip}}(f)}{m+n} (\star_1)\Bigg] + 2\mathbb{E} \sup_{h \in \mathcal{H}}\Bigg[\frac{J_{\text{Lip}}(f)}{(m+n)}(\star_2)\Bigg].
\end{eqnarray*}
Regarding the term $(\star_2)$, we have:
\begin{align*}
    (\star_2)
    \stackrel{\text{identical distribution}}{\sim} & \frac{1}{\sqrt{m+n-1}} \sum_{j=1}^{m+n-1} \left(\sum_{k=1,k\neq j}^m\xi_{k, j}^Th_\mathcal{X}(X_k) + \sum_{k=m+1,k\neq j}^{m+n} \xi_{k, j}^Th_\mathcal{Y}(Y_{k-m})\right)\\
    \stackrel{\text{identical distribution}}{\sim}& \sum_{k=1}^m\xi_{k, k}^Th_\mathcal{X}(X_k) + \sum_{k=m+1}^{m+n} \xi_{k, k}^Th_\mathcal{Y}(Y_{k-m}),
\end{align*}
where the last part is derived from the fact that $\forall 1\le k \le m$, random variable 
$$
\frac{1}{\sqrt{m+n-1}} \sum_{j=1,j\neq k}^{m+n} \xi_{k, j}^Th_\mathcal{X}(X_k)
$$ 
follows the same Guassian distribution as $\xi_{k, k}^Th_\mathcal{X}(X_k)$. 
For $m < k \le m+n$, a similar result is true to justify the second term in the above expression. 
Consequently, recall the definition of terms $(\star_1)$ and $(\star_2)$ given in \eqref{eq:star1} and \eqref{eq:star2}, respectively, we have
\begin{align*}
    & 2\mathbb{E}\sup_{h \in \mathcal{H}}\sum_{k=1}^{m+n}\left\langle(\xi_{k, 1}, \dots, \xi_{k, m+n}), v_k\left(h(\mathbf{X}, \mathbf{Y})\right)\right\rangle\\
    \leq& 2\mathbb{E}_{\mathbf{X}, \mathbf{Y}}\mathbb{E}_{\boldsymbol{\xi}}\sup_{h \in \mathcal{H}}\Bigg[\frac{2M_{\text{Lip}}(f)}{m+n}(\star_1)\Bigg] + 2\mathbb{E}_{\mathbf{X}, \mathbf{Y}}\mathbb{E}_{\boldsymbol{\xi}} \sup_{h \in \mathcal{H}}\Bigg[\frac{J_{\text{Lip}}(f)}{m+n}(\star_1)\Bigg].
\end{align*}
Recall the definition of the Gaussian complexity for $\mathcal{H}(\mathbf{X}, \mathbf{Y})=\{h(\mathbf{X}, \mathbf{Y})\mid h \in \mathcal{H}\}$, i.e., define $\boldsymbol{\xi}_i \stackrel{\text{i.i.d.}}{\sim} \mathcal{N}(\mathbf{0}, I_d), \forall i$, we have 
$$
    \mathcal{G}(\mathcal{H}(\mathbf{X}, \mathbf{Y})) = \mathbb{E}_{\boldsymbol{\xi}}\left[\sup_{h \in \mathcal{H}}\left(\sum_{k=1}^m\boldsymbol{\xi}_{k}^Th_\mathcal{X}(X_k) + \sum_{k=m+1}^{m+n} \boldsymbol{\xi}_{k}^Th_\mathcal{Y}(Y_{k-m})\right) \Bigg| (\mathbf{X}, \mathbf{Y})\right],
$$
which is equivalent with the term $(\star_1)$.
Consequently, combining the above results, we have
\begin{eqnarray}
    &&\mathbb{E}\sup_{h \in \mathcal{H}}\sum_{k=1}^{m+n}\left\langle\boldsymbol{\xi}_k, v_k\left(h(\mathbf{X}, \mathbf{Y}), h\left(\mathbf{X}^{\prime}, \mathbf{Y}^{\prime}\right)\right)\right\rangle \nonumber\\ 
    &\leq& 2\mathbb{E}_{\mathbf{X}, \mathbf{Y}}\left[\frac{2M_{\text{Lip}}(f)}{m+n}\mathcal{G}(\mathcal{H}(\mathbf{X}, \mathbf{Y}))\right] + 2\mathbb{E}_{\mathbf{X}, \mathbf{Y}}\left[\frac{J_{\text{Lip}}(f)}{m+n}\mathcal{G}(\mathcal{H}(\mathbf{X}, \mathbf{Y}))\right] \nonumber\\
    &=& \frac{2(2M_{\text{Lip}}(f) + J_{\text{Lip}}(f))}{m+n}\mathbb{E}[\mathcal{G}(\mathcal{H}(\mathbf{X}, \mathbf{Y}))]. \label{eq:p42_maurertwosample}
\end{eqnarray}
Note that the term $\mathbb{E} \sup _{h \in \mathcal{H}} f(h(\mathbf{X}, \mathbf{Y}))-\mathbb{E}f\left(h\left(\mathbf{X}, \mathbf{Y}\right)\right)$ is no greater than the expression $\mathbb{E} \sup _{h \in \mathcal{H}} f(h(\mathbf{X}, \mathbf{Y}))-f\left(h\left(\mathbf{X}^{\prime}, \mathbf{Y}'\right)\right)$ through the following derivation steps:
\begin{eqnarray}
    &&\mathbb{E} \sup _{h \in \mathcal{H}} f(h(\mathbf{X}, \mathbf{Y}))-\mathbb{E}f\left(h\left(\mathbf{X}, \mathbf{Y}\right)\right)\nonumber \\
    &=& \mathbb{E}_{\mathbf{X}, \mathbf{Y}} \sup _{h \in \mathcal{H}} f(h(\mathbf{X}, \mathbf{Y}))-\mathbb{E}_{\mathbf{X}', \mathbf{Y'}}f\left(h\left(\mathbf{X}', \mathbf{Y}'\right)\right)\nonumber \\
    &=& \mathbb{E}_{\mathbf{X}, \mathbf{Y}} \sup _{h \in \mathcal{H}} \mathbb{E}_{\mathbf{X}', \mathbf{Y'}} \left[f(h(\mathbf{X}, \mathbf{Y}))-f\left(h\left(\mathbf{X}', \mathbf{Y}'\right)\right)\right]\nonumber \\
    &\leq & \mathbb{E}_{\mathbf{X}, \mathbf{Y}} \mathbb{E}_{\mathbf{X}', \mathbf{Y'}} \sup _{h \in \mathcal{H}} \left[f(h(\mathbf{X}, \mathbf{Y}))-f\left(h\left(\mathbf{X}', \mathbf{Y}'\right)\right)\right]\nonumber \\
    &=& \mathbb{E} \sup _{h \in \mathcal{H}} \left[f(h(\mathbf{X}, \mathbf{Y}))-f\left(h\left(\mathbf{X}^{\prime}, \mathbf{Y}'\right)\right)\right]. \label{eq:p42-1}
\end{eqnarray}
In summary, combine \eqref{eq:p42-1}, \eqref{eq:p42_maurertwosample}, and Lemma \ref{lem:maurer_twosample}, we have
\begin{align*}
    \mathbb{E} \left[\sup _{h \in \mathcal{H}} f(h(\mathbf{X}, \mathbf{Y}))-\mathbb{E}f\left(h\left(\mathbf{X}', \mathbf{Y}'\right)\right)\right] \leq  \frac{\sqrt{2\pi}}{m + n}(2M_{\text{Lip}}(f) + J_{\text{Lip}}(f))\mathbb{E}[\mathcal{G}(\mathcal{H}(\mathbf{X}, \mathbf{Y}))].
\end{align*}
Moreover, based on McDiarmid's inequality, $\forall \delta \in (0, 1)$, with probability at least $1-\delta$, we have:
\begin{eqnarray}
&& \sup _{h \in \mathcal{H}} f(h(\mathbf{X}, \mathbf{Y}))-\mathbb{E}f\left(h\left(\mathbf{X}, \mathbf{Y}\right)\right) \nonumber  \\
& \leq & \frac{\sqrt{2\pi}}{m+n}(2M_{\text{Lip}}(f) + J_{\text{Lip}}(f))\mathbb{E}[\mathcal{G}(\mathcal{H}(\mathbf{X}, \mathbf{Y}))] + M(f)\sqrt{\frac{\ln(1/\delta)}{m+n}}. \label{eq:p42-2}
\end{eqnarray}
That is, Maurer's uniform concentration inequality (Lemma \ref{lem:maurer}) can be extended to the case where the involved statistic is a two-sample statistic.

Next, we consider to replace the difference $f(h(\mathbf{X}, \mathbf{Y}))-\mathbb{E}f\left(h\left(\mathbf{X}^{\prime}, \mathbf{Y}'\right)\right)$ in the above inequalities with its absolute value.
First, we replace the function $f$ in the above inequalties with $-f$.
Recall the seminorms mentioned in Assumption \ref{assum:maurer}, we have $M_{\text{Lip}}(-f) = M_{\text{Lip}}(f)$, $J_{\text{Lip}}(-f) = J_{\text{Lip}}(f)$, and $M(f) = M(-f)$.
That is, for any $\delta \in (0, 1)$, with probability at least $1-\delta$, we have
\begin{eqnarray*}
    &&\sup_{h \in \mathcal{H}}\mathbb{E}\left[f\left(h(\mathbf{X}', \mathbf{Y}')\right)\right] - f(h(\mathbf{X}, \mathbf{Y}))\\
    &=& \sup_{h \in \mathcal{H}} [-f(h(\mathbf{X}, \mathbf{Y}))] - \mathbb{E}\left[-f\left(h(\mathbf{X}', \mathbf{Y}')\right)\right]\\
    &\stackrel{\mbox{\eqref{eq:p42-1}, \eqref{eq:p42-2}}}{\leq} & \frac{\sqrt{2 \pi}}{m+n}\left(2 M_{\text{Lip}}(-f)+J_{\text{Lip}}(-f)\right)\mathbb{E}[\mathcal{G}(\mathcal{H}(\mathbf{X}, \mathbf{Y}))]+M(-f) \sqrt{\frac{\ln (1 / \delta)}{m+n}}\\
    &=& \frac{\sqrt{2 \pi}}{m+n}\left(2 M_{\text{Lip}}(f)+J_{\text{Lip}}(f)\right)\mathbb{E}[\mathcal{G}(\mathcal{H}(\mathbf{X}, \mathbf{Y}))]+M(f) \sqrt{\frac{\ln (1 / \delta)}{m+n}}.
\end{eqnarray*}
Consequently, combining the above inequality with \eqref{eq:p42-2}, we have
\begin{align*}
    &\sup_{h \in \mathcal{H}} \left|f(h(\mathbf{X}, \mathbf{Y})) - \mathbb{E}\left[f(h(\mathbf{X}', \mathbf{Y}'))\right]\right| \\
    =& \sup_{h \in \mathcal{H}}\max\left\{f(h(\mathbf{X}, \mathbf{Y})) - \mathbb{E}\left[f(h(\mathbf{X}', \mathbf{Y}'))\right], \mathbb{E}\left[f(h(\mathbf{X}', \mathbf{Y}'))\right] - f(h(\mathbf{X}, \mathbf{Y}))\right\}\\
    =& \max\left\{\sup_{h \in \mathcal{H}}f(h(\mathbf{X}, \mathbf{Y})) - \mathbb{E}\left[f(h(\mathbf{X}', \mathbf{Y}'))\right], \sup_{h \in \mathcal{H}}\mathbb{E}\left[f(h(\mathbf{X}', \mathbf{Y}'))\right] - f(h(\mathbf{X}, \mathbf{Y}))\right\}.
\end{align*}

Notice that each of the following events holds with a probability smaller than $\delta /2$:
\begin{align*}
    & \sup_{h \in \mathcal{H}}f(h(\mathbf{X}, \mathbf{Y})) - \mathbb{E}\left[f(h(\mathbf{X}', \mathbf{Y}'))\right] \\
    >& \frac{\sqrt{2 \pi}}{m+n}\left(2 M_{\text {Lip}}(f)+J_{\text{Lip}}(f)\right) \mathbb{E}\left[\mathcal{G}(\mathcal{H}(\mathbf{X}, \mathbf{Y}))\right]+M(f) \sqrt{\frac{\ln (2 / \delta)}{m+n}};
\end{align*}
and 
\begin{align*}
   & \sup_{h \in \mathcal{H}} \mathbb{E}\left[f(h(\mathbf{X}', \mathbf{Y}'))\right] - f(h(\mathbf{X}, \mathbf{Y})) \\
   >& \frac{\sqrt{2 \pi}}{m+n}\left(2 M_{\text {Lip}}(f)+J_{\text{Lip}}(f)\right) \mathbb{E}\left[\mathcal{G}(\mathcal{H}(\mathbf{X}, \mathbf{Y}))\right]+M(f) \sqrt{\frac{\ln (2 / \delta)}{m+n}}.
\end{align*}
In addition, $\forall C \in \mathbb{R}$, we have
\begin{small}
\begin{align*}
    &\left\{\max\left\{\sup_{h \in \mathcal{H}}f(h(\mathbf{X}, \mathbf{Y})) - \mathbb{E}\left[f(h(\mathbf{X}', \mathbf{Y}'))\right], \sup_{h \in \mathcal{H}}\mathbb{E}\left[f(h(\mathbf{X}', \mathbf{Y}'))\right] - f(h(\mathbf{X}, \mathbf{Y}))\right\} > C\right\}\\
    \Leftrightarrow & \left\{\sup_{h \in \mathcal{H}} \mathbb{E}\left[f(h(\mathbf{X}', \mathbf{Y}'))\right] - f(h(\mathbf{X}, \mathbf{Y})) > C\right\} \bigcup \left\{\sup_{h \in \mathcal{H}}f(h(\mathbf{X}, \mathbf{Y})) - \mathbb{E}\left[f(h(\mathbf{X}', \mathbf{Y}'))\right] > C\right\}.
\end{align*}
\end{small}
Notice $\operatorname{Pr}(A\cup B) \leq \operatorname{Pr}(A) + \operatorname{Pr}(B)$, given events $A$ and $B$. 
Consequently, with probability smaller than $\delta$, we have
\begin{align*}
   \max\bigg\{\sup_{h \in \mathcal{H}} & f(h(\mathbf{X}, \mathbf{Y})) - \mathbb{E}\left[f(h(\mathbf{X}', \mathbf{Y}'))\right], \sup_{h \in \mathcal{H}}\mathbb{E}\left[f(h(\mathbf{X}', \mathbf{Y}'))\right] - f(h(\mathbf{X}, \mathbf{Y}))\bigg\}  \\
   & > \frac{\sqrt{2 \pi}}{m+n}\left(2 M_{\text {Lip}}(f)+J_{\text{Lip}}(f)\right) \mathbb{E}\left[\mathcal{G}(\mathcal{H}(\mathbf{X}, \mathbf{Y}))\right]+M(f) \sqrt{\frac{\ln (2 / \delta)}{m+n}}.
\end{align*}
That is, with probability at least $1-\delta$, we have
\begin{align*}
   \max\bigg\{\sup_{h \in \mathcal{H}} & f(h(\mathbf{X}, \mathbf{Y})) - \mathbb{E}\left[f(h(\mathbf{X}', \mathbf{Y}'))\right], \sup_{h \in \mathcal{H}}\mathbb{E}\left[f(h(\mathbf{X}', \mathbf{Y}'))\right] - f(h(\mathbf{X}, \mathbf{Y}))\bigg\} \\
   & \leq \frac{\sqrt{2 \pi}}{m+n}\left(2 M_{\text {Lip}}(f)+J_{\text{Lip}}(f)\right) \mathbb{E}\left[\mathcal{G}(\mathcal{H}(\mathbf{X}, \mathbf{Y}))\right]+M(f) \sqrt{\frac{\ln (2 / \delta)}{m+n}}.
\end{align*}

Next, we consider extending the finite cardinality constraint mentioned in Lemma \ref{lem:maurer} to the finite covering number constraint.
To begin with, notice that the upper bounds provided in Lemma \ref{lem:maurer} are not related to the specific cardinality of the function class $\mathcal{H}$.
Suppose $\mathcal{H}$ is a function set that is equipped with a finite covering number, then $\forall \epsilon > 0$, $\exists \mathcal{H}_\epsilon \subseteq \mathcal{H}$, s.t. $\forall h \in \mathcal{H}$, $\exists h_\epsilon \in \mathcal{H}_\epsilon$, s.t. $\sup_{\mathbf{u} \in \mathcal{U}^{m+n}}\|h(\mathbf{u})-h_\epsilon(\mathbf{u})\|\leq \epsilon$. 
Moreover, since $M_{\text{Lip}}(f) < \infty$, $\forall \mathbf{u}, \mathbf{u}' \in \mathcal{U}^n$, we have 
\begin{eqnarray*}
    && f(\mathbf{u}) - f(\mathbf{u}')\\
    &=&f(u_1, \dots, u_{m+n}) - f(u_1', \dots, u_{m+n}')\\
    &=&f(u_1, \dots) - f(u_1', u_2, \dots) + f(u_1', u_2, \dots) - \cdots \\
    & & \quad + f(\dots,u_{m+n-1}', u_{m+n}) - f(\dots,u_{m+n-1}', u_{m+n}')\\
    &\stackrel{\text{Assumption \ref{assum:maurer}}}{\leq}& \frac{1}{m+n}M_{\text{Lip}}(f)\sum_{i=1}^{m+n}\|u_i - u_i'\|\\
    &\stackrel{\text{QM-AM Inequality}}{\leq}& M_{\text{Lip}}(f)\sqrt{\frac{\sum_{i=1}^{m+n}\|u_i - u_i'\|^2}{m+n}}\\
    &=& \frac{M_{\text{Lip}}(f)}{\sqrt{m+n}} \|\mathbf{u}-\mathbf{u}'\|.
\end{eqnarray*}

Consequently, $\forall \epsilon > 0, h \in \mathcal{H}$, we have
\begin{equation}
\label{eq:epsilon}
\begin{aligned}
&\left|\mathbb{E}_{\mathbf{X}^{\prime}}\left[f(h(\mathbf{X}', \mathbf{Y}'))\right]-f(h(\mathbf{X}, \mathbf{Y}))\right|\\ 
\stackrel{\phantom{\text{Definition of }\mathcal{G}_\epsilon}}{=}& \bigg|\mathbb{E}_{\mathbf{X}^{\prime}}\left[f(h)\right] - \mathbb{E}_{\mathbf{X}^{\prime}}\left[f\left(h_\epsilon\right)\right] + \mathbb{E}_{\mathbf{X}^{\prime}}\left[f\left(h_\epsilon\right)\right] - f(h_\epsilon) + f(h_\epsilon) - f(h)\bigg|\\
\stackrel{\phantom{\text{Definition of }\mathcal{G}_\epsilon}}{\leq}&\bigg|\mathbb{E}_{\mathbf{X}^{\prime}}\left[f(h)\right] - \mathbb{E}_{\mathbf{X}^{\prime}}\left[f\left(h_\epsilon\right)\right]\bigg| + \bigg|\mathbb{E}_{\mathbf{X}^{\prime}}\left[f\left(h_\epsilon\right)\right] - f(h_\epsilon)\bigg| + \bigg|f(h_\epsilon) - f(h)\bigg|\\
\stackrel{\substack{\text{Definition of }\mathcal{G}_\epsilon\\\text{Lipschitz of }f}}{\leq}& \frac{2\epsilon M_{\text{Lip}}(f)}{\sqrt{m+n}} + \bigg|\mathbb{E}_{\mathbf{X}^{\prime}}\left[f\left(h_\epsilon\right)\right] - f(h_\epsilon)\bigg|\\
\stackrel{\phantom{\text{Definition of }\mathcal{G}_\epsilon}}{\leq}& \frac{2\epsilon M_{\text{Lip}}(f)}{\sqrt{m+n}} + \sup_{h \in \mathcal{H}_\epsilon}\bigg|\mathbb{E}_{\mathbf{X}^{\prime}}\left[f(h(\mathbf{X}', \mathbf{Y}'))\right] - f(h(\mathbf{X}, \mathbf{Y}))\bigg|.
\end{aligned}
\end{equation}

Notice that the bound for the function class $\mathcal{H}_\epsilon$ is less than or equal to the one for $\mathcal{H}$. More specifically, we have:
\begin{eqnarray*}
& &  \mathbb{E}\left[\sup_{h \in \mathcal{H}_\epsilon}\left|\mathbb{E}_{\mathbf{X}^{\prime}}\left[f(h(\mathbf{X}', \mathbf{Y}'))\right] - f(h(\mathbf{X}, \mathbf{Y}))\right|\right] \\
&\stackrel{\eqref{eq:p42-1}}{\leq}& \frac{\sqrt{2 \pi}}{m+n}\left(2 M_{\text {Lip}}(f)+J_{\text {Lip}}(f)\right) \mathbb{E}[\mathcal{G}(\mathcal{H}_\epsilon(\mathbf{X}, \mathbf{Y}))]\\
&\stackrel{\mathcal{H}_\epsilon \subseteq \mathcal{H}}{\leq}& \frac{\sqrt{2 \pi}}{m+n}\left(2 M_{\text {Lip}}(f)+J_{\text {Lip}}(f)\right) \mathbb{E}[\mathcal{G}(\mathcal{H}(\mathbf{X}, \mathbf{Y}))].
\end{eqnarray*}
Moreover, $\forall \delta \in (0, 1)$, with probability at least $1-\delta$, we have
\begin{eqnarray*}
    &&\sup_{h \in \mathcal{H}_\epsilon}\left|\mathbb{E}_{\mathbf{X}^{\prime}}\left[f(h(\mathbf{X}', \mathbf{Y}'))\right] - f(h(\mathbf{X}, \mathbf{Y}))\right|\\ &\stackrel{\eqref{eq:p42-2}}{\leq}&  \frac{\sqrt{2 \pi}}{m+n}\left(2 M_{\text {Lip}}(f)+J_{\text {Lip}}(f)\right) \mathbb{E}[\mathcal{G}(\mathcal{H}_\epsilon(\mathbf{X}, \mathbf{Y}))] + M(f)\sqrt{\frac{\ln(2/\delta)}{m+n}}\\
    &\stackrel{\mathcal{H}_\epsilon \subseteq \mathcal{H}}{\leq}& \frac{\sqrt{2 \pi}}{m+n}\left(2 M_{\text {Lip}}(f)+J_{\text {Lip}}(f)\right) \mathbb{E}[\mathcal{G}(\mathcal{H}(\mathbf{X}, \mathbf{Y}))] + M(f)\sqrt{\frac{\ln(2/\delta)}{m+n}}.
\end{eqnarray*}
Combining the above two inequalities and \eqref{eq:epsilon}, letting $\epsilon \rightarrow 0$, the extension is proved.
\end{proof}

\subsection{Proof of Proposition \ref{prop:Gaussian}}
\label{sec:proof-2}
\begin{proof}
    The first inequality is referred to \cite[Theorem 2.26]{wainwright2019high}.
    To employ the McDiarmid's inequality, let $f(X_1, \dots, X_n):= \mathcal{G}(\mathcal{H}(\mathbf{X}))$. We have
    \begin{align*}
        & \left|f(x, \dots, x_n)-f(x_1, \dots, x_{i-1}, \widetilde{x}, x_{i+1}, \dots, x_n)\right|\\
        =& \left|\mathbb{E}_\xi\left[\sup_{h}\sum_{j=1}^n \langle \xi_j, h(x_j) \rangle\right]-\mathbb{E}_\xi\left[\sup_{h}\sum_{j \neq i} \langle \xi_j, h(x_j) \rangle + \langle \xi_i, h(\widetilde{x}) \rangle\right]\right|\\
        \leq& \left|\mathbb{E}_\xi\left[\sup_h \langle \xi_i, h(x_i)-h(\widetilde{x})\rangle \right]\right|.
    \end{align*}
    Derived from Cauchy Schwarz's inequality, we have
    \begin{align*}
        & \left|\mathbb{E}_\xi\left[\sup_h \langle \xi_i, h(x_i)-h(\widetilde{x})\rangle \right]\right|\\
        \leq & \left|\mathbb{E}_\xi\left[\sup_h \| \xi_i\| \|h(x_i)-h(\widetilde{x})\| \right]\right|\\
        \leq & D(\mathcal{H}(\mathcal{X}))\sqrt{d}.
    \end{align*}
    Combining the above quantity with McDiarmid's inequality (Lemma \ref{lem:mcdiarmid}), the second inequality is derived.
\end{proof}

\subsection{Proof of Lemma \ref{lem:contraction}}
\label{sec:proof-4.6}
\begin{proof}
    Recall the definition of Gaussian complexity (Definition \ref{def:Gauss}).
    We have
    \begin{align*}
        \mathcal{G}(\phi \circ \mathcal{F}(\mathbf{X})) := \mathbb{E}_\xi \sup_{f \in \mathcal{F}} \sum_{i=1}^n\sum_{j=1}^{d_1}\xi_{i,j}\phi(f(X_i)_j).
    \end{align*}
    Based on the symmetric property of the Gaussian random variables, we have
    \begin{align*}
        \mathcal{G}(\phi \circ \mathcal{F}(\mathbf{X})) = \frac{1}{2}\Bigg\{\mathbb{E}_\xi& \left[\sup_{f \in \mathcal{F}} \sum_{i=1}^{n-1}\sum_{j=1}^{d_1}\xi_{i,j}\phi(f(X_i)_j) + \sum_{j=1}^{d_1} \xi_{n, j}\phi(f(X_n)_j) \right] \\
        &+ \mathbb{E}_\xi \left[\sup_{f \in \mathcal{F}} \sum_{i=1}^{n-1}\sum_{j=1}^{d_1}\xi_{i,j}\phi(f(X_i)_j) - \sum_{j=1}^{d_1} \xi_{n, j}\phi(f(X_n)_j) \right]\Bigg\}.
    \end{align*}
    Define the functions $f_1 \in \mathcal{F}$ and $f_2 \in \mathcal{F}$ as follows:
    \begin{align*}
        f_1 := \arg\max_f \sum_{i=1}^{n-1}\sum_{j=1}^{d_1}\xi_{i,j}\phi(f(X_i)_j) + \sum_{j=1}^{d_1} \xi_{n, j}\phi(f(X_n)_j),\\
        f_2 := \arg\max_f \sum_{i=1}^{n-1}\sum_{j=1}^{d_1}\xi_{i,j}\phi(f(X_i)_j) - \sum_{j=1}^{d_1} \xi_{n, j}\phi(f(X_n)_j).
    \end{align*}
    That is, $\mathcal{G}(\phi \circ \mathcal{F}(\mathbf{X}))$ is equal to the following expression:
    \begin{align*}
        \frac{1}{2}\mathbb{E}_\xi \Bigg[ \sum_{i=1}^{n-1}\sum_{j=1}^{d_1}\xi_{i,j}\left(\phi(f_1(X_i)_j) + \phi(f_2(X_i)_j)\right) + \sum_{j=1}^{d_1} \xi_{n, j}\left(\phi(f_1(X_n)_j) - \phi(f_2(X_n)_j)\right)\Bigg].
    \end{align*}
    Based on the fact that the function $\phi$ is $l_\phi$-Lipschitz, we have
    \begin{align*}
         &\frac{1}{2}\mathbb{E}_\xi \Bigg[ \sum_{i=1}^{n-1}\sum_{j=1}^{d_1}\xi_{i,j}\left(\phi(f_1(X_i)_j) + \phi(f_2(X_i)_j)\right) + \sum_{j=1}^{d_1} \xi_{n, j}\left(\phi(f_1(X_n)_j) - \phi(f_2(X_n)_j)\right)\Bigg]\\ 
         =& \frac{1}{2}\mathbb{E}_\xi \Bigg[ \sum_{i=1}^{n-1}\sum_{j=1}^{d_1}\xi_{i,j}\left(\phi(f_1(X_i)_j) + \phi(f_2(X_i)_j)\right) + \sum_{j=1}^{d_1} l_\phi \xi_{n, j}s_j \left(f_1(X_n)_j - f_2(X_n)_j\right)\Bigg]\\
         =& \frac{1}{2}\mathbb{E}_\xi \Bigg[\sum_{i=1}^{n-1}\sum_{j=1}^{d_1}\xi_{i,j}\phi(f_1(X_i)_j) + \sum_{j=1}^{d_1} l_\phi s_j \xi_{n, j}f_1(X_n)_j \Bigg]\\ 
         & \quad \quad \quad \quad \quad \quad \quad \quad \quad + \frac{1}{2}\mathbb{E}_\xi \Bigg[\sum_{i=1}^{n-1}\sum_{j=1}^{d_1}\xi_{i,j}\phi(f_2(X_i)_j) - \sum_{j=1}^{d_1} l_\phi s_j \xi_{n, j}f_2(X_n)_j \Bigg].
    \end{align*}
    Here, $s_j$ refers to the sign of $f_1(X_n)_j - f_2(X_n)_j$.
    Note that: 
    \begin{align*}
    &\mathbb{E}_\xi \Bigg[\sum_{i=1}^{n-1}\sum_{j=1}^{d_1}\xi_{i,j}\phi(f_1(X_i)_j) + \sum_{j=1}^{d_1} l_\phi s_j \xi_{n, j}f_1(X_n)_j \Bigg]\\
    =& \mathbb{E}_\xi \Bigg[\sum_{i=1}^{n-1}\sum_{j=1}^{d_1}\xi_{i,j}\phi(f_1(X_i)_j) + l_\phi \sum_{j=1}^{d_1} \xi_{n, j}f_1(X_n)_j \Bigg]\\
    \leq&\mathbb{E}_\xi \sup_{f \in \mathcal{F}}\Bigg[\sum_{i=1}^{n-1}\sum_{j=1}^{d_1}\xi_{i,j}\phi(f(X_i)_j) + l_\phi \sum_{j=1}^{d_1} \xi_{n, j}f(X_n)_j \Bigg].
    \end{align*}
    Similarly, we have 
    \begin{align*}
    &\mathbb{E}_\xi \Bigg[\sum_{i=1}^{n-1}\sum_{j=1}^{d_1}\xi_{i,j}\phi(f_2(X_i)_j) - \sum_{j=1}^{d_1} l_\phi s_j \xi_{n, j}f_2(X_n)_j \Bigg]\\
    \leq&\mathbb{E}_\xi \sup_{f \in \mathcal{F}}\Bigg[\sum_{i=1}^{n-1}\sum_{j=1}^{d_1}\xi_{i,j}\phi(f(X_i)_j) - l_\phi \sum_{j=1}^{d_1} \xi_{n, j}f(X_n)_j \Bigg]\\
    =& \mathbb{E}_\xi \sup_{f \in \mathcal{F}}\Bigg[\sum_{i=1}^{n-1}\sum_{j=1}^{d_1}\xi_{i,j}\phi(f(X_i)_j) + l_\phi \sum_{j=1}^{d_1} \xi_{n, j}f(X_n)_j \Bigg].
    \end{align*}
    Consequently,
    \begin{align*}
        \mathcal{G}(\phi \circ \mathcal{F}(\mathbf{X})) \leq \mathbb{E}_\xi \sup_{f \in \mathcal{F}}\Bigg[\sum_{i=1}^{n-1}\sum_{j=1}^{d_1}\xi_{i,j}\phi(f(X_i)_j) + l_\phi \sum_{j=1}^{d_1} \xi_{n, j}f(X_n)_j \Bigg].
    \end{align*}
    Through mathematical induction, we have
    \begin{align*}
        \mathcal{G}(\phi \circ \mathcal{F}(\mathbf{X})) \leq l_\phi \mathbb{E}_\xi \sup_{f \in \mathcal{F}}\Bigg[\sum_{i=1}^{n}\sum_{j=1}^{d_1}\xi_{i,j}f(X_i)_j \Bigg] = l_\phi \mathcal{G}(\mathcal{F}(\mathbf{X})).
    \end{align*}
\end{proof}

\subsection{Proof of Proposition \ref{prop:linear}}
\label{sec:proof-4.7}
\begin{proof}
    Consider the following linear function set mapping from $\mathbb{R}^{d_0}$ to $\mathbb{R}^{d_1}$ under regularization constraints:
    \begin{align*}
        \mathcal{F}:= \{x \mapsto \boldsymbol{w} \cdot x \mid \boldsymbol{w} \in \mathbb{R}^{d}, \|\boldsymbol{w}\|_1 \leq \omega\}.
    \end{align*}
    From the definition of Gaussian complexity (Definition \ref{def:Gauss}), we have:
    \begin{eqnarray*}
        \mathcal{G}(\mathcal{F}(\mathbf{X})) &=& \mathbb{E}_\xi \sup_{\|\boldsymbol{w}\|_1\leq \omega} \sum_{i=1}^n \xi_{i} \boldsymbol{w}^T X_i\\
        &=& \mathbb{E}_\xi \sup_{\|\boldsymbol{w}\|_{1} \leq \omega} \boldsymbol{w}^T \sum_{i=1}^n X_i \xi_{i}\\
        &\stackrel{\text{Hölder's Inequality}}{\leq}& \mathbb{E}_\xi \sup_{\|\boldsymbol{w}\|_{1} \leq \omega} \|\boldsymbol{w}\|_1 \left\|\sum_{i=1}^n X_i \xi_{i}\right\|_\infty\\
        &=& \omega \mathbb{E}_\xi\left\|\sum_{i=1}^n X_i \xi_{i}\right\|_\infty\\
        &=& \omega \mathbb{E}_\xi\max_k \left|\sum_{i=1}^n \xi_{i} X_{i,k} \right|\\.
    \end{eqnarray*}
    Let $Y_{k}:= \sum_{i=1}^n \xi_{i} X_{i,k}$, $k = 1, \dots, d_0$.
    We have $Y_{k} \sim \mathcal{N}\left(0, \sum_{i=1}^n X_{i,k}^2\right)$. 
    Consider the normalized random variable $Z_{k} = Y_{k} / \sqrt{\sum_{i=1}^n X_{i,k}^2}$, we have $Z_{k} \stackrel{\text{i.i.d.}}{\sim} \mathcal{N}(0, 1)$.
    That is, 
    \begin{eqnarray*}
        \max_{k}|Y_{k}| &=& \max_{k} \sqrt{\sum_{i=1}^n X_{i,k}^2}|Z_{k}|\\
        &\leq& \max_k\sqrt{\sum_{i=1}^n X_{i,k}^2} \max_{k}|Z_{k}|.
    \end{eqnarray*}
    Consider the random vector $\max_{j,k} |Z_{k}|$.
    For any $\lambda > 0$, we have
    \begin{eqnarray*}
        \mathbb{E}\left[\max_{k} |Z_{k}| \right] &=&  \mathbb{E}\left[\frac{1}{\lambda} \log \left(\max_{k} e^{\lambda\left|Z_{k}\right|}\right)\right]\\
        &\leq& \mathbb{E}\left[\frac{1}{\lambda} \log \left(\sum_{k} e^{\lambda\left|Z_{k}\right|}\right)\right]\\
        &\stackrel{\text{Jensen's Inequality}}{\leq}& \frac{1}{\lambda}\log \mathbb{E}\left[ \left(\sum_{k} e^{\lambda\left|Z_{k}\right|}\right)\right]\\
        &=& \frac{1}{\lambda} \log \left(d \mathbb{E}\left[ \left( e^{\lambda\left|Z_0\right|}\right)\right]\right).
    \end{eqnarray*}
    Through integral, we have
    \begin{eqnarray*}
        \mathbb{E}\left[\exp(\lambda |Z_0|)\right] \leq 2 e^{\lambda^2 / 2}.
    \end{eqnarray*}
    Combining the above results, we have
    \begin{eqnarray*}
        \mathbb{E}\left[\max_k |Z_k| \right] &\leq& \frac{\log(2d)}{\lambda} + \frac{\lambda}{2}.
    \end{eqnarray*}
    Let $\lambda = \sqrt{2\log(2d)}$, we have
    \begin{eqnarray*}
        \mathbb{E}\left[\max_k |Z_k| \right] &\leq& \sqrt{2{\log(2d)}}.
    \end{eqnarray*}
    Consequently, we have
    \begin{align*}
        \mathcal{G}(\mathcal{F}(\mathbf{X})) \leq \omega \sqrt{2\log(2d)} \max_{k}\sqrt{\sum_{i=1}^n X_{i, k}^2}.
    \end{align*}
\end{proof}

\subsection{Proof of Proposition \ref{prop:fnn}}
\label{sec:proof-prop4.8}
\begin{proof}
    The Gaussian complexity of a feed-forward neural network can be bounded recursively by considering each layer at a time.
    We demonstrate it through the following proposition. The proof is referred to Section \ref{sec:proof-recursive}
    \begin{proposition}
    \label{prop:recursive}
        Let $\mathcal{L}$ be a class of functions from $\mathbb{R}^d$ to $\mathbb{R}$ that includes the zero function. Let $\sigma: \mathbb{R} \rightarrow \mathbb{R}$ be $\gamma$-Lipschitz.
        Define the function set $\mathcal{L}^\prime$ as follows:
        $$
            \mathcal{L}^{\prime}:=\left\{x \in \mathbb{R}^d \mapsto \sigma\left(\sum_{j=1}^m w_j l_j(x)\right) \in \mathbb{R}:\|w\|_1 \leq \omega, l_1, \ldots, l_m \in \mathcal{L}\right\}.
        $$ 
        Then, for any data matrix $\mathbf{X}:=(X_1, \dots, X_n)^T \in \mathbb{R}^{n \times d}$, we have
        \begin{eqnarray*}
            \mathcal{G}(\mathcal{L}'(\mathbf{X})) \leq 2\gamma\omega \mathcal{G}(\mathcal{L}(\mathbf{X})).
        \end{eqnarray*}
    \end{proposition}
    Recall that the activation function of the last layer of the defined neural network is an identity function, i.e., $\sigma^{(\iota)}(x) = x$, we can apply Proposition \ref{prop:recursive} with $\gamma = 1$ once and apply it with $\gamma = \lambda$ for $\iota - 2$ times.
    Consequently, we have
    $$
        \mathcal{G}(\mathcal{F}(\mathbf{X})) \leq (2\omega)^{\iota - 1}\lambda^{\iota - 2}\sigma \circ \mathcal{G}(\mathcal{F}_{\text{linear}}(\mathbf{X})),
    $$
    where $\mathcal{F}_{\text{linear}}$ is the function set defined in Proposition \ref{prop:linear}.
    From the contraction inequality (Lemma \ref{lem:contraction}) and Proposition \ref{prop:linear}, we have
    $$
        \mathcal{G}(\mathcal{F}(\mathbf{X})) \leq (2\omega)^\iota \lambda^{\iota - 1} \sqrt{2 \log (2 d)} \max_k
        \sqrt{\sum_{i=1}^n X_{i, k}^2}.
    $$
\end{proof}

\subsubsection{Proof of Proposition \ref{prop:recursive}}
\label{sec:proof-recursive}
\begin{proof}
    To begin with, we define a function set $\mathcal{F}$ as follows: 
    $$
        \mathcal{F}:=\left\{x \in \mathbb{R}^d \rightarrow \sum_{i=1}^m w_j l_j(x) \in \mathbb{R}:\|w\|_1 \leq \omega, l_1, \ldots, l_m \in \mathcal{L}\right\}.
    $$ 
    Utilize the contraction inequality (Lemma \ref{lem:contraction}), we have
    $\mathcal{G}(\mathcal{L}'(\mathbf{X}))\leq \gamma \mathcal{G}(\mathcal{F}(\mathbf{X}))$.
    Note that 
    $$
        \mathcal{G}(\mathcal{F}(\mathbf{X})) = \mathcal{G}(\mathcal{F}'(\mathbf{X})),
    $$
    where $\mathcal{F}':= \left\{x \in \mathbb{R}^d \rightarrow \sum^m w_j l_j(x) \in \mathbb{R}:\|w\|_1=\omega, l_1, \ldots, l_m \in \mathcal{L}\right\}$.
    In addition, for any $\omega \in \mathbb{R}^m$ such that $\|\omega\|_1 = 1$, we have
    $$
        \sum_{i} w_i l_i = \sum_{i: w_i \geq 0} w_i (l_i - 0) + \sum_{i, w_i < 0}|w_i|(0-l_i),
    $$
    which is a convex combination of elements in $\mathcal{L}-\mathcal{L}$.
    Recall the definition of Gaussian complexity, i.e., given a set $\mathcal{T}$, we have $\mathcal{G}(\mathcal{T}):=\sup_{\boldsymbol{t} \in \mathcal{T}} \sum_{i} t_i \xi_i$. 
    Note that the Gaussian complexity for a set $\mathcal{T}$ is equal to the one for the convex hull of $\mathcal{T}$, i.e.,
    $\mathcal{G}(\mathcal{T}) =\mathcal{G}(\operatorname{conv}(\mathcal{T}))$, where $\operatorname{conv}(\mathcal{T}) = \{\sum_{i} \lambda_i \boldsymbol{t}^{(i)} \mid \boldsymbol{t}^{(i)} \in \mathcal{T}, \sum_i \lambda_i = 1, \lambda_i \geq 0, \forall i\}, $
    Hence, by applying in order the convex hall property, the summation property, and the scalar multiplication property of Gaussian complexity, we have
    \begin{eqnarray*}
        \mathcal{G}(\mathcal{F}(\mathbf{X})) &\leq& \omega \mathcal{G}(\operatorname{conv}(\mathcal{L}-\mathcal{L})(\mathbf{X}))\\
        &=& \omega \mathcal{G}((\mathcal{L}-\mathcal{L})(\mathbf{X}))\\
        &=& \omega \mathcal{G}(\mathcal{L}(\mathbf{X})) + \omega \mathcal{G}(-\mathcal{L}(\mathbf{X}))\\
        &=& 2\omega \mathcal{G}(\mathcal{L}(\mathbf{X})).
    \end{eqnarray*}
    In summary, we have
    \begin{align*}
        \mathcal{G}(\mathcal{L}'(\mathbf{X})) \leq 2\gamma\omega \mathcal{G}(\mathcal{L}(\mathbf{X}))
    \end{align*}
\end{proof}

\subsection{Proof of Corollary \ref{coro:MMD}}
\label{sec:proof-3}
\begin{proof}
We first consider the differences between empirical estimators $\widehat{\gamma}_{k, u}$, $\widehat{\gamma}_{k, b}$, and $\widehat{\gamma}_{k, e}$. 
Based on Definition \ref{def:empiricalMMDestimators}, i.e., 
\begin{equation*}
    \begin{aligned}
        &\widehat{\gamma}_{k, u}^2(\mathbf{X}, \mathbf{Y}) = \frac{1}{m(m-1)}\sum_{i\neq j}^m k(X_i, X_j) + \frac{1}{n(n-1)}\sum_{i\neq j}^n  k(Y_i, Y_j) - \frac{2}{mn}\sum_{i=1}^m\sum_{j=1}^n k(X_i, Y_j);\\
        &\widehat{\gamma}_{k,b}^2(\mathbf{X}, \mathbf{Y}) = \frac{1}{m^2} \sum_{i, j=1}^m k\left(X_i, X_j\right) +\frac{1}{n^2} \sum_{i, j=1}^n k\left(Y_i, Y_j\right) -\frac{2}{m n} \sum_{i, j=1}^{m, n} k\left(X_i, X_j\right).
    \end{aligned}
    \end{equation*}
We have
\begin{align*}
    \left|\widehat{\gamma}_{k,u}^2(\mathbf{x}, \mathbf{y}) - \widehat{\gamma}_{k,b}^2(\mathbf{x}, \mathbf{y})\right| = \Bigg|\frac{1}{m^2(m-1)} \sum_{i=1}^m \bigg[\sum_{j \neq i}k\left(x_i, x_j\right)-(m-1)k(x_i, x_i)\bigg] + \\
    \frac{1}{n^2(n-1)} \sum_{i=1}^n \bigg[\sum_{j \neq i}k\left(y_i, y_j\right)-(n-1)k(y_i, y_i)\bigg]\Bigg|.
\end{align*}

Suppose $\sup_{u \in \mathcal{U}}k(u,u) \leq \nu$, then $\forall u, u' \in \mathcal{U}$, we have
\begin{align}
\label{eq:k(u, u')}
    k(u, u') = \langle k(u, \cdot), k(u', \cdot) \rangle_\mathcal{F} \stackrel{\text{Cauchy Schwarz's}}{\leq} \|k(u, \cdot)\|_\mathcal{F}\|k(u', \cdot)\|_\mathcal{F} = \sqrt{k(u, u) k(u', u')} \leq \nu.
\end{align}
Consequently,
\begin{align*}
    \left|\frac{1}{m^2(m-1)} \sum_{i=1}^m \left[\sum_{j \neq i}k\left(x_i, x_j\right)-(m-1)k(x_i, x_i)\right]\right| \leq \frac{m(m-1)}{m^2(m-1)}2\nu = \frac{2\nu}{m}.
\end{align*}
Similar results holds for $\frac{1}{n^2(n-1)} \sum_{i=1}^n \left[\sum_{j \neq i}k\left(y_i, y_j\right)-(n-1)k(y_i, y_i)\right]$.
As a consequence, let $\rho_x = m/(m+n)$ and $n = n/(m+n)$, we have
\begin{align*}
    \left|\widehat{\gamma}_{k,u}^2(\mathbf{X}, \mathbf{Y}) - \widehat{\gamma}_{k,b}^2(\mathbf{X}, \mathbf{Y})\right| \leq 2\nu \left(\frac{1}{m}+\frac{1}{n}\right) = 2\nu\frac{\rho_x^{-1}\rho_y^{-1}}{m+n}.
\end{align*}
Suppose the sample size $m = n$, then 
recall Definition \ref{def:empiricalMMDestimators}, we have
    \begin{equation*}
    \begin{aligned}
        &\widehat{\gamma}_{k, u}^2(\mathbf{X}, \mathbf{Y}) = \frac{1}{n(n-1)}\sum_{i\neq j}^n k(X_i, X_j) + \frac{1}{n(n-1)}\sum_{i\neq j}^n  k(Y_i, Y_j) - \frac{2}{n^2}\sum_{i=1}^n\sum_{j=1}^n k(X_i, Y_j);\\
        & \widehat{\gamma}_{k,e}^2(\mathbf{X}, \mathbf{Y})= \frac{1}{n(n-1)} \sum_{i=1}^n \sum_{j \neq i}^n k\left(X_i, X_j\right)+k\left(Y_i, Y_j\right)-k\left(X_i, Y_j\right)-k\left(X_j, Y_i\right).        
    \end{aligned}
    \end{equation*}
Consequently, we have
\begin{align*}
    \left|\widehat{\gamma}_{k,u}^2(\mathbf{x}, \mathbf{y}) - \widehat{\gamma}_{k,b}^2(\mathbf{x}, \mathbf{y})\right| =&  
    \left|\frac{2}{n^2}\sum_{i=1}^n \sum_{j=1}^n k(x_i, y_j) - \frac{2}{n(n-1)}\sum_{i=1}^n\sum_{j \neq i}k(x_i, y_j)\right|\\
    =&\frac{2}{n}\left|\sum_{i=1}^n\left[\frac{1}{n}\sum_{j=1}^nk(x_i, y_j) - \frac{1}{n-1}\sum_{j \neq i} k(x_i, y_j)\right]\right|\\
    =&\frac{2}{n^2}\left|\sum_{i=1}^n\left[k(x_i, y_i) - \frac{1}{(n-1)}\sum_{j\neq i}^n k(x_i, y_j)\right]\right|\\
    \leq& \frac{4\nu}{n}.
\end{align*}
The last inequality is derived from the assumption that $\sup_{u \in \mathcal{U}}k(u, u) \leq \nu$ and \eqref{eq:k(u, u')}.

Next, we consider the seminorms $M_{\text{Lip}}$, $J_{\text{Lip}}$, and $M$ of the unbiased estimator $\widehat{\gamma}_{k, u}$ under Assumption \ref{assum:BandL}.
To obtain upper bounds of the seminorms of $\widehat{\gamma}_{k,u}^2$, we first calculate the $k$-th partial difference, that is, for $k=1,\dots,m$, we have
\begin{align*}
    D_{xx'}^k\widehat{\gamma}_{k,u}^2(\mathbf{x}, \mathbf{y}) = \frac{2}{m(m-1)}\sum_{i \neq k}(k(x, x_i) - k(x', x_i)) - \frac{2}{mn}\sum_{j=1}^n (k(x, y_j) - k(x',y_j)).
\end{align*}
Similarly, for $k=m+1,\dots,m+n$, we have
\begin{align*}
    D_{yy'}^k\widehat{\gamma}_{k,u}^2(\mathbf{x}, \mathbf{y}) = \frac{2}{n(n-1)}\sum_{i \neq k}(k(y, y_i) - k(y', y_i)) - \frac{2}{mn}\sum_{j=1}^m (k(y, x_j) - k(y',x_j)).
\end{align*}
Suppose $\sup_{u \in \mathcal{U}}k(u,u) \leq \nu$, then since $\langle k(u_1, \cdot), k(u_2, \cdot)\rangle_\mathcal{F} = k(u_1, u_2)$, $\forall u_1, u_2$, the following inequality can be induced, that is, $\forall u, u' \in \mathcal{U}$, we have
\begin{equation}
\label{eq:k(u,u')}
\begin{aligned}
    k(u,u') &\stackrel{\phantom{\text{Cauchy Schwarz's Inequality}}}{=} \langle k(u, \cdot), k(u', \cdot)\rangle_{\mathcal{F}}\\
    &\stackrel{\text{Cauchy Schwarz's Inequality}}{\leq} \|k(u, \cdot)\|_{\mathcal{F}}\|k(u', \cdot)\|_{\mathcal{F}}\\
    &\stackrel{\phantom{\text{Cauchy Schwarz's Inequality}}}{=} \sqrt{k(u,u) k(u',u')}\\
    &\stackrel{\sup_{u \in \mathcal{U}}k(u,u) \leq \nu}{\mathmakebox[\widthof{$\stackrel{\text{Cauchy Schwarz's Inequality}}{\leq}$}]{\leq}} \nu.
\end{aligned}
\end{equation}

Based on the above result, for $k=1,\dots,m$, we have

\begin{align*}
    &D_{xx'}^k\widehat{\gamma}_{k,u}^2(\mathbf{x}, \mathbf{y})\\ 
    \stackrel{\phantom{\sup_{u,u' \in \mathcal{U}}k(u,u') \leq \nu}}{=}& \frac{2}{m(m-1)}\sum_{i \neq k}(k(x, x_i) - k(x', x_i)) - \frac{2}{mn}\sum_{j=1}^n (k(x, y_j) - k(x',y_j))\\ \stackrel{\text{Triangle's inequality}}{\mathmakebox[\widthof{$\stackrel{\sup_{u,u' \in \mathcal{U}}k(u,u') \leq \nu}{\leq}$}]{\leq}}& \frac{2}{m(m-1)}\sum_{i \neq k}\left|k(x, x_i) - k(x', x_i))\right| + \frac{2}{mn}\sum_{j=1}^n \left|k(x, y_j) - k(x',y_j))\right|\\
    \stackrel{\sup_{u,u' \in \mathcal{U}}k(u,u') \leq \nu}{\leq}& \frac{8\nu}{m}.
\end{align*}

Similarly, for $k=m+1,\dots,m+n$, we have

\begin{align*}
    D_{yy'}^k\widehat{\gamma}_{k,u}^2(\mathbf{x}, \mathbf{y}) \leq \frac{8\nu}{n}.
\end{align*}

On the other hand, suppose the reproducing kernel function is Lipschitz, i.e., there exists a constant $l > 0$, such that $M_\text{Lip}(k) \leq l$. 
Then we have

\begin{align*}
    D_{xx'}^k\widehat{\gamma}_{k,u}^2(\mathbf{x}, \mathbf{y}) =& \frac{2}{m(m-1)}\sum_{i \neq k}(k(x, x_i) - k(x', x_i)) - \frac{2}{mn}\sum_{j=1}^n (k(x, y_j) - k(x',y_j))\\
    \leq & \frac{2l}{m}\|x-x'\|.
\end{align*}
Similarly, for $k=m+1, \dots, m+n$, we have
\begin{align*}
    D_{yy'}^k\widehat{\gamma}_{k,u}^2(\mathbf{x}, \mathbf{y}) 
    \leq & \frac{2l}{n}\|y-y'\|.
\end{align*}

\noindent To provide an upper bound for $J_{\text{Lip}}(\widehat{\gamma}_{k,u}^2)$, we first consider $D_{z z^{\prime}}^l D_{y y^{\prime}}^k \widehat{\gamma}_{k,u}^2$.

\noindent For $k = 1,\dots,m$ and $j=1,\dots,m$, we have
\begin{align*}
    & D_{z z^{\prime}}^j D_{x x^{\prime}}^k \widehat{\gamma}_{k,u}^2\\ 
    \stackrel{\phantom{M_{\text{Lip}}(k) \leq l}}{=} & D_{z, z'}^j\left[\frac{2}{m(m-1)}\sum_{i \neq k}(k(x, x_i) - k(x', x_i)) - \frac{2}{mn}\sum_{i=1}^n (k(x, y_i) - k(x',y_i))\right]\\
    \stackrel{\phantom{M_{\text{Lip}}(k) \leq l}}{=} &\frac{2}{m(m-1)}\left[k(x, z) - k(x', z) - k(x, z') + k(x', z')\right]\\
    \stackrel{M_{\text{Lip}}(k) \leq l}{\leq} & \frac{2l\|x-x'\|}{m(m-1)}.
\end{align*}
For $k = 1,\dots,m$ and $j=m+1,\dots,m+n$, we have
\begin{align*}
    &D_{z z^{\prime}}^j D_{x x^{\prime}}^k \widehat{\gamma}_{k,u}^2 \\
    \stackrel{\phantom{M_\text{Lip}(k) \leq l}}{=}& D_{z, z'}^j\left[\frac{2}{m(m-1)}\sum_{i \neq k}(k(x, x_i) - k(x', x_i)) - \frac{2}{mn}\sum_{i=1}^n (k(x, y_i) - k(x',y_i))\right]\\
    \stackrel{\phantom{M_\text{Lip}(k) \leq l}}{=}& \frac{2}{mn}\left[k(x, z) - k(x', z) - k(x, z') + k(x', z')\right]\\
    \stackrel{M_\text{Lip}(k) \leq l}{\leq}& \frac{2l\|x-x'\|}{mn}.
\end{align*}
Combining the above results, for $k = 1,\dots,m$, we have
\begin{align*}
    \max_j D_{z z^{\prime}}^j D_{x x^{\prime}}^k \widehat{\gamma}_{k,u}^2 \leq \frac{2l\|x-x'\|}{m}\max\{\frac{1}{m-1}, \frac{1}{n}\}.
\end{align*}
Similarly, for $k = m+1,\dots,m+n$, we have
\begin{align*}
    \max_j D_{z z^{\prime}}^j D_{y y^{\prime}}^k \widehat{\gamma}_{k,u}^2 \leq \frac{2l\|y-y'\|}{n}\max\{\frac{1}{n-1}, \frac{1}{m}\}.
\end{align*}

Denote $m/(m+n)$ as $\rho_x$, and $n/(m+n)$ as $\rho_y$, where $\rho_x+\rho_y = 1$.
Based on the above results, we can provide upper bounds for seminorms $M(\widehat{\gamma}_{k,u}^2)$, $M_{\text{Lip}}(\widehat{\gamma}_{k,u}^2)$, and $J_{\text{Lip}}(\widehat{\gamma}_{k,u}^2)$ as follows.
\begin{align*}
    &M(\widehat{\gamma}_{k,u}^2) =(m+n)\max_k{D_{y y^{\prime}}^k \widehat{\gamma}_{k,u}^2} \leq 8\nu \max\{\frac{m+n}{m}, \frac{m+n}{n}\} = 8\nu \max\{\rho_x^{-1}, \rho_y^{-1}\},\\
    &M_{\text{Lip}}(\widehat{\gamma}_{k,u}^2) = (m+n)\max_k\sup_{y,y'}\frac{D_{y y^{\prime}}^k \widehat{\gamma}_{k,u}^2}{\|y-y'\|} \leq 2l \max\{\frac{m+n}{m}, \frac{m+n}{n}\} = 2l\max\{\rho_x^{-1}, \rho_y^{-1}\},\\
    &J_{\text{Lip}}(\widehat{\gamma}_{k,u}^2) = (m+n)^2\max_{k \neq l}\sup_{y,y', z, z'} \frac{D_{z z^{\prime}}^l D_{y y^{\prime}}^k \widehat{\gamma}_{k,u}^2}{\|y-y'\|}\\
    & \quad \quad \quad \quad \leq 2l \max \left\{\frac{(m+n)^2}{mn}, \frac{(m+n)^2}{m(m-1)}, \frac{(m+n)^2}{n(n-1)}\right\}\\
    & \quad \quad \quad \quad {\leq} 2l \max \left\{\frac{(m+n)^2}{m(m-1)}, \frac{(m+n)^2}{n(n-1)}\right\}\\
    & \quad \quad \stackrel{\frac{m}{m-1}, \frac{n}{n-1} \leq 2}{\leq} 2l\max\{\rho_x^{-2}, \rho_y^{-2}\}.
\end{align*}
Combining the above results with our main result (Theorem \ref{thm:general}), the inequality for MMD is derived.
\end{proof}

\subsection{Proof of Corollary \ref{coro:HSIC}}
\label{sec:proof-5.4}
\begin{proof}
    First, let $k$ be a reproducing kernel function defined as follows: 
    $$
    k((x,y), (x', y')) = k_\mathcal{X}(x, x')k_\mathcal{Y}(y, y'), \quad \forall x, x' \in \mathcal{X}, \, \forall y, y' \in \mathcal{Y}.
    $$
    We have
    \begin{align*}
        \sup_{x, x', y, y'}k((x, y), (x', y')) =& \sup_{x, x', y, y'}k_\mathcal{X}(x, x')k_\mathcal{Y}(y, y')\\
        \leq& \sup_{x, x'}k_\mathcal{X}(x, x')\sup_{y, y'} k_\mathcal{Y}(y, y')\\
        =& \nu_\mathcal{X}\nu_\mathcal{Y}.
    \end{align*}
    Second, we have
    \begin{eqnarray*}
        && M_{\text{Lip}}(k) \\
        &=& 2\max_{(x, y), (x', y'), (x_0, y_0)}\frac{k((x, y), (x_0, y_0)) - k((x', y'), (x_0, y_0))}{\|(x,y)-(x', y')\|}\\
        &=& 2\max_{(x, y), (x', y'), (x_0, y_0)}\frac{k_\mathcal{X}(x, x_0)k_\mathcal{Y}(y, y_0) - k_\mathcal{X}(x', x_0)k_\mathcal{Y}(y', y_0)}{\sqrt{\|x-x'\|^2 + \|y-y'\|^2}}.
    \end{eqnarray*}
    Note that:
    \begin{eqnarray*}
        &&k_\mathcal{X}(x, x_0)k_\mathcal{Y}(y, y_0) - k_\mathcal{X}(x', x_0)k_\mathcal{Y}(y', y_0)\\ 
        &=&k_\mathcal{Y}(y, y_0)\left(k_\mathcal{X}(x, x_0) - k_\mathcal{X}(x', x_0)\right) + k_\mathcal{X}(x', x_0) \left(k_\mathcal{Y}(y, y_0) - k_\mathcal{Y}(y', y_0)\right)\\
        &\leq& \nu_\mathcal{Y}\frac{\|x-x'\|}{2}M_{\text{Lip}}(k_\mathcal{X}) + \nu_\mathcal{X}\frac{\|y-y'\|}{2}M_{\text{Lip}}(k_\mathcal{Y})\\
        &=& \nu_\mathcal{Y}l_\mathcal{X}\frac{\|x-x'\|}{2} + \nu_\mathcal{X}l_\mathcal{Y}\frac{\|y-y'\|}{2}.
    \end{eqnarray*}
    Consequently, based on the above results, we have
    \begin{eqnarray*}
        && M_{\text{Lip}}(k) \\
        &\leq& \frac{\nu_\mathcal{Y}l_\mathcal{X}{\|x-x'\|} + \nu_\mathcal{X}l_\mathcal{Y}\|y-y'\|}{\sqrt{\|x-x'\|^2 + \|y-y'\|^2}}\\
        &\leq& \frac{\max\{\nu_\mathcal{Y}l_\mathcal{X}, \nu_\mathcal{X}l_\mathcal{Y}\}\left(\|x-x'\| + \|y-y'\| \right)}{\sqrt{\|x-x'\|^2 + \|y-y'\|^2}}\\
        &\leq&\max\{\nu_\mathcal{Y}l_\mathcal{X}, \nu_\mathcal{X}l_\mathcal{Y}\} \frac{\sqrt{2\left(\|x-x'\|^2 + \|y-y'\|^2\right)}}{\sqrt{\|x-x'\|^2 + \|y-y'\|^2}}\\
        &=& \sqrt{2}\max\{\nu_\mathcal{Y}l_\mathcal{X}, \nu_\mathcal{X}l_\mathcal{Y}\}.
    \end{eqnarray*}
\end{proof}

\subsection{Proof of Lemma \ref{ex:dk}}
\label{sec:proof-5.7}
\begin{proof}
    First, we have
    \begin{eqnarray*}
        k(u, u') &=& \frac{1}{2}\left[\left\|u-u_0\right\|+\left\|u^{\prime}-u_0\right\|-\left\|u-u^{\prime}\right\|\right]\\
        &\leq& \frac{1}{2}\left[\left\|u-u_0\right\|+\left\|u^{\prime}-u_0\right\|\right]\\
        &\leq& \frac{1}{2}\left[\nu_{\text{dis}}+\nu_{\text{dis}}\right]\\
        &=& \nu_{\text{dis}}.
    \end{eqnarray*}
    Secondly,
    \begin{eqnarray*}
        && M_\text{Lip}(k) \\
        &=& 2\sup_{u, u', v}\frac{k(u, v)-k(u',v)}{\|u-u'\|}\\
        &=& \sup_{u, u', v}\frac{\left[\left\|u-u_0\right\|+\left\|v-u_0\right\|-\left\|u-v\right\|\right]-\left[\left\|u'-u_0\right\|+\left\|v-u_0\right\|-\left\|u'-v\right\|\right]}{\|u-u'\|}\\
        &=& \sup_{u, u', v}\frac{\left\|u-u_0\right\|-\left\|u-v\right\|-\left\|u'-u_0\right\|+\left\|u'-v\right\|}{\|u-u'\|}\\
        &\leq&\frac{\|u-u'\| + \|u-u'\|}{\|u-u'\|}\\
        &=& 2,
    \end{eqnarray*}
    where the inequality step is derived based on the triangle's inequality of the norm $\|\cdot\|$.
\end{proof}

\subsection{Proof of Lemma \ref{ex:bilkernel}}\label{sec:proof-4}
\begin{proof}
Based on the formulation, suppose $\sup_{u \in \mathcal{U}}\|u\| \leq b_{\text{bi}}$, we have
\begin{align*}
    k(u, u') = u^Tu' \stackrel{\text{Cauchy Schwarz}}{\leq} \|u\|\|u'\| \leq b_{\text{bi}}^2, \quad \forall u, u'.
\end{align*}
Given $u, u_1, u_2 \in \mathcal{U}$, we have
\begin{eqnarray*}
    k(u, u_1) - k(u, u_2) &=& u^T(u_1-u_2)\\
    &\stackrel{\text{Cauchy Schwarz}}{\leq}& \|u_1-u_2\|\|u\|\\
    &\leq& b_{\text{bi}}\|u_1-u_2\|.
\end{eqnarray*}
Based on the definition of seminorm $M_\text{Lip}$ (Assumption \ref{assum:maurer}), we have $M_\text{Lip}(k) \leq 2b_{\text{bi}}$.
\end{proof}

\subsection{Proof of Lemma \ref{ex:polykernel}}\label{sec:proof-5}
\begin{proof}
To begin with, since $\sup_{u \in \mathcal{U}} \|u\| \leq b_{\text{pl}}$, we have the following inequality based on Cauchy Schwarz's inequality:
$$
    \sup_{u \in \mathcal{U}} u^Tu \leq b_{\text{pl}}^2.
$$
Since $k(u, u)$ is a monotone increasing function with respect to the quantity $u^Tu$, we have
$$
k(u, u) = (\alpha u^Tu + 1)^d \leq \left(\alpha b_{\text{pl}}^2 + 1\right)^d, \quad \forall u \in \mathcal{U}.
$$
Moreover, let $f(t):= (\alpha t + 1)^d$, we have $f'(t) = \alpha d(\alpha t + 1)^{d-1}$.
Since $u^Tu' \leq b_\text{pl}^2$, $\forall u, u' \in \mathcal{U}$, we have $f'(u^Tu') \leq \alpha d(\alpha b_\text{pl}^2 + 1)^{d-1}$.
Consequently, $\forall u, u_1, u_2 \in \mathcal{U}$, we have
\begin{eqnarray*}
    k(u, u_1) - k(u, u_2) &=& (\alpha u^Tu_1 + 1)^d - (\alpha u^Tu_2 + 1)^d\\
    &\leq& \alpha d(\alpha b_\text{pl}^2 + 1)^{d-1}\left|u^T(u_1-u_2)\right|\\
    &\stackrel{\text{Cauchy Schwarz's}}{\leq}& \alpha d(\alpha b_\text{pl}^2 + 1)^{d-1}\left\|u\right\|\left\|(u_1-u_2)\right\|\\
    &\stackrel{\sup_{u \in \mathcal{U}\|u\| \leq b_{\text{pl}}}}{\leq}& \alpha d b_{\text{pl}}(\alpha b_\text{pl}^2 + 1)^{d-1}\|u_1-u_2\|.
\end{eqnarray*}
Based on the definition of seminorm $M_\text{Lip}$ (Assumption \ref{assum:maurer}), we have 
$$
    M_\text{Lip}(k) \leq 2\alpha d b_{\text{pl}}(\alpha b_\text{pl}^2 + 1)^{d-1}.
$$
\end{proof}

\subsection{Proof of Lemma \ref{lem:sigmoid}}\label{sec:proof-6}

\begin{proof}
    Recall the $\operatorname{tanh}$ function, i.e.,
\begin{align*}
    \operatorname{tanh}(x) = \frac{e^{2 x}-1}{e^{2 x}+1}.
\end{align*}
Consequently, 
\begin{align*}
    \operatorname{tanh}(\alpha u^Tu' + \beta) \leq 1, \quad \forall u, u'.
\end{align*}
Moreover, we have
\begin{align*}
    \frac{d}{dx}\operatorname{tanh}(x) =& \frac{2e^{2x}[(e^{2x}+1)-(e^{2x}-1)]}{(e^{2 x}+1)^2}\\
    =& \frac{2e^{2x}[(e^{2x}+1)-(e^{2x}-1)]}{(e^{2 x}+1)^2}\\
    =& \frac{4}{(e^{2x} + 1)(e^{-2x} + 1)}.
\end{align*}
Regarding the term $(e^{2x} + 1)(e^{-2x} + 1)$, it is observable that it reaches its minimum at the point $x = 0$, where it equals to $4$.
Consequently, we have $\frac{d}{dx}\operatorname{tanh}(x) \leq 1$.
Given $u, u_1, u_2 \in \mathcal{U}$, we have
\begin{eqnarray*}
    &&\operatorname{tanh}(\alpha u^Tu_1 + \beta) - \operatorname{tanh}(\alpha u^Tu_2 + \beta)\\
    &\leq &\left|(\alpha u^Tu_1 + \beta) - (\alpha u^Tu_2 + \beta)\right|\\
    &=& \alpha |u^T(u_1-u_2)|\\
    &\stackrel{\text{Cauchy Schwarz}}{\leq} & \alpha\|u\|\|u_1-u_2\|\\
    &\leq& \alpha b_\text{sig}\|u_1-u_2\|.
\end{eqnarray*}
Based on the definition of seminorm $M_\text{Lip}$ (Assumption \ref{assum:maurer}), we have $M_\text{Lip}(k) \leq 2\alpha b_\text{sig}$.
\end{proof}

\subsection{Proof of Lemma \ref{lem:translation}}\label{sec:proof-7}

\begin{proof}
    From the definition of function $\widetilde{k}$, we have
    \begin{align*}
        \sup_{u \in \mathcal{U}} k(u,u) = \sup_{u \in \mathcal{U}}\widetilde{k}(u-u)
        =\widetilde{k}(\mathbf{0}_d)
        \leq \nu_{\text{t}}.
    \end{align*}
    For the second parts, $\forall u \in \mathcal{U}$, consider the function $k(u, \cdot)$, $\forall u_1, u_2 \in \mathcal{U}$, we have
    \begin{align*}
        k(u, u_1)-k(u, u_2) =& \widetilde{k}(u-u_1)-\widetilde{k}(u-u_2)\\
        \leq& l_{\text{t}} \|u_1-u_2\|,
    \end{align*}
    where the first equation is given based on the definition of function $\widetilde{k}$, and the second one is derived based on the Lipschitz assumption of the function $\widetilde{k}$. That is, $\forall u \in \mathcal{U}$, the function $k(u, \cdot)$ is $l_{\text{t}}$-Lipschitz.
    Consequently, based on the definition of seminorm $M_\text{Lip}$ (Assumption \ref{assum:maurer}), we have $M_\text{Lip}(k) \leq 2l_t$.
\end{proof}

\subsection{Proof of Lemma \ref{ex:Gk}}\label{sec:proof-8}

\begin{proof}
    From the definition of the Gaussian kernel, it could be observed that it is a translation-invariant kernel, equipped with the function: $\widetilde{k}(u)=\exp\left(-\sigma^{-2}\|u\|^2\right)$, $\forall u \in \mathcal{U}$.
Regarding this function, we have:
\begin{align*}
    \exp\left(-\sigma^{-2}\|\mathbf{0}_d\|^2\right) \leq 1.
\end{align*}
Moreover, considering the derivative $\nabla\widetilde{k}$, we have $\left\|\nabla\widetilde{k}(u)\right\| = \frac{2}{\sigma^2}\|u\| \exp \left(-\sigma^{-2}\|u\|^2\right).$
Denote the term $\|u\|/\sigma$ as $t$, then the derivative $\left\|\nabla\widetilde{k}(u)\right\| = {2}t\exp(-t^2)/\sigma$. Again, taking derivative of the term $2t\exp(-t^2)/\sigma$ with respect to $t$, we have
\begin{align*}
    \frac{d}{dt}\frac{2t\exp(-t^2)}{\sigma} =& \frac{2}{\sigma}\left(\exp(-t^2) - 2t^2\exp(-t^2)\right),
\end{align*}
which means that $\left\|\nabla\widetilde{k}(u)\right\|$ takes its supremum when $t = 1/\sqrt{2}$, that is, we have
\begin{align*}
\left\|\nabla\widetilde{k}(u)\right\| \leq \frac{\sqrt{2}}{\sigma}\exp(-1/2).
\end{align*}
Adopting the results built for the translation invariant kernel, the result for the Gaussian kernel is derived.
\end{proof}

\subsection{Proof of Lemma \ref{ex:Lk}}\label{sec:proof-9}
\begin{proof}
    It could be observed that the Laplacian kernel is a translation-invariant kernel, equipped with the function $\widetilde{k}(u)=\exp(-\sigma^{-1}\|u\|_1)$. Similarly, it could be derived that $\widetilde{k}(u)=\exp(-\sigma^{-1}\|\mathbf{0}\|_1) \leq 1$.
    In addition, $\forall u:=(u_1,\dots,u_d) \in \mathcal{U}$, the element-wise derivative of $\widetilde{k}$ is considered as follows:
    \begin{align*}
        \left|\frac{d}{du_i} \exp(-\sigma^{-1}\|u\|_1)\right| \leq \sigma^{-1}\exp(-\sigma^{-1}\|u\|_1) \leq \sigma^{-1}.
    \end{align*}
    As a consequence, the norm of the derivative is less than or equal to $\sigma^{-1}\sqrt{d}$. That is, the Lipschitz constant is equal to $\sigma^{-1}\sqrt{d}$.
\end{proof}

\subsection{Proof of Corollary \ref{ex:DISCA}}
\label{sec:proof-cor6.1}
\begin{proof}
    We first confirm the upper bound for the Gaussian complexity, i.e., $\mathbb{E}\mathcal{G}(P_{S^{p-1}}\mathbf{X}, \mathbf{Y})$.
    That is,
    \begin{eqnarray*}
        &&\mathbb{E}\mathcal{G}(P_{S^{p-1}}\mathbf{X}, \mathbf{Y})\\ 
        &=& \mathbb{E}\left[\sup_{u \in \mathcal{S}}\sum_{i=1}^n\langle \xi_i, (u^TX_i, Y_i) \rangle\right]\\
        &=& \mathbb{E}\left[\sup_{u \in \mathcal{S}}\sum_{i=1}^n\langle \xi_{i,x}, u^TX_i \rangle + \langle \xi_{i,y}, Y_i \rangle\right]\\
        &=& \mathbb{E}\left[\sup_{u \in \mathcal{S}}\sum_{i=1}^n\langle \xi_{i,x}, u^TX_i \rangle \right]\\
        &=& \mathbb{E}\sup_{u \in \mathcal{S}} \left[u^T \sum_{i=1}^n X_i \xi_{i,x} \right]\\
        &\stackrel{\text{Cauchy Schwarz's Inequality}}{\leq}& \mathbb{E}\sup_{u \in \mathcal{S}} \left[\|u\| \cdot \left\|\sum_{i=1}^n X_i \xi_{i,x} \right\|\right]\\
        &=& \mathbb{E} \left\|\sum_{i=1}^n X_i \xi_{i,x} \right\|\\
        &=& \mathbb{E}\sqrt{\sum_{j=1}^p\left(\sum_{i=1}^n X_{i,j} \xi_{i,x}\right)^2}\\
        &\stackrel{\text{Jensen's Inequality}}{\leq}& \sqrt{\sum_{j=1}^p\mathbb{E}\left(\sum_{i=1}^n X_{i,j} \xi_{i,x}\right)^2}\\
        &\stackrel{\mathbb{E}\xi_{{i_1},x}\xi_{{i_2},x} = 0}{=}& \sqrt{\sum_{j=1}^p\mathbb{E}\sum_{i=1}^n \left(X_{i,j} \xi_{i,x}\right)^2}\\
        &=& \sqrt{\sum_{j=1}^p\sum_{i=1}^n X_{i,j}^2}\\
        &=& \sqrt{\sum_{i=1}^n \|X_i\|^2}\\
        &\stackrel{\sup_{x \in \mathcal{X}}\|x\| \leq \nu_{\text{dis}}}{\leq}& \sqrt{n \nu_{\text{dis}}^2}\\
        &=&\sqrt{n} \nu_{\text{dis}}.
    \end{eqnarray*}
    Combining the above result with Corollary \ref{cor:dCov}, the uniform concentration inequality for the function set defined in Corollary \ref{ex:DISCA} is proved.
    Based on Proposition \ref{prop:intro}, the proof of Corollary \ref{ex:DISCA} is completed.
\end{proof}

\subsection{Proof of Corollary \ref{ex:ICA}}
\label{sec:proof-cor6.2}
\begin{proof}
    Recall $\boldsymbol{S}:= \mathbf{Y}\mathbf{W}_{\theta}^T = (\boldsymbol{S}_1(\theta), \dots, \boldsymbol{S}_d(\theta))$.
    As stated on \cite[p. 5]{matteson2017independent}, the specific formulation of the rotation matrix $\mathbf{W}_\theta$ is given as follows:
    $$
        \mathbf{W}_\theta=\mathbf{Q}^{(d-1)} \cdots \mathbf{Q}^{(1)}, \quad \text { in which } \quad \mathbf{Q}^{(k)}=\mathbf{Q}_{k, d}\left(\theta_{k, d}\right) \cdots \mathbf{Q}_{k, k+1}\left(\theta_{k, k+1}\right) ,
    $$
    where $\mathbf{Q}_{i,j}(\psi)$ corresponds to the Givens (plane) rotation matrix equipped with position $(i,j)$ and angle $\psi$, that is, the identity matrix $\boldsymbol{I}_d$ with the $(i, i)$ and $(j, j)$ elements replaced by $\cos (\psi)$, the $(i, j)$ element replaced by $-\sin (\psi)$, and the $(j, i)$ element replaced by $\sin (\psi)$.
    
    Regarding matrix $(\boldsymbol{S}_k, \boldsymbol{S}_{k+1}, \dots, \boldsymbol{S}_{d})$, we have
    \begin{align*}
        \boldsymbol{S} \begin{pmatrix}
            \boldsymbol{0}_{(k-1)\times (d-k+1)}\\
            I_{d-k+1}
        \end{pmatrix} =& (\boldsymbol{S}_1, \dots, \boldsymbol{S}_d) \begin{pmatrix}
            \boldsymbol{0}_{(k-1)\times (d-k+1)}\\
            \boldsymbol{I}_{d-k+1}
        \end{pmatrix}\\
        =& (\boldsymbol{S}_k, \dots, \boldsymbol{S}_d).
    \end{align*}
    Consequently, we have
    \begin{eqnarray*}
        \mathbb{E}\left[\sup_{\theta^{(k:k)} \in \Theta_k}\sum_{i=1}^n \langle \xi_i, (\boldsymbol{S}_k, \boldsymbol{S}_{k^+})_i\rangle\right]
        &=& \mathbb{E}\left[\sup_{\theta^{(k:k)} \in \Theta_k} \sum_{i=1}^n Y_i^T\mathbf{W}_\theta \begin{pmatrix}
            \boldsymbol{0}_{(k-1)\times (d-k+1)}\\
            \boldsymbol{I}_{d-k+1}
        \end{pmatrix} \xi_i \right]
    \end{eqnarray*}
    It can be observed that $\mathbf{Q}^{(d-1)} \cdots \mathbf{Q}^{(k)}$ can be expressed as the following matrix:
    $$
    \begin{pmatrix}
        \boldsymbol{I}_{k-1} & \\
        & \mathbf{W}_{k^+}
    \end{pmatrix},
    $$
    where $\mathbf{W}_{k^+}$ is a $\mathcal{SO}(d-k+1)$ matrix.
    Consequently, we have
    \begin{eqnarray*}
        &&Y_i^T\mathbf{W}_\theta \begin{pmatrix}
            \boldsymbol{0}_{(k-1)\times (d-k+1)}\\
            I_{d-k+1}
        \end{pmatrix} \xi_i\\
        &=& Y_i^T\mathbf{Q}^{(d-1)}\cdots \mathbf{Q}^{(1)} \begin{pmatrix}
            \boldsymbol{0}_{(k-1)\times (d-k+1)}\\
            I_{d-k+1}
        \end{pmatrix} \xi_i\\
        &=& Y_i^T\begin{pmatrix}
        \boldsymbol{I}_{k-1} & \\
        & \mathbf{W}_{k^+}
    \end{pmatrix} \mathbf{Q}^{(k-1)}\cdots \mathbf{Q}^{(1)} \begin{pmatrix}
            \boldsymbol{0}_{(k-1)\times (d-k+1)}\\
            I_{d-k+1}
        \end{pmatrix} \xi_i.
    \end{eqnarray*}
    Note that the variation of $\theta^{(k:k)}$ will only lead to the variation of matrix $\mathbf{Q}^{(k)}$, which is only reflected on $\mathbf{W}_{k^+}$.
    Moreover, since $\mathbf{Q}^{(k-1)}\cdots \mathbf{Q}^{(1)}$ is an orthogonal matrix, we decompose it as follows for simplicity:
    \begin{align*}
        \mathbf{Q}^{(k-1)}\cdots \mathbf{Q}^{(1)} := \left(\boldsymbol{O}_1, \boldsymbol{O}_2\right), \quad \text{where }\boldsymbol{O}_1 \in \mathbb{R}^{d \times (k-1)}, \,\boldsymbol{O}_2 \in \mathbb{R}^{d \times (d-k+1)}.
    \end{align*}
    Here, note that $\boldsymbol{O}_2^T\boldsymbol{O}_2 = \boldsymbol{I}_{d-k+1}$.
    Consequently, let $\mathbf{Y}:=(Y_1, \dots, Y_n)^T \in \mathbb{R}^{n \times d}$, $\boldsymbol{\xi}:= (\xi_1, \dots, \xi_n)^T \in \mathbb{R}^{n \times (d-k+1)}$. 
    We have
    \begin{eqnarray*}
        &&\mathbb{E}\left[\sup_{\theta^{(k:k)} \in \Theta_k}\sum_{i=1}^n \langle \xi_i, (\boldsymbol{S}_k, \boldsymbol{S}_{k^+})_i \rangle \right]\\
        &=& \mathbb{E}\left[\sup_{\theta^{(k:k)} \in \Theta_k} \sum_{i=1}^n Y_{i}^T
        \begin{pmatrix}
            \boldsymbol{I}_{k-1} & \\
            & \mathbf{W}_{k^+}
        \end{pmatrix}
        \left(\boldsymbol{O}_1, \boldsymbol{O}_2\right)
        \begin{pmatrix}
            \boldsymbol{0}_{(k-1)\times (d-k+1)}\\
            I_{d-k+1}
        \end{pmatrix}
        \xi_i\right]\\
        &=& \mathbb{E}\left[\sup_{\theta^{(k:k)} \in \Theta_k} \operatorname{tr}\left(\mathbf{Y}
        \begin{pmatrix}
            \boldsymbol{I}_{k-1} & \\
            & \mathbf{W}_{k^+}
        \end{pmatrix}
        \left(\boldsymbol{O}_1, \boldsymbol{O}_2\right)
        \begin{pmatrix}
            \boldsymbol{0}_{(k-1)\times (d-k+1)}\\
            I_{d-k+1}
        \end{pmatrix} \boldsymbol{\xi}^T\right)\right]\\
        &\stackrel{\operatorname{tr}(\mathbf{AB}) = \operatorname{tr}(\mathbf{BA})}{=}& \mathbb{E}\left[\sup_{\theta^{(k:k)} \in \Theta_k} \operatorname{tr}\left(
        \begin{pmatrix}
            \boldsymbol{0}_{(k-1)\times n}\\
            \boldsymbol{\xi}^T
        \end{pmatrix} \mathbf{Y}
        \begin{pmatrix}
            \boldsymbol{I}_{k-1} & \\
            & \mathbf{W}_{k^+}
        \end{pmatrix}
        \left(\boldsymbol{O}_1, \boldsymbol{O}_2\right)
        \right)\right]\\
        &=& \mathbb{E}\left[\sup_{\theta^{(k:k)} \in \Theta_k} \operatorname{tr}\left(
        \boldsymbol{\xi}^T \mathbf{Y}
        \begin{pmatrix}
            \boldsymbol{I}_{k-1} & \\
            & \mathbf{W}_{k^+}
        \end{pmatrix}
        \boldsymbol{O}_2
        \right)\right]\\
        &=& \mathbb{E}\left[\sup_{\theta^{(k:k)} \in \Theta_k} \operatorname{tr}\left(\boldsymbol{O}_2^T
        \begin{pmatrix}
            \boldsymbol{I}_{k-1} & \\
            & \mathbf{W}_{k^+}^T
        \end{pmatrix}
        \mathbf{Y}^T\boldsymbol{\xi}
        \right)\right].
    \end{eqnarray*}
    Denote the matrix 
    $\boldsymbol{O}_2^T
    \begin{pmatrix}
        \boldsymbol{I}_{k-1} & \\
        & \mathbf{W}_{k^+}^T
    \end{pmatrix}
    \mathbf{Y}^T \in \mathbb{R}^{(d-k+1) \times n}$ as $\boldsymbol{V}_0^T$, it can be observed that $\boldsymbol{V}_0^T\boldsymbol{V}_0 = \boldsymbol{I}_{d-k+1}$.
    Consequently, we have
    \begin{eqnarray*}
    \mathbb{E}\left[\sup_{\theta^{(k:k)} \in \Theta_k} \operatorname{tr}\left(\boldsymbol{O}_2^T
        \begin{pmatrix}
            \boldsymbol{I}_{k-1} & \\
            & \mathbf{W}_{k^+}^T
        \end{pmatrix}
        \mathbf{Y}^T\boldsymbol{\xi}
        \right)\right]
        = \mathbb{E}\left[\sup_{\theta^{(k:k)} \in \Theta_k} \operatorname{tr}\left(\boldsymbol{V}_0^T\boldsymbol{\xi}
        \right)\right].
    \end{eqnarray*}
    From \textit{von Neumann's trace inequality}, we have
    \begin{align*}
        \operatorname{tr}\left(\boldsymbol{V}_0^T\boldsymbol{\xi}
        \right) \leq \sum_{i=1}^{d-k+1} \sigma_i(\boldsymbol{\xi}),
    \end{align*}
    where $\sigma_i(\boldsymbol{\xi})$ represents the $i$th largest singular value for matrix $\boldsymbol{\xi}$.
    Since the nuclear norm of a matrix is smaller than or equal to the Frobenius norm of a matrix, we have
    \begin{eqnarray*}
        \mathbb{E}\left[\sup_{\theta^{(k:k)} \in \Theta_k} \operatorname{tr}\left(\boldsymbol{V}_0^T\boldsymbol{\xi}
        \right)\right] &\leq& \mathbb{E} \|\boldsymbol{\xi}\|_F\\
        &\stackrel{\text{Jensen's inequality}}{\leq}& \sqrt{\mathbb{E}\sum_{i=1}^n \sum_{j=1}^{d-k+1} \xi_{i,j}^2}\\
        &=& \sqrt{n (d-k+1)}.
    \end{eqnarray*}
    Substituting the upper bound for the Gaussian complexity into the uniform concentration inequality for dCov, i.e., Corollary \ref{cor:dCov}, the uniform concentration inequality for the aforementioned function class for this example is derived.
    Afterwards, employing Proposition \ref{prop:intro} leads to the listed inequality.
\end{proof}

\subsection{Proof of Corollary \ref{cor:generalizedMMD}}
\label{sec:proof-cor6.3}
\begin{proof}
    Notice that $\gamma^2(X, Y; \mathcal{K})$ in this specified case is equivalent to $\sup_{h \in \mathcal{H}}\gamma_k^2(h_\mathcal{X}(X), h_\mathcal{Y}(Y))$, where $k$ refers to the Gaussian kernel with bandwidth $1$, i.e, $k(u, u') = \exp(\|u-u'\|^2)$, $h(X, Y) = (X/\sigma, Y/\sigma)$, $\sigma \in [\sigma_l, \sigma_h]$.
    Consider the Gaussian complexity of this function class, that is, 
    \begin{eqnarray*}
        \mathcal{G}(\mathcal{H}(\mathbf{X}, \mathbf{Y})) &=& \mathbb{E}_{\xi}\sup_{\sigma \in [\sigma_l, \sigma_h]} \left[\sum_{i=1}^m \langle \xi_{i,x}, X_i/\sigma\rangle + \sum_{j=1}^n\langle \xi_{j,y}, Y_j/\sigma\rangle\right]\\
        &=& \mathbb{E}_{\xi}\sup_{\sigma \in [\sigma_l, \sigma_h]}\frac{1}{\sigma} \left[\sum_{i=1}^m \langle \xi_{i,x}, X_i\rangle + \sum_{j=1}^n\langle \xi_{j,y}, Y_j \rangle\right].
    \end{eqnarray*}
    Note that $\sum_{i=1}^m\langle \xi_{i,x}, X_i\rangle + \sum_{j=1}^n \langle \xi_{j,y}, Y_j\rangle \sim \mathcal{N}\left(0, \sum_{i=1}^mX_i^TX_i + \sum_{j=1}^n Y_j^TY_j\right)$.
    Consequently, let $\xi$ be a random variable following standard normal distribution, we have
    \begin{eqnarray*}
         \mathcal{G}(\mathcal{H}(\mathbf{X}, \mathbf{Y})) &=& \left[\sum_{i=1}^mX_i^TX_i + \sum_{j=1}^n Y_j^TY_j\right]^{1/2} \mathbb{E}_{\xi}\sup_{\sigma \in [\sigma_l, \sigma_h]}\frac{\xi}{\sigma}\\
         &=& \left[\sum_{i=1}^mX_i^TX_i + \sum_{j=1}^n Y_j^TY_j\right]^{1/2} \mathbb{E}_{\xi} \left[I_{\xi \geq 0} / \sigma_l + I_{\xi < 0} / \sigma_r\right]\xi\\
         &=& \left[\sum_{i=1}^mX_i^TX_i + \sum_{j=1}^n Y_j^TY_j\right]^{1/2} \mathbb{E}_{\xi} \left[\left(\frac{1}{\sigma_l} - \frac{1}{\sigma_r}\right)I_{\xi \geq 0} + \frac{1}{\sigma_r}\right]\xi\\
         &=& \left[\sum_{i=1}^mX_i^TX_i + \sum_{j=1}^n Y_j^TY_j\right]^{1/2}\left(\frac{1}{\sigma_l} - \frac{1}{\sigma_r}\right)\frac{1}{\sqrt{2\pi}}.
    \end{eqnarray*}
    Take expectations of the above quantity, we have
    \begin{eqnarray*}
        \mathbb{E}\left[\mathcal{G}(\mathcal{H}(\mathbf{X}, \mathbf{Y}))\right] &\leq& \frac{1}{\sqrt{2\pi}}\left(\frac{1}{\sigma_l} - \frac{1}{\sigma_r}\right)\mathbb{E}\left[\sum_{i=1}^mX_i^TX_i + \sum_{j=1}^n Y_j^TY_j\right]^{1/2}\\
        &\stackrel{\text{Jensen's Inequality}}{\leq}& \frac{1}{\sqrt{2\pi}}\left(\frac{1}{\sigma_l} - \frac{1}{\sigma_r}\right)\left[\mathbb{E}\sum_{i=1}^mX_i^TX_i + \mathbb{E}\sum_{j=1}^n Y_j^TY_j\right]^{1/2}\\
        &=& \frac{1}{\sqrt{2\pi}}\left(\frac{1}{\sigma_l} - \frac{1}{\sigma_r}\right) \sqrt{m\mathbb{E}\|X\|^2 + n \mathbb{E}\|Y\|^2}.
    \end{eqnarray*}
    Consequently, substitute the constants $\nu$ and $l$ regarding the reproducing kernel $k$ with the ones for the Gaussian kernel equipped with bandwidth $1$ (see Table \ref{tab:kernels}), and the function set $\mathcal{H}$ in Corollary \ref{coro:MMD} with the one stated above, the upper bound for $\widehat{\gamma}_{k, u}$ in Corollary \ref{coro:MMD} is expressed as follows:
    \begin{align*}
        C_1\sqrt{\frac{\log (2 / \delta)}{m+n}}
        + {8C_0\sqrt{2/e}}\left(\frac{1}{\sigma_l} - \frac{1}{\sigma_r}\right)\sqrt{\frac{\rho_x\mathbb{E}\|X\|^2 + \rho_y \mathbb{E}\|Y\|^2}{m+n}},
    \end{align*}
    where $C_0 = \max\bigg\{\rho_y^{-1}\left(1 + \rho_y^{-1}\right), \rho_x^{-1}\left(1 + \rho_x^{-1}\right)\bigg\}$, $C_1 = 8\max\left\{\rho_x^{-1}, \rho_y^{-1}\right\}$.
    
    \noindent Consequently, let $C_2 = {8C_0\sqrt{2/e}}\left(\frac{1}{\sigma_l} - \frac{1}{\sigma_r}\right)\sqrt{{\rho_x\mathbb{E}\|X\|^2 + \rho_y \mathbb{E}\|Y\|^2}}$, the upper bound can be expressed as follows:
    \begin{align*}
        C_1\sqrt{\frac{\log (2 / \delta)}{m+n}} + \frac{C_2}{\sqrt{m+n}}.
    \end{align*}
    Utilizing Proposition \ref{prop:intro}, the conclusion in Corollary \ref{cor:generalizedMMD} is derived.
\end{proof}

\subsection{Proof of Corollary \ref{coro:mmd(m,n)}}
\label{sec:proof-cor6.5}
\begin{proof}
    In the following, we first take the U-statistic estimator $\widehat{\gamma}_{k,e}$ into account.
To derive the generalization error bound for $\widehat{\theta}_{n,n}$, we first break the difference into two terms. That is,
\begin{align*}
    &\gamma_k^2\left(h_{\widehat{\theta}_{n,n}}(X), Y\right) - \inf_{\theta \in \Theta}\gamma_k^2(h_{\theta}(X), Y)\\
    =& \underbrace{\gamma_k^2\left(h_{\widehat{\theta}_{n,n}}(X), Y\right) - \widehat{\gamma}_k^2\left(h_{\widehat{\theta}_{n,n}}(\mathbf{X}), \mathbf{Y}\right)}_{(\star)_1} + \underbrace{\inf_{\theta\in \Theta}\widehat{\gamma}_k^2(h_\theta(\mathbf{X}), \mathbf{Y}) - \inf_{\theta \in \Theta}\gamma_k^2(h_\theta(X), Y)}_{(\star)_2}.
\end{align*}
Here, the above formula is derived based on the definition (\ref{eq:intro_MDE(MMD,m,n)}) of $\widehat{\theta}_{n,n}$. That is,
$$
\widehat{\gamma}_k^2(h_{\widehat{\theta}_{n,n}}(\mathbf{X}), \mathbf{Y})
= \inf_{\theta \in \Theta}\widehat{\gamma}_k^2(h_\theta(X), Y).
$$
Since $\widehat{\theta}_{n,n} \in \Theta$, the following inequality holds:
\begin{align*}
    (\star)_1 \leq \sup_{\theta \in \Theta} \left|\gamma_k^2\left(h_{\theta}(X), Y\right) - \widehat{\gamma}_k^2\left(h_{\theta}(\mathbf{X}), \mathbf{Y}\right)\right|.
\end{align*}
Regarding the second term $(\star)_2$, based on Proposition \ref{prop:intro}, we have
\begin{align*}
    (\star)_2 \leq \sup_{\theta \in \Theta} \left|\gamma_k^2\left(h_{\theta}(X), Y\right) - \widehat{\gamma}_k^2\left(h_{\theta}(\mathbf{X}), \mathbf{Y}\right)\right|.
\end{align*}
Let the following function class be the one mentioned in Corollary \ref{coro:MMD}:
\begin{align}
\label{eq:intro_G_MDE}
    \mathcal{H}:=\{(g_{\theta}, \operatorname{id}_{\mathbb{R}^q}) \mid \theta \in \Theta\},
\end{align}
where $\operatorname{id}_{\mathbb{R}^q}$ is the identity map in $\mathbb{R}^q$. In that case, Corollary \ref{coro:MMD} shows that: $\forall \delta \in (0,1)$, with probability at least $1-\delta$, we have
\begin{align*}
    \sup_{\theta \in \Theta} \left|\gamma_k^2\left(h_{\theta}(X), Y\right) - \widehat{\gamma}_k^2\left(h_{\theta}(\mathbf{X}), \mathbf{Y}\right)\right| \leq C(n, \delta).
\end{align*}
Combining the above result with the upper bounds of $(\star)_1$ and $(\star)_2$, the generalization bound of estimator $\widehat{\theta}_{n,n}$ is given as follows: $\forall \delta \in (0,1)$, with probability at least $1-\delta$, we have
\begin{align}
    \gamma_k^2\left(h_{\widehat{\theta}_{m,n}}(X), Y\right) \leq \inf_{\theta \in \Theta}\gamma_k^2(h_{\theta}(X), Y) + 2C(n, \delta).
\end{align}
Replace the constant $C(n, \delta)$ with the upper bound presented in Corollary \ref{coro:MMD}, the inequality for the U-statistic estimator is derived.
Suppose the sample size $m \neq n$, and the unbiased estimate $\widehat{\gamma}_{k,u}^2$ or the biased estimate $\widehat{\gamma}_{k,b}^2$ is employed in the formulation of the estimator $\widehat{\theta}_{m,n}$ \eqref{eq:intro_MDE(MMD,m,n)}. 
The above generalization error bound for $\widehat{\gamma}_{k,u}^2$ and $\widehat{\gamma}_{k,b}^2$  can be derived in similar steps.

For the error bound of the minimized empirical MMD, i.e., $\min_{\theta \in \Theta}\widehat{\gamma}_k(h_{\theta}(\mathbf{X}), \mathbf{Y})$, according to Proposition \ref{prop:intro}, we have
\begin{align*}
    \left|\min_{\theta \in \Theta}\widehat{\gamma}_k^2(h_{\theta}(\mathbf{X}), \mathbf{Y}) - \min_{\theta \in \Theta}\gamma_k^2(g_\theta(X), Y)\right|
    \leq \sup_{\theta \in \Theta}\left|\widehat{\gamma}_k^2(h_{\theta}(\mathbf{X}), \mathbf{Y}) - \gamma_k^2(g_\theta(X), Y)\right|.
\end{align*}
Consequently, our main result can be directly employed to derive the estimation error bounds.
\end{proof}

\subsection{Proof of Corollay \ref{coro:MMDGAN}}
\label{sec:proof-cor6.6}
\begin{proof}
    To begin with, we replace the function class mentioned in Corollay \ref{coro:MMD} with the following one, i.e.,
\begin{align*}
    \{(f\circ h, f) \mid (f\circ h, f)(\mathbf{X}, \mathbf{Y}) = f(h(X_1), \dots, h(X_m), Y_1,\dots, Y_n), h \in \mathcal{H}, f \in \mathcal{F}\},
\end{align*}
the left-hand-side of Corollay \ref{coro:MMD} is given as follows:
\begin{align*}
    &\sup_{h \in \mathcal{H}, f \in \mathcal{F}}\left|\widehat{\gamma}_{k, u}^2(f(h(\mathbf{X})), f(\mathbf{Y}))-{\gamma}_{k}^2(f(h(X)), f(Y))\right|\\
    \stackrel{\text{Definition of }k\circ f}{=}&\sup_{h \in \mathcal{H}, f \in \mathcal{F}}\left|\widehat{\gamma}_{k\circ f, u}^2(h(\mathbf{X}), \mathbf{Y})-{\gamma}_{k \circ f}^2(h(X), Y)\right|.
\end{align*}
Similar to the steps mentioned in Section \ref{sec:proof-cor6.5}, the Gaussian complexity can be replaced by the following summation:
\begin{align*}
    \mathbb{E}\left[\mathcal{G}(\mathcal{F}\circ \mathcal{H}(\mathbf{X}))\right] + \mathbb{E}\left[\mathcal{G}(\mathcal{F}(\mathbf{Y}))\right].
\end{align*}
Consequently, take the unbiased estimate $\widehat{\gamma}_{k, u}$ as an illustrative example, $\forall \delta \in (0, 1)$, with probability at least $1-\delta$, we have
\begin{align*}
    \sup_{h \in \mathcal{H}, f \in \mathcal{F}}\left|\widehat{\gamma}_{k\circ f, u}^2(h(\mathbf{X}), \mathbf{Y})-{\gamma}_{k \circ f}^2(h(X), Y)\right| \leq 
    8\nu \max\left\{\rho_x^{-1}, \rho_y^{-1}\right\}\sqrt{\frac{\log (2 / \delta)}{m+n}}
    + \\
    \frac{4\sqrt{2\pi}l}{m+n}\max\bigg\{\rho_y^{-1}
    \left(1 + \rho_y^{-1}\right), \rho_x^{-1}\left(1 + \rho_x^{-1}\right)\bigg\} \left(\mathbb{E}\left[\mathcal{G}(\mathcal{F}\circ \mathcal{H}(\mathbf{X}))\right] + \mathbb{E}\left[\mathcal{G}(\mathcal{F}(\mathbf{Y}))\right]\right).
\end{align*}
    Note that the difference $\gamma_{k \circ f^*}^2\left(h^*(X), Y\right) - \inf_{f \in \mathcal{F}}\sup_{h \in \mathcal{H}}\gamma_{k \circ f}^2\left(h(X), Y\right)$ is equivalent to the following expression:
    \begin{align*}
        \gamma_{k \circ f^*}^2\left(h^*(X), Y\right)& - \inf_{h \in \mathcal{H}}\sup_{f \in \mathcal{F}}\widehat{\gamma}_{k\circ f}^2\left(h(\mathbf{X}), \mathbf{Y}\right)\\
        &+ \inf_{h \in \mathcal{H}}\sup_{f \in \mathcal{F}}\widehat{\gamma}_{k\circ f}^2\left(h(\mathbf{X}), \mathbf{Y}\right) - \inf_{h \in \mathcal{H}}\sup_{f \in \mathcal{F}}\gamma_{k \circ f}^2\left(h(X), Y\right).
    \end{align*}
    In addition, similar to the procedure stated in the proof of Corollary \ref{coro:mmd(m,n)}, the following inequality is an observable conclusion: 
    \begin{align*}
        \gamma_{k \circ f^*}^2\left(g^*(X), Y\right) - &\inf_{h \in \mathcal{H}}\sup_{f \in \mathcal{F}}\widehat{\gamma}_{k\circ f}^2\left(h(\mathbf{X}), \mathbf{Y}\right)\\ 
        &\leq \sup_{f\in \mathcal{F}, h \in \mathcal{H}} \left|\widehat{\gamma}^2_{k \circ f}\left(h(\mathbf{X}), \mathbf{Y}\right)-\gamma^2_{k \circ f}(h(X), Y)\right|.
    \end{align*}
    That is, the first two terms mentioned in Corollary \ref{coro:MMDGAN} can be bounded based on the above observation.
    Consequently, the proof of the generalization bound is reduced to the proof of an upper bound of the term $\inf_{h \in \mathcal{H}}\sup_{f \in \mathcal{F}}\widehat{\gamma}_{k\circ f}^2\left(h(\mathbf{X}), \mathbf{Y}\right) - \inf_{h \in \mathcal{H}}\sup_{f \in \mathcal{F}}{\gamma}_{k\circ f}^2\left(h(\mathbf{X}), \mathbf{Y}\right)$.
    In the following, we abbreviate $\widehat{\gamma}_{k\circ f}^2\left(h(\mathbf{X}), \mathbf{Y}\right)$ as $\widehat{\gamma}_{k \circ f}^2$, and $\gamma_{k \circ f}^2\left(h(X), Y\right)$ as $\gamma_{k \circ f}^2$ for simplicity, then we have
    \begin{eqnarray*}
        \inf_{h \in \mathcal{H}}\sup_{f \in \mathcal{F}} \widehat{\gamma}_{k \circ f}^2 - \inf_{h \in \mathcal{H}}\sup_{f \in \mathcal{F}} {\gamma}_{k \circ f}^2 &=& \inf_{h \in \mathcal{H}}\sup_{f \in \mathcal{F}} \left[\left(\widehat{\gamma}_{k \circ f}^2-\gamma_{k \circ f}^2\right)+\gamma_{k \circ f}^2\right] - \inf_{h \in \mathcal{H}}\sup_{f \in \mathcal{F}} {\gamma}_{k \circ f}^2\\
        &\stackrel{\text{Proposition \ref{prop:intro}}}{\leq}& \sup_{h \in \mathcal{H}}\left|\sup_{f \in \mathcal{F}}\left[\left(\widehat{\gamma}_{k \circ f}^2-\gamma_{k \circ f}^2\right)+\gamma_{k \circ f}^2\right] - \sup_{f \in \mathcal{F}} {\gamma}_{k \circ f}^2\right|\\
        &\stackrel{\text{Proposition \ref{prop:intro}}}{\leq}& \sup_{h \in \mathcal{H}}\sup_{f \in \mathcal{F}}\left|\left[\left(\widehat{\gamma}_{k \circ f}^2-\gamma_{k \circ f}^2\right)+\gamma_{k \circ f}^2\right] - {\gamma}_{k \circ f}^2\right|\\
        &=& \sup_{h \in \mathcal{H}, f \in \mathcal{F}} \left|\widehat{\gamma}_{k \circ f}^2-\gamma_{k \circ f}^2\right|.
    \end{eqnarray*}
    Combining the above two results, the inequality is derived.
\end{proof}

\subsection{Proof of Proposition \ref{prop:alternative-1}}
\label{sec:proof-prop42}

\begin{proof}[Proof of Proposition \ref{prop:alternative-1}]
Given a pair of random vectors $X$, $Y$, the corresponding data matrices $\mathbf{X}:=(X_1, \dots, X_m)^T$, $\mathbf{Y}:=(Y_1, \dots, Y_n)^T$, and an associated function set $\mathcal{H}:=\{h \mid h(x, y) = (h_\mathcal{X}(x), h_\mathcal{Y}(y))\}$.
Let ${\boldsymbol{X}}_m$ and ${\boldsymbol{Y}}_n$ be random vectors following the empirical distributions $\frac{1}{m}\sum_{i=1}^m \boldsymbol{\delta}_{X_i}$ and $\frac{1}{n}\sum_{j=1}^n \boldsymbol{\delta}_{Y_j}$ gathered from the data matrices $\boldsymbol{X}_m$ and $\boldsymbol{Y}_n$, respectively. 
According to \cite{gretton2012kernel}, the biased estimator $\widehat{\gamma}_{k,b}(\mathbf{X}, \mathbf{Y})$ of MMD (Definition \ref{def:empiricalMMDestimators}) can be considered as the population MMD, i.e., $\gamma_k$ (Definition \ref{def:MMD}), between random vectors ${\boldsymbol{X}}_m$ and ${\boldsymbol{Y}}_n$.
Thus, we have
\begin{eqnarray*}
    &&\sup_{h \in \mathcal{H}}|\widehat{\gamma}_{k,b}(h_\mathcal{X}(\mathbf{X}), h_\mathcal{Y}(\mathbf{Y})) - {\gamma}_k(h_\mathcal{X}(X), h_\mathcal{Y}(Y))|\\
    &=& \sup_{h \in \mathcal{H}}|{\gamma}_k(h_\mathcal{X}(\boldsymbol{X}_m), h_\mathcal{Y}(\boldsymbol{Y}_n)) - {\gamma}_k(h_\mathcal{X}(X), h_\mathcal{Y}(Y))|.
\end{eqnarray*}
Here, the term $\sup_{h \in \mathcal{H}}|{\gamma}_k(h_\mathcal{X}(\boldsymbol{X}_m), h_\mathcal{Y}(\boldsymbol{Y}_n)) - {\gamma}_k(h_\mathcal{X}(X), h_\mathcal{Y}(Y))|$ is smaller than or equal to the following summation:
\begin{align*}
    \sup_{h \in \mathcal{H}}&|{\gamma}_k(h_\mathcal{X}(\boldsymbol{X}_m), h_\mathcal{Y}(\boldsymbol{Y}_n)) - {\gamma}_k(h_\mathcal{X}(\boldsymbol{X}_m), h_\mathcal{Y}(Y))|\\
    &+ \sup_{h \in \mathcal{H}}|{\gamma}_k(h_\mathcal{X}(\boldsymbol{X}_m), h_\mathcal{Y}(Y)) - {\gamma}_k(h_\mathcal{X}(X), h_\mathcal{Y}(Y))|.
\end{align*}
Note that MMD is a probability metric.
Consequently, based on the triangular inequality for a probability metric, we can combine the above derivations and have the following inequality:
\begin{eqnarray*}
    &&\sup_{h \in \mathcal{H}}|\widehat{\gamma}_{k,b}(h_\mathcal{X}(\mathbf{X}), h_\mathcal{Y}(\mathbf{Y})) - {\gamma}_k(h_\mathcal{X}(X), h_\mathcal{Y}(Y))|\\
    &\leq& \sup_{h_\mathcal{X}}\gamma_k(h_\mathcal{X}(\boldsymbol{X}_m), h_\mathcal{X}(X)) + \sup_{h_\mathcal{Y}}\gamma_k(h_\mathcal{Y}(\boldsymbol{Y}_n), h_\mathcal{Y}(Y)).
\end{eqnarray*}
Next, we focus on terms $\sup_{h_\mathcal{X}}\gamma_k(h_\mathcal{X}(\boldsymbol{X}_m), h_\mathcal{X}(X))$ and $\sup_{h_\mathcal{Y}}\gamma_k(h_\mathcal{Y}(\boldsymbol{Y}_n), h_\mathcal{Y}(Y))$.
Let $\mathbf{X}':=(X_1', \dots, X_m')$ be an independent copy of the data matrix $\mathbf{X}$, $\rho_1, \dots, \rho_m$ be independent uniform $\{\pm 1\}$-valued random variables.
By invoking the symmetrization for $\sup_{h_\mathcal{X}}\gamma_k(h_\mathcal{X}(\boldsymbol{X}_m), h_\mathcal{X}(X))$, we have
\begin{eqnarray*}
    &&\mathbb{E}\sup_{h_\mathcal{X}}\gamma_k(h_\mathcal{X}(\boldsymbol{X}_m), h_\mathcal{X}(X))\\
    &\stackrel{\text{Definition \ref{def:MMD}}}{=}& \mathbb{E}\sup_{h_\mathcal{X}}\left\|\mathbb{E}\left[k(\cdot, h_\mathcal{X}(X))\right] - \frac{1}{m}\sum_{i=1}^m k(\cdot, h_\mathcal{X}(X_i))\right\|_\mathcal{F}\\
    &=& \mathbb{E}_X \sup_{h_\mathcal{X}}\left\|\mathbb{E}_{X'}\left[\frac{1}{m}\sum_{i=1}^m k(\cdot, h_\mathcal{X}(X_i'))\right] - \frac{1}{m}\sum_{i=1}^m k(\cdot, h_\mathcal{X}(X_i))\right\|_\mathcal{F}\\
    &=& \mathbb{E}_X \sup_{h_\mathcal{X}}\left\|\mathbb{E}_{X'} \frac{1}{m}\sum_{i=1}^m \left[k(\cdot, h_\mathcal{X}(X_i)) - k(\cdot, h_\mathcal{X}(X_i'))\right]\right\|_\mathcal{F}\\
    &\stackrel{\text{Jensen's Inequality }(\|\cdot\|_\mathcal{F})}{\leq}& \mathbb{E} \sup_{h_\mathcal{X}}\left\|\frac{1}{m}\sum_{i=1}^m \left[k(\cdot, h_\mathcal{X}(X_i)) - k(\cdot, h_\mathcal{X}(X_i'))\right]\right\|_\mathcal{F}.
    \end{eqnarray*}
The last inequality is derived based on Jensen's inequality, i.e., $\|\mathbb{E}(\cdot)\|_\mathcal{F} \leq \mathbb{E}\|(\cdot)\|_\mathcal{F}$.
Next, since $\mathbf{X}'$ is an independent copy of matrix $\mathbf{X}$, let $\{\rho_i\}_{i=1}^m$ be a set of Rademacher random variables, the symmetric property leads to the following equation:
    \begin{eqnarray*}
    &&\mathbb{E} \sup_{h_\mathcal{X}}\left\|\frac{1}{m}\sum_{i=1}^m \left[k(\cdot, h_\mathcal{X}(X_i)) - k(\cdot, h_\mathcal{X}(X_i'))\right]\right\|_\mathcal{F}\\
    &=& \mathbb{E} \sup_{h_\mathcal{X}}\left\|\frac{1}{m}\sum_{i=1}^m \rho_i\left[k(\cdot, h_\mathcal{X}(X_i)) - k(\cdot, h_\mathcal{X}(X_i'))\right]\right\|_\mathcal{F}\\
    &\leq& 2 \mathbb{E} \sup_{h_\mathcal{X}}\left\|\frac{1}{m}\sum_{i=1}^m \rho_i k(\cdot, h_\mathcal{X}(X_i))\right\|_\mathcal{F}.
    \end{eqnarray*}
Then, based on the property of the reproducing kernel function $k$, that is, $\langle k(x, \cdot), k(x', \cdot)\rangle_\mathcal{F} = k(x, x')$, $\forall x, x' \in \mathcal{U}$, we have
    \begin{eqnarray*}
    &&2 \mathbb{E} \sup_{h_\mathcal{X}}\left\|\frac{1}{m}\sum_{i=1}^m \rho_i k(\cdot, h_\mathcal{X}(X_i))\right\|_\mathcal{F}\\
    &=& 2 \mathbb{E} \sup_{h_\mathcal{X}}\sqrt{\frac{1}{m}\sum_{i=1}^m\sum_{j=1}^m \rho_i \rho_j k(h_\mathcal{X}(X_i), h_\mathcal{X}(X_j))}\\
    &\stackrel{\text{Jensen's Inequality}}{\leq}& \frac{2}{\sqrt{m}}\left[\mathbb{E} \sup_{h_\mathcal{X}}\left|\sum_{i=1}^m\sum_{j=1}^m \rho_i \rho_j k(h_\mathcal{X}(X_i), h_\mathcal{X}(X_j))\right|\right]^{1/2}.
\end{eqnarray*}
Similar bounds can be built for $\mathbb{E}\sup_{h_\mathcal{X}} \gamma_k(h_\mathcal{X}(\boldsymbol{X}_m), h_\mathcal{X}(X))$.

Recall $\sup_{u} k(u,u) \leq \nu$ (Assumption \ref{assum:BandL}), the bounded difference property required in McDiarmid's inequality (Lemma \ref{lem:mcdiarmid}) is satisfied for $\sup_{h_\mathcal{X}}|\widehat{\gamma}_{k,b}(h_\mathcal{X}(\boldsymbol{X}_m), h_\mathcal{X}(X))|$.
Consequently, combining the above results with McDiarmid's inequality (Lemma \ref{lem:mcdiarmid}), the inequality in Proposition \ref{prop:alternative-1} can be derived.
\end{proof}

\vskip 0.2in
\bibliography{refs}

\end{document}